\documentclass[a4paper,fleqn]{cas-dc}

\usepackage[numbers]{natbib}
\usepackage{placeins} 
\usepackage{caption}
\usepackage{graphicx}
\usepackage{subcaption}

\def\tsc#1{\csdef{#1}{\textsc{\lowercase{#1}}\xspace}}
\tsc{WGM}
\tsc{QE}
\tsc{EP}
\tsc{PMS}
\tsc{BEC}
\tsc{DE}

\begin{document}
\let\WriteBookmarks\relax
\def\floatpagepagefraction{1}
\def\textpagefraction{.001}
\shorttitle{Knowledge-Based Systems}
\shortauthors{Guangrui Bai et~al.}

\title [mode = title]{Rethinking Low-Light Image Enhancement: A Log-Domain Intensity--Chromaticity Decoupling Perspective} 

\author[addr1]{Guangrui~Bai}

\author[addr1]{Yifan~Mei}
\author[addr1]{Yahui~Deng}
\author[addr2]{Yuhan~Chen}
\author[addr1]{Yuze~Qiu}

\author[addr1]{Wenhai~Liu}
\author[addr1]{Erbao~Dong\corref{cor1}}[orcid=0000-0002-4062-9730]
\ead{ebdong@ustc.edu.cn}

\address[addr1]{Key Laboratory of Precision and Intelligent Chemistry, Department of Precision Machinery and Precision Instrumentation, University of Science and Technology of China, Hefei, Anhui 230026, China.}

\address[addr2]{College of Mechanical and Vehicle Engineering, Chongqing University, Chongqing 400044, China}

\tnotetext[1]{This work was supported by the National Key R\&D Program of China (Grant No. 2018YFB1307400) and the State Grid Anhui Science and Technology Project.}


\begin{abstract}
	Low-light image enhancement is commonly formulated as a direct RGB-to-RGB restoration problem, where brightness recovery, chromatic correction, and noise suppression are jointly modeled in a coupled space. 
	However, such coupled enhancement often leads to unstable exposure correction, chromatic noise amplification, color shift, and detail degradation, especially under severe underexposure and complex real-world noise. 
	This raises a fundamental question: should low-light enhancement be formulated as a direct RGB-domain restoration problem?
	In this paper, we rethink low-light image enhancement from a representation perspective and propose a log-domain intensity--chromaticity decoupling framework. 
	Instead of directly learning RGB enhancement, we reformulate the task as separate modeling of intensity recovery and chromatic correction. 
	A reversible decoupling transformation and its inverse reconstruction are derived, where the intensity component represents the pixel-wise intensity envelope and the chromaticity component encodes inter-channel relative ratios. 
	Based on this representation, a dual-branch interaction network is constructed to separately model exposure compensation and chromatic consistency. 
	Explicit constraints derived from the decoupled representation are further imposed during reconstruction to suppress abnormal channel amplification and reduce chromatic noise amplification.
	Extensive experiments on LOLv2-Real, MIT-Adobe FiveK, and LSRW demonstrate that the proposed method achieves competitive or superior performance in both quantitative metrics and visual quality, with 29.71 dB PSNR and 0.89 SSIM on LOLv2-Real.
	Experiments on DarkFace further show that the enhanced images improve downstream face detection under low-light conditions.
	Code and pretrained models are available at: \url{https://github.com/mubaisam/ICD}.
\end{abstract}

%
%


\maketitle

\begin{table*}[t]
	\centering
	\caption{Representative LLIE methods grouped by paradigm and representation design.}
	\label{tab:method_comparison}
	\footnotesize
	\setlength{\tabcolsep}{4.0pt}
	\renewcommand{\arraystretch}{1.12}
	\begin{tabular}{p{0.9cm} p{1.2cm} p{2.55cm} p{3.0cm} p{3.35cm} p{4.1cm}}
		\toprule
		\textbf{Year} 
		& \textbf{Venue} 
		& \textbf{Method} 
		& \textbf{Representation} 
		& \textbf{Core mechanism} 
		& \textbf{Remaining gap} \\
		\midrule
		
		\rowcolor{gray!10}
		\multicolumn{6}{l}{\textbf{Prior-based methods}} \\
		\midrule
		1997 & TIP 
		& MSRCR~\cite{Jobson1997_MSRCR} 
		& Illumination--reflectance 
		& Multi-scale Retinex enhancement 
		& Decomposition is often ill-posed. \\
		
		2016 & TIP 
		& LIME~\cite{guo2016lime} 
		& Illumination map/RGB 
		& Illumination-map estimation 
		& Hand-crafted prior; scene-dependent robustness. \\
		
		\midrule
		\rowcolor{gray!10}
		\multicolumn{6}{l}{\textbf{Retinex-based methods}} \\
		\midrule
		2021 & IJCV 
		& KinD++~\cite{zhang2021KIND++} 
		& Retinex-decomposed space 
		& Joint decomposition and restoration 
		& Limited explicit intensity--chromaticity separation. \\
						2021 & CVPR 
		& RUAS~\cite{liu2021ruas} 
		& Illumination-centric space 
		& Retinex-inspired architecture search 
		& Mainly focuses on illumination adjustment. \\
		2022 & CVPR 
		& URetinex~\cite{wu2022uretinex} 
		& Retinex-decomposed space 
		& Deep unfolding Retinex enhancement 
		& Illumination--reflectance separation remains dominant. \\
		
		2023 & CVPR 
		& PairLIE~\cite{fu2023PairLIE} 
		& Paired low-light RGB 
		& Pairwise illumination estimation 
		& Color correction is not explicitly decoupled from intensity. \\

		\midrule
		\rowcolor{gray!10}
		\multicolumn{6}{l}{\textbf{Unsupervised methods}} \\
		\midrule
		2020 & CVPR 
		& Zero-DCE~\cite{guo2020zerodce} 
		& RGB curve space 
		& Zero-reference curve estimation 
		& Brightness and color are adjusted in coupled RGB space. \\
		
		2021 & TPAMI 
		& Zero-DCE++~\cite{li2021zerodce++} 
		& RGB curve space 
		& Efficient curve-based enhancement 
		& RGB-domain coupling remains. \\
		
		2021 & TIP 
		& EnlightenGAN~\cite{jiang2021enlightengan} 
		& RGB image space 
		& Unpaired adversarial enhancement 
		& Color and noise control remain implicit. \\

		2022 & CVPR 
		& SCI~\cite{ma2022SCI} 
		& Illumination space 
		& Self-calibrated illumination learning 
		& Limited explicit chromaticity modeling. \\
				2024 & CVPR 
		& ZERO-IG~\cite{shi2024zeroIG} 
		& Illumination-guided space 
		& Zero-shot enhancement and denoising 
		& Per-image adaptation; limited chromatic decoupling. \\
		
		\midrule
		\rowcolor{gray!10}
		\multicolumn{6}{l}{\textbf{Transformer/diffusion methods}} \\
		\midrule
		2023 & AAAI 
		& LLFormer~\cite{wang2023LLFormer} 
		& RGB/UHD image space 
		& Long-range transformer modeling 
		& Computationally intensive for high-resolution inputs. \\
		
		2023 & ICCV 
		& Diff-Retinex~\cite{yi2023Diffretinex} 
		& Retinex+diffusion 
		& Conditional generative restoration 
		& Multi-stage pipeline with diffusion sampling. \\
				2023 & ICCV 
		& Retinexformer~\cite{cai2023retinexformer} 
		& Retinex-guided space 
		& Illumination-guided transformer 
		& Still illumination-centered. \\
		2024 & ECCV 
		& LightenDiffusion~\cite{lightdiffusion_2024_ECCV} 
		& Latent Retinex space 
		& Unsupervised diffusion enhancement 
		& Diffusion inference remains costly. \\
		
		\midrule
		\rowcolor{gray!10}
		\multicolumn{6}{l}{\textbf{Color-space/noise-aware methods}} \\
		\midrule
		2018 & CVPR 
		& SID~\cite{seeinthedark} 
		& RAW sensor domain 
		& Short-to-long exposure restoration 
		& Requires paired RAW data; RGB output remains coupled. \\
		
		2022 & BMVC 
		& IAT~\cite{Cui_2022_BMVC} 
		& ISP-aware representation 
		& Local--global ISP adjustment 
		& Focuses on ISP parameterization. \\
		
		2022 & CVPR 
		& SNR-Aware~\cite{xu2022snr} 
		& RGB/SNR-guided space 
		& SNR-guided adaptive restoration 
		& No explicit intensity--chromaticity separation. \\

		2025 & CVPR 
		& HVI~\cite{yan2025hvi} 
		& Learned color space 
		& Color--intensity decoupling 
		& Log-domain constrained reconstruction is not explicitly explored. \\
		
		2025 & TPAMI 
		& NoiSER~\cite{zhang2024NoiSER} 
		& RGB/noise-driven learning 
		& Learning from Gaussian noise 
		& Noise-aware, but RGB coupling remains. \\
		
		\bottomrule
	\end{tabular}
\end{table*}
\section{Introduction}
High-quality imaging is fundamental to reliable visual perception in real-world intelligent systems. Outdoor robots, surveillance platforms, mobile cameras, and edge devices often operate under non-uniform illumination, nighttime scenes, shadows, and limited sensor dynamic range, resulting in low contrast, noise amplification, and loss of discriminative structures. These degradations corrupt feature extraction and reduce prediction reliability in downstream tasks such as object detection, semantic segmentation, tracking, and scene understanding. Low-light image enhancement (LLIE) is therefore not only a low-level restoration problem, but also an important front-end for robust machine perception in practical deployments~\cite{yang2018grand,cadena2017past,shahria2022comprehensive}.

Existing LLIE methods have evolved from prior-driven models to learning-based restoration frameworks. Traditional methods, including histogram equalization, gamma correction, and Retinex-based models, are computationally efficient and physically intuitive, but their assumptions are often too restrictive for complex real scenes~\cite{Wang1999_DSIHE,Zuiderveld1994_CLAHE,Land1971,Jobson1997_MSRCR}. Deep learning methods trained with paired or unpaired data have substantially improved visibility and detail recovery on standard benchmarks~\cite{seeinthedark,Chen2018LOLv1,Yang_2021LOLv2,zhang2021KIND++,jiang2021enlightengan,guo2020zerodce}, while recent transformer- and diffusion-based models have further improved perceptual quality through stronger long-range modeling or generative priors~\cite{wang2023LLFormer,cai2023retinexformer,yi2023Diffretinex,lightdiffusion_2024_ECCV}. 
Despite these advances, most existing methods still formulate LLIE as a coupled RGB-domain mapping, where exposure correction, chromatic adjustment, and noise suppression are implicitly handled within a single transformation.
This raises a more fundamental question: should low-light image enhancement be formulated as a direct RGB-to-RGB restoration problem at all?
We argue that the primary difficulty of LLIE lies not in insufficient network capacity, but in the entangled representation itself. 
When exposure recovery, chromatic correction, and noise suppression are jointly optimized in RGB space, brightness amplification and color distortion become inherently coupled, making stable enhancement fundamentally difficult under severe degradation.

The problem becomes more severe in extremely dark scenes, where low photon counts substantially reduce the signal-to-noise ratio and make shot noise and sensor read noise more dominant. Under low illumination, strong enhancement may perturb inter-channel color ratios, leading to color bias, exposure artifacts, and chromatic noise. Recent color-space studies have shown that representation design is important for controlling color distortion in low-light enhancement~\cite{yan2025hvi}. In practice, a single RGB-space mapping is expected to brighten severely under-exposed regions, preserve already visible regions, maintain color consistency, and suppress channel-wise corruption simultaneously. This entanglement reduces controllability and weakens physical interpretability.

This issue becomes more severe in extremely dark scenes. When photon counts are low, the signal-to-noise ratio drops substantially, and the effects of shot noise and sensor read noise become more prominent~\cite{seeinthedark,Foi2008PGNoise}. Aggressive amplification may therefore magnify image structures and noise at the same time. Although noise-aware and Retinex-inspired methods have shown the benefit of spatially adaptive enhancement~\cite{xu2022snr,liu2021ruas,wu2022uretinex}, they still face a trade-off between artifact suppression and texture preservation. As a result, even recent strong methods may improve global brightness but still struggle to preserve local exposure, color stability, and fine details in challenging scenes.

Recent progress in LLIE increasingly depends on large transformer or diffusion models~\cite{wang2023LLFormer,cai2023retinexformer,yi2023Diffretinex,lightdiffusion_2024_ECCV}, suggesting that performance improvements are often pursued through greater model capacity rather than better representation. 
However, larger backbones do not fundamentally resolve the instability caused by coupled RGB-domain enhancement. 
A practical LLIE framework should therefore improve robustness from the representation level, instead of relying on increasingly complex RGB-to-RGB mappings.

We address this problem by reformulating LLIE in a log-domain intensity--chromaticity decoupled space, termed the intensity--chromaticity decoupled (ICD) space. 
Instead of directly enhancing RGB channels, an image is decomposed into an intensity component and a chromaticity component in this representation. 
The intensity component captures the absolute intensity envelope and dominant structural information, while the chromaticity component describes the relative ratios among RGB channels. 
LLIE is therefore reformulated as intensity recovery and chromatic correction in a physically meaningful decoupled domain, rather than unconstrained RGB-to-RGB regression. 
This representation provides better controllability for suppressing color instability and chromatic noise amplification under strong enhancement.

Building on the proposed intensity--chromaticity decoupled (ICD) space, we develop ICDNet, a dual-stream enhancement network for low-light image enhancement. 
The intensity and chromaticity branches are processed separately and coupled through lightweight cross-stream interaction, enabling coordinated exposure recovery and chromatic correction. 
To improve efficiency on high-resolution inputs, we introduce a space-to-channel folding strategy that rearranges local spatial neighborhoods into the channel dimension, allowing most feature extraction to be performed at a lower spatial resolution. 
The final RGB image is reconstructed through a physically constrained inverse mapping, where chromaticity is restricted to a feasible domain and stabilized by a zero-anchor condition. 
This formulation enables stable exposure recovery, robust chromatic correction, and efficient deployment under complex low-light conditions.

In summary, the main contributions of this work are as follows:
\begin{itemize}
	
	\item[1)] \textbf{Representation reformulation for LLIE.}
	We revisit low-light image enhancement from a representation perspective and propose a log-domain intensity--chromaticity decoupling formulation, which reformulates RGB-domain enhancement as separate intensity recovery and chromatic correction in a physically meaningful space.
	
	\item[2)] \textbf{Constrained decoupled enhancement framework.}
	We develop a dual-branch enhancement framework with cross-branch interaction and constrained inverse reconstruction. Explicit non-positive and zero-anchor constraints are imposed on chromaticity to improve color stability and suppress chromatic noise amplification under strong enhancement.
	
	\item[3)] \textbf{Robust performance and practical efficiency.}
	Extensive experiments on multiple public benchmarks and downstream face detection tasks demonstrate that the proposed method achieves superior enhancement quality, stable color restoration, and strong cross-scene generalization, while maintaining efficient inference for practical deployment.
	
\end{itemize}

\begin{figure*}[t]
	\centering
	\includegraphics[width=0.99\linewidth]{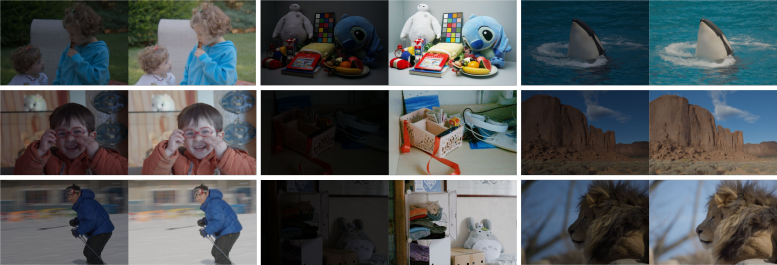}
	\caption{Visual comparison of low-light image enhancement results on representative scenes.}
	\label{fig:loss_Ablation}
\end{figure*}

\section{Related Work}
\label{sec:related_work}

Low-light image enhancement (LLIE) can be broadly grouped into conventional image-processing methods and learning-based methods.
This section reviews these two research lines and discusses why representation coupling remains a fundamental limitation in existing LLIE frameworks.

\subsection{Conventional Methods}
\label{sec:rw_conventional}

Early LLIE methods were mainly developed from hand-crafted priors, including histogram equalization, gamma correction, Retinex theory, and illumination-map estimation~\cite{Wang1999_DSIHE,Zuiderveld1994_CLAHE,Land1971,Jobson1997_MSRCR,guo2016lime}.
Histogram-based methods enhance global or local contrast by redistributing image intensities, while gamma correction adjusts image brightness through nonlinear intensity remapping.
Retinex-based and illumination-map methods model low-light enhancement from the perspective of illumination estimation or illumination--reflectance decomposition.
These methods provide compact and interpretable enhancement schemes, but their performance is often tied to the validity of the adopted priors.
Under severe noise, non-uniform illumination, or mixed degradations, fixed hand-crafted assumptions may limit their adaptability to real scenes.

\subsection{Learning-Based Methods}
\label{sec:rw_learning}

With the availability of paired datasets and deep networks, LLIE has been increasingly formulated as a data-driven restoration problem~\cite{seeinthedark,Chen2018LOLv1,xu2020learning,zhang2019kindling,zhang2021KIND++,fu2023PairLIE}. 
Supervised methods learn mappings from paired low/normal-light images or raw short-/long-exposure data, while Retinex-based networks incorporate illumination--reflectance decomposition into the learning framework. 
Recent studies further improve restoration stability by introducing dual-domain priors, hierarchical illumination guidance, degradation-aware modeling, and probabilistic reformulation~\cite{qi2025dual,lv2026bip,sedeeq2025low,wei2026rethinking}.

These methods provide effective data-driven formulations for low-light restoration. 
In most cases, however, the enhanced image is still reconstructed in the RGB domain, where exposure correction, color adjustment, and noise suppression are jointly modeled.
To reduce the dependence on paired supervision, unpaired, zero-reference, and self-supervised methods have been proposed~\cite{jiang2021enlightengan,guo2020zerodce,li2021zerodce++,liu2021ruas,ma2022SCI}. 
EnlightenGAN learns enhancement from unpaired data. 
Zero-DCE and Zero-DCE++ formulate LLIE as curve estimation under non-reference losses. 
RUAS and SCI perform unsupervised illumination correction through Retinex-inspired or self-calibrated mechanisms. 
These methods reduce the need for paired training data, but most of them still operate on coupled image representations.

Recent transformer- and diffusion-based methods introduce long-range modeling and generative priors into LLIE~\cite{wang2023LLFormer,cai2023retinexformer,yi2023Diffretinex,lightdiffusion_2024_ECCV}. 
LLFormer and Retinexformer use transformer architectures for high-resolution or illumination-guided enhancement, while Diff-Retinex and LightenDiffusion adopt diffusion-based restoration. 
These methods extend the modeling capacity of LLIE, but their enhancement processes are still mainly formulated in RGB, Retinex, or latent spaces.

Several recent studies further consider noise modeling and representation design. 
SNR-aware LLIE introduces spatially adaptive restoration according to local signal-to-noise ratio~\cite{xu2022snr}. 
ZERO-IG combines denoising and adaptive enhancement in a zero-shot setting~\cite{shi2024zeroIG}, and NoiSER studies LLIE from a noise-driven learning perspective~\cite{zhang2024NoiSER}. 
In parallel, HVI redesigns the color representation for low-light enhancement~\cite{yan2025hvi}, and IAT formulates image enhancement from an ISP-aware perspective~\cite{Cui_2022_BMVC}.
These studies suggest that noise modeling, color representation, and image formation priors are critical to robust low-light restoration. 
However, explicit log-domain separation between absolute intensity recovery and inter-channel chromatic correction, together with physically constrained inverse reconstruction, remains largely underexplored.

%

\section{Methodology}
\label{sec:methodology}

Instead of directly learning a coupled RGB-to-RGB enhancement mapping, we reformulate low-light image enhancement in a log-domain intensity--chromaticity decoupled space. 
Each pixel is decomposed into an intensity component that represents the absolute intensity envelope and a chromaticity component that encodes inter-channel relative ratios. 
This reparameterization separates exposure recovery from chromatic correction, reducing the coupling among brightness amplification, color distortion, and noise propagation in conventional RGB-domain enhancement. 
The enhanced image is reconstructed through a constrained inverse transformation, where explicit chromaticity constraints are imposed to suppress abnormal channel-wise amplification and improve color stability under strong enhancement.

\subsection{Image Formation Model}
\label{sec:image_formation_prior}

For a color image, the response of channel \(c\in\{R,G,B\}\) at pixel location \(x\) can be written as
\begin{equation}
	I_c(x)
	=
	\int_{\Lambda} E(x,\lambda) S(x,\lambda) \rho_c(\lambda) d\lambda,
	\label{eq:imaging_integral}
\end{equation}
where \(\Lambda\) denotes the visible spectrum, \(E(x,\lambda)\) is the illumination spectrum, \(S(x,\lambda)\) is the surface reflectance, and \(\rho_c(\lambda)\) is the camera spectral response.

Following Retinex theory and intrinsic image decomposition~\cite{Land197,barrow1978recover}, the image formation can be approximated as the interaction between illumination and reflectance. 
Under the local color constancy assumption, the illumination spectrum can be decomposed as
\begin{equation}
	E(x,\lambda) = L(x)\tilde{E}(\lambda) + \delta_E(x,\lambda),
	\label{eq:illum_decompose}
\end{equation}
where \(L(x)\) denotes the scalar illumination intensity, \(\tilde{E}(\lambda)\) is the normalized spectral shape, and \(\delta_E(x,\lambda)\) accounts for residual variations.

By absorbing wavelength-dependent terms into an effective reflectance, Eq.~\eqref{eq:imaging_integral} can be approximated as
\begin{equation}
	I_c(x) = L(x)R_c(x) + n_c(x), \qquad c\in\{R,G,B\},
	\label{eq:imaging_lr}
\end{equation}
where \(R_c(x)\) denotes the effective reflectance and \(n_c(x)\) collects noise and modeling residuals.

In vector form,
\begin{equation}
	\mathbf{I}(x) = L(x)\mathbf{R}(x) + \mathbf{n}(x),
	\label{eq:vector_lr1}
\end{equation}
where \(\mathbf{I}(x)=[I_R(x), I_G(x), I_B(x)]^\top\). 
When illumination varies smoothly and residual terms are moderate, \(L(x)\) acts as a shared multiplicative factor across channels, while inter-channel ratios are primarily determined by \(\mathbf{R}(x)\). 
This observation motivates separating the absolute intensity scale from relative chromaticity. 
In practice, deviations from this model are treated as residual factors and handled by the subsequent data-driven model.

\subsection{Intensity--Chromaticity Reparameterization}
\label{sec:icd_definition}

Motivated by Eq.~\eqref{eq:vector_lr1}, we define a pixel-wise reparameterization that separates the intensity envelope from inter-channel chromatic ratios. 
For each pixel \(x\), the intensity component is defined as
\begin{equation}
	I_{\max}(x)=\max_{c\in\{R,G,B\}} I_c(x),
	\label{eq:intensity_def}
\end{equation}
which represents the pixel-wise intensity envelope.

To ensure numerical stability, a small constant \(\epsilon>0\) is introduced. 
The log-domain chromaticity of channel \(c\) is defined as
\begin{equation}
	\begin{aligned}
		C_c(x)
		&=
		\log\!\big(I_c(x)+\epsilon\big)
		-
		\log\!\big(I_{\max}(x)+\epsilon\big) \\
		&=
		\log
		\frac{I_c(x)+\epsilon}{I_{\max}(x)+\epsilon}.
	\end{aligned}
	\label{eq:chromaticity_def}
\end{equation}

The RGB image is thus reparameterized as
\begin{equation}
	\begin{aligned}
		\mathbf{I}(x)
		&\longmapsto
		\big(I_{\max}(x),\mathbf{C}(x)\big), \\
		\mathbf{C}(x)
		&=
		\big(C_R(x),C_G(x),C_B(x)\big)^\top.
	\end{aligned}
	\label{eq:forward_icd}
\end{equation}
where \(I_{\max}(x)\) captures the intensity envelope and \(\mathbf{C}(x)\) encodes the relative chromatic structure.

By construction, \(C_c(x)\le 0\) for all channels \(c\), which imposes an explicit non-positive upper bound on chromaticity. 
Moreover, for each pixel, at least one channel satisfies \(C_c(x)=0\), serving as a normalization anchor.

The inverse transformation admits a closed form:
\begin{equation}
	I_c(x)
	=
	\big(I_{\max}(x)+\epsilon\big)\exp\!\big(C_c(x)\big)-\epsilon,
	\quad c\in\{R,G,B\}.
	\label{eq:inverse_icd}
\end{equation}
Thus, the mapping is well-defined and admits a closed-form inverse for non-negative inputs.

Compared with direct RGB parameterization, this formulation explicitly disentangles intensity scaling from inter-channel ratios, providing a more controllable representation for stable enhancement under severe degradation.
\begin{figure}[t] 
	\centering \includegraphics[width=0.99\linewidth]{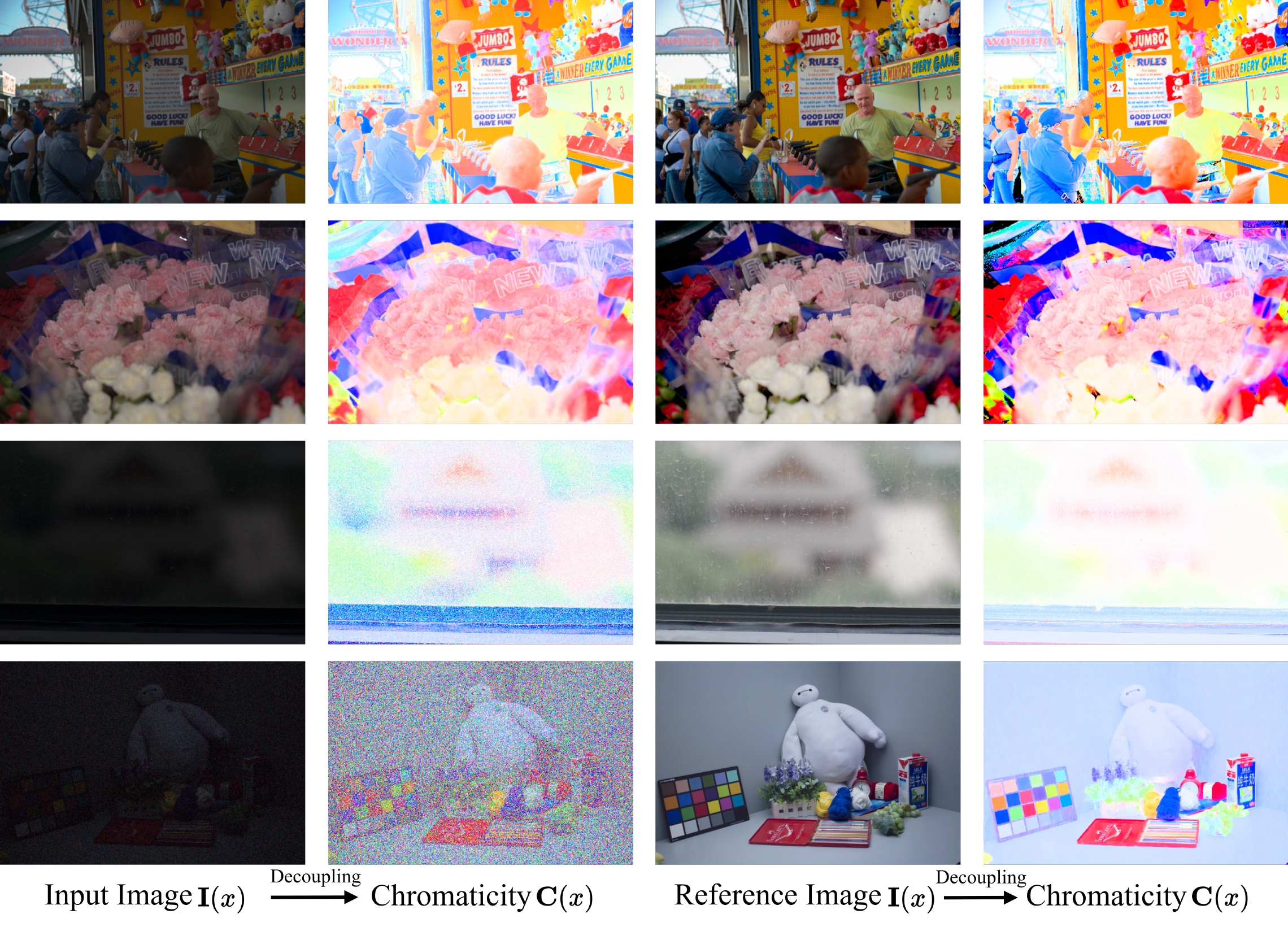} \caption{Visualization of log-chromaticity components for low-light inputs and normal-exposure references.
		The first two columns are from the MIT dataset (primarily illumination differences), and the last two columns are from the LOLv2 dataset (extremely low-light with strong noise).} 
	\label{fig:loss_Ablation}
	\vspace{-2em}
\end{figure}

\subsection{Properties of the Decoupled Representation}
\label{sec:icd_properties}

The proposed representation has several properties that support constrained reconstruction and stable chromatic correction.

\textbf{Non-positive upper bound.}
For any \(c\in\{R,G,B\}\), \(I_c(x)\le I_{\max}(x)\), and thus
\begin{equation}
	\frac{I_c(x)+\epsilon}{I_{\max}(x)+\epsilon}\le 1.
\end{equation}
It follows that
\begin{equation}
	C_c(x)
	=
	\log
	\frac{I_c(x)+\epsilon}{I_{\max}(x)+\epsilon}
	\le 0.
	\label{eq:non_positive_bound}
\end{equation}
The chromaticity component is upper-bounded by zero. 
This defines a natural feasible domain that can be enforced during reconstruction to suppress abnormal channel amplification and color artifacts.

\textbf{Zero-anchor property.}
For each pixel \(x\), there exists at least one channel \(c^\star\) such that
\begin{equation}
	I_{c^\star}(x)=I_{\max}(x),
\end{equation}
which yields
\begin{equation}
	C_{c^\star}(x)
	=
	\log
	\frac{I_{\max}(x)+\epsilon}{I_{\max}(x)+\epsilon}
	=
	0.
	\label{eq:zero_anchor}
\end{equation}
Each pixel contains at least one zero-valued chromaticity component, which serves as a pixel-wise normalization anchor.

\textbf{Relative-ratio property.}
From Eq.~\eqref{eq:chromaticity_def}, when \(I_c(x)\) and \(I_{\max}(x)\) are both much larger than \(\epsilon\), we have
\begin{equation}
	C_c(x)
	=
	\log
	\frac{I_c(x)+\epsilon}{I_{\max}(x)+\epsilon}
	\approx
	\log
	\frac{I_c(x)}{I_{\max}(x)}.
	\label{eq:relative_ratio}
\end{equation}
Chromaticity mainly encodes relative inter-channel ratios, whereas \(I_{\max}(x)\) carries the absolute intensity scale.

\textbf{Approximate illumination invariance.}
Under the imaging model \(I_c(x)=L(x)R_c(x)+n_c(x)\), when the residual term \(n_c(x)\) is relatively small, does not change the channel ordering, and the signal level is sufficiently larger than \(\epsilon\), we have
\begin{equation}
	I_{\max}(x)\approx L(x)R_{\max}(x),
\end{equation}
where \(R_{\max}(x)=\max_c R_c(x)\). 
Substituting into Eq.~\eqref{eq:chromaticity_def} yields
\begin{equation}
	C_c(x)
	\approx
	\log
	\frac{L(x)R_c(x)}{L(x)R_{\max}(x)}
	=
	\log
	\frac{R_c(x)}{R_{\max}(x)}.
\end{equation}
Common illumination scaling is mainly reflected in the intensity component, whereas chromaticity is governed by inter-channel reflectance ratios and remains approximately invariant to illumination changes.

\subsection{Noise Behavior in ICD Space}
\label{sec:noise_behavior_icd}

Low-light enhancement is inherently affected by noise propagation. 
Let the observed RGB signal at pixel \(x\) be
\begin{equation}
	\mathbf{I}(x)=\mathbf{S}(x)+\boldsymbol{\eta}(x),
	\label{eq:obs_noise_model}
\end{equation}
where \(\mathbf{S}(x)\) denotes the latent clean signal and \(\boldsymbol{\eta}(x)\) denotes the observation noise. 
For a direct RGB-domain enhancement mapping \(\hat{\mathbf{I}}(x)=f(\mathbf{I}(x))\), a first-order approximation around \(\mathbf{S}(x)\) gives
\begin{equation}
	\delta\hat{\mathbf{I}}(x)
	\approx
	\mathbf{J}_f\big(\mathbf{S}(x)\big)\boldsymbol{\eta}(x),
	\label{eq:rgb_noise_linearization}
\end{equation}
where \(\mathbf{J}_f\) is the Jacobian of the mapping with respect to the input RGB vector. 
In dark regions, enhancement usually requires large local responses, and the corresponding Jacobian can amplify noise together with image structures. 
Channel-wise noise discrepancies may then be transformed into chromatic artifacts or local color bias.
%

\begin{figure*}[t]
	\centering
	\includegraphics[width=0.99\linewidth]{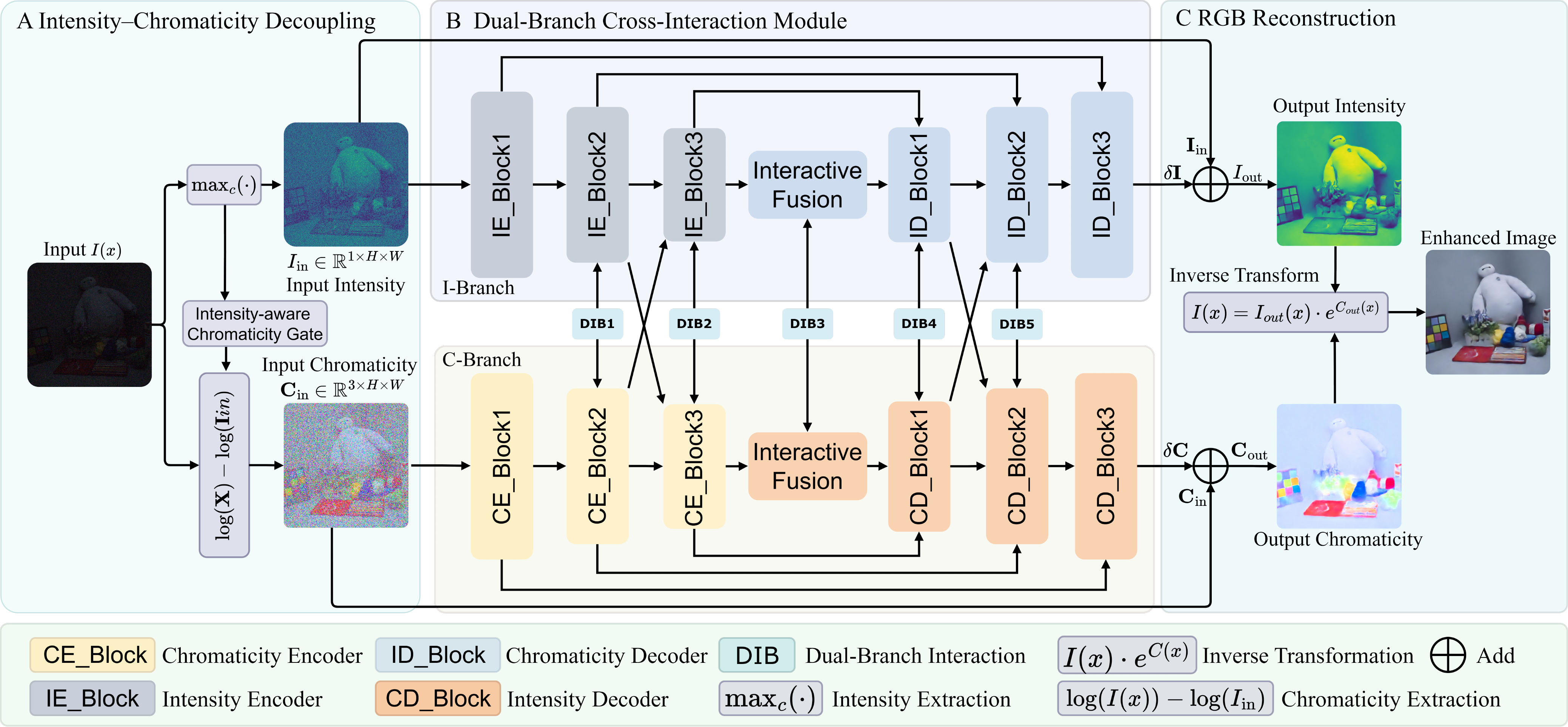}
	\caption{
		Framework of the proposed ICDNet. 
		The input low-light image is transformed into a log-domain intensity--chromaticity decoupled representation, where the intensity component captures the absolute intensity scale and the chromaticity component encodes inter-channel relative ratios. 
		The two components are processed by a dual-branch interaction backbone to predict intensity and chromaticity residuals. 
		The enhanced RGB image is reconstructed through an inverse transformation with explicit chromaticity constraints to improve color stability and reduce chromatic noise amplification.
	}
	\label{fig:framework}
\end{figure*}

The ICD representation changes how such perturbations enter the enhancement process. 
For analysis, we consider a local region where the maximum response is attained by channel \(m\) and the channel ordering remains unchanged:
\begin{equation}
	I_{\max}(x)=I_m(x).
\end{equation}
The log-chromaticity component can then be written as
\begin{equation}
	C_c(x)
	=
	\log
	\frac{I_c(x)+\epsilon}{I_m(x)+\epsilon}.
\end{equation}
Substituting \(I_c(x)=S_c(x)+\eta_c(x)\) and applying a first-order expansion gives
\begin{equation}
	C_c(x)
	\approx
	\log
	\frac{S_c(x)+\epsilon}{S_m(x)+\epsilon}
	+
	\frac{\eta_c(x)}{S_c(x)+\epsilon}
	-
	\frac{\eta_m(x)}{S_m(x)+\epsilon}.
	\label{eq:chroma_noise_linearization}
\end{equation}
This expression shows that noise in chromaticity appears through relative inter-channel perturbations. 
In extremely dark regions, where the signal level is low, these perturbations can still be unreliable due to the small denominators in Eq.~\eqref{eq:chroma_noise_linearization}. 
Therefore, ICD should not be interpreted as eliminating noise by itself. 
Instead, it separates intensity recovery from chromatic correction, allowing brightness amplification to be mainly modeled by the intensity component while chromatic responses are handled separately.

The non-positive chromaticity constraint further stabilizes inverse reconstruction. 
Since the chromaticity component satisfies \(C_c(x)\leq 0\), we enforce
\begin{equation}
	C_c^{\mathrm{out}}(x)\leq 0,
\end{equation}
which gives
\begin{equation}
	\exp\!\big(C_c^{\mathrm{out}}(x)\big)\leq 1.
\end{equation}
Using the inverse transformation,
\begin{equation}
	I_{\mathrm{out},c}(x)
	=
	\big(I_{\max}^{\mathrm{out}}(x)+\epsilon\big)
	\exp\!\big(C_c^{\mathrm{out}}(x)\big)
	-\epsilon,
	\label{eq:constrained_rgb_reconstruction}
\end{equation}
we obtain
\begin{equation}
	I_{\mathrm{out},c}(x)\leq I_{\max}^{\mathrm{out}}(x),
	\qquad c\in\{R,G,B\}.
	\label{eq:channel_upperbound}
\end{equation}
Each reconstructed RGB channel is therefore bounded by the output intensity envelope, which limits abnormal channel-wise over-amplification. 
This constraint helps reduce the risk of chromatic noise, color speckles, and local color bias under strong enhancement.

Together with constrained reconstruction, the ICD representation mitigates chromatic noise amplification at the representation level rather than as a post-processing step.

\section{Proposed Method}
\label{sec:method}

Low-light image enhancement aims to improve scene visibility while preserving natural color and structural details. However, in the standard RGB representation, brightness correction, color adjustment, and noise amplification are interdependent, complicating dark region enhancement due to color distortion and noise amplification. To address this, we reformulate LLIE in a log-domain intensity-chromacity decoupling space, where intensity recovery and chromatic correction are modeled separately.

As shown in Fig.~\ref{fig:framework}, the proposed framework consists of three components: intensity-chromacity decoupling, a dual-branch interaction backbone, and constrained RGB reconstruction. Given an input image, the network first decomposes it into intensity and chromaticity components in the log domain. These components are then processed through dedicated branches, with cross-branch interaction, to jointly model brightness enhancement and color consistency. Finally, the enhanced image is reconstructed through an inverse transformation with explicit constraints.

\subsection{Intensity--Chromaticity Decoupling}
\label{sec:icd_decoupling}

To reduce the synchronized noise amplification caused by coupled RGB-domain enhancement, we decouple the absolute intensity scale and inter-channel chromatic ratios at the representation level. 
Given an input image \(\mathbf{I}_{\mathrm{in}}\in[0,1]^{3\times H\times W}\), the intensity component is defined as
\begin{equation}
	I_{\max}(\mathbf{I}_{\mathrm{in}})(x)
	=
	\max_{c\in\{R,G,B\}}
	I_{\mathrm{in},c}(x),
	\label{eq:net_Iin}
\end{equation}
where \(I_{\mathrm{in},c}(x)\) denotes the response of channel \(c\) at pixel location \(x\). 
This component describes the absolute intensity scale of each pixel.

A small constant \(\epsilon=10^{-4}\) is introduced for numerical stability. 
For any \(c\in\{R,G,B\}\), the chromaticity component is defined as
\begin{equation}
	C_c(\mathbf{I}_{\mathrm{in}})(x)
	=
	\log\!\big(
	I_{\mathrm{in},c}(x)+\epsilon
	\big)
	-
	\log\!\big(
	I_{\max}(\mathbf{I}_{\mathrm{in}})(x)+\epsilon
	\big),
	\label{eq:net_Cin}
\end{equation}
which measures the log-ratio between channel \(c\) and the maximum channel response.

The corresponding vector form is
\begin{equation}
	\begin{gathered}
		I_{\max}(\mathbf{I}_{\mathrm{in}})
		=
		\max_{c\in\{R,G,B\}}
		\mathbf{I}_{\mathrm{in},c},
		\\
		\mathbf{C}(\mathbf{I}_{\mathrm{in}})
		=
		\log(\mathbf{I}_{\mathrm{in}}+\epsilon)
		-
		\log\!\big(
		I_{\max}(\mathbf{I}_{\mathrm{in}})+\epsilon
		\big),
	\end{gathered}
	\label{eq:net_ic_form}
\end{equation}
where
\(
I_{\max}(\mathbf{I}_{\mathrm{in}})
\in
[0,1]^{1\times H\times W}
\)
is a single-channel scalar field and
\(
\mathbf{C}(\mathbf{I}_{\mathrm{in}})
\in
\mathbb{R}^{3\times H\times W}
\)
is a three-channel vector field.

For each pixel, the chromaticity satisfies
\[
C_c(x)
\in
\left[
\log\frac{\epsilon}{1+\epsilon},\,0
\right],
\qquad
c\in\{R,G,B\},
\]
and at least one channel satisfies
\[
C_c(x)=0.
\]
Therefore, the chromaticity component is naturally upper-bounded by zero and always contains a zero-anchor channel at each pixel.

In ideal cases where low-light degradation mainly corresponds to a common intensity scaling, the raw chromaticity \(\mathbf{C}_{\mathrm{raw}}\) is expected to remain approximately stable. 
However, real low-light images usually contain additional noise, sensor artifacts, and local distortions, especially in extremely dark regions where chromaticity estimates can be dominated by unreliable channel-wise perturbations. 
Before feeding the chromaticity component into the chromaticity branch, we therefore introduce an intensity-aware chromaticity gate to attenuate low-confidence chromatic cues.

Let
\(
\mathbf{C}_{\mathrm{raw}}=\mathbf{C}(\mathbf{I}_{\mathrm{in}})
\).
The gated chromaticity input is defined as
\begin{equation}
	\mathbf{C}_{\mathrm{in}}
	=
	\mathcal{G}_{c}
	\big(
	I_{\max}(\mathbf{I}_{\mathrm{in}})
	\big)
	\odot
	\mathbf{C}_{\mathrm{raw}},
	\label{eq:chroma_gate}
\end{equation}
where \(\mathcal{G}_{c}(\cdot)\) denotes an intensity-aware chromaticity gate. 
Specifically, the gate is parameterized as
\begin{equation}
	\mathcal{G}_{c}(I)
	=
	\alpha
	+
	(1-\alpha)
	\left[
	\sin
	\left(
	\frac{\pi}{2}I
	\right)
	\right]^{\gamma},
	\label{eq:chroma_gate_func}
\end{equation}
where \(\alpha\) controls the lower confidence bound and \(\gamma\) is a learnable parameter constrained within a predefined range. 
This gate keeps chromaticity nearly unchanged in brighter regions while suppressing unreliable chromatic responses in low-intensity regions. 
Since \(\mathcal{G}_{c}(I)\in[\alpha,1]\), this operation preserves the non-positive upper bound and zero-anchor property of the chromaticity representation.

This reparameterization separates low-light enhancement into two subproblems: intensity recovery and chromatic correction. 
The intensity component models brightness compensation, while the chromaticity component preserves relative color structure and provides a more stable basis for suppressing chromatic noise amplification under strong enhancement.
\subsection{Dual-Branch Interaction Backbone}
\label{sec:dual_branch_backbone}

Based on the decoupled representation, we construct a dual-branch backbone consisting of an intensity branch and a chromaticity branch. 
The intensity branch takes \(I_{\max}(\mathbf{I}_{\mathrm{in}})\) as input and models exposure compensation together with global brightness recovery, while the chromaticity branch takes \(\mathbf{C}(\mathbf{I}_{\mathrm{in}})\) as input and focuses on inter-channel ratio correction, local texture restoration, and color consistency preservation.

Both branches follow an encoder--decoder architecture with symmetric multi-scale feature extraction and reconstruction. 
Skip connections are introduced between corresponding encoder and decoder stages to preserve structural details and stabilize feature propagation across scales.

To enable controlled information exchange between the two branches, dual-branch interaction blocks (DIBs) are inserted at multiple corresponding stages. 
Let \(\mathbf{F}_{\mathrm{I}}^{(l)}\) and \(\mathbf{F}_{\mathrm{C}}^{(l)}\) denote the intensity and chromaticity features at the \(l\)-th interaction stage, respectively. 
The interaction process is formulated as
\begin{equation}
	\tilde{\mathbf{F}}_{\mathrm{I}}^{(l)},
	\,
	\tilde{\mathbf{F}}_{\mathrm{C}}^{(l)}
	=
	\mathrm{DIB}^{(l)}
	\left(
	\mathbf{F}_{\mathrm{I}}^{(l)},
	\mathbf{F}_{\mathrm{C}}^{(l)}
	\right),
	\label{eq:dib_exchange}
\end{equation}
where
\(
\tilde{\mathbf{F}}_{\mathrm{I}}^{(l)}
\)
and
\(
\tilde{\mathbf{F}}_{\mathrm{C}}^{(l)}
\)
denote the updated features after interaction. 
This design allows structural and intensity cues from the intensity branch to regularize chromatic correction, while edge and local texture information from the chromaticity branch provide complementary guidance for intensity recovery.

An additional fusion module is introduced at the bottleneck stage to further aggregate deep features from both branches and enhance global cross-branch representation. 
Based on the fused features, the decoder predicts the intensity residual and the chromaticity residual as
\begin{equation}
	\Delta I
	=
	\mathcal{G}_{\mathrm{I}}(\mathbf{F}_{\mathrm{I}}),
	\qquad
	\Delta \mathbf{C}
	=
	\mathcal{G}_{\mathrm{C}}(\mathbf{F}_{\mathrm{C}}),
	\label{eq:delta_predict_new}
\end{equation}
where
\(
\mathcal{G}_{\mathrm{I}}
\)
and
\(
\mathcal{G}_{\mathrm{C}}
\)
denote the prediction heads of the two branches, respectively. 
Residual prediction reformulates enhancement as a correction process in the decoupled space, which improves optimization stability and facilitates detail-preserving reconstruction.
\subsection{Constrained RGB Reconstruction}
\label{sec:constrained_reconstruction}

Let the input image be denoted as \(\mathbf{I}_{\mathrm{in}}\), with corresponding intensity and chromaticity components defined as
\begin{equation}
I_{\max,\mathrm{in}} = I_{\max}(\mathbf{I}_{\mathrm{in}}), \quad \mathbf{C}_{\mathrm{in}} = \mathbf{C}(\mathbf{I}_{\mathrm{in}}).
\end{equation}

The network predicts intensity residual \(\Delta I\) and chromaticity residual \(\Delta \mathbf{C}\), which are added to the input components as:
\begin{equation}
	\begin{aligned}
		\tilde{I}_{\max,\mathrm{out}} &= I_{\max,\mathrm{in}} + \Delta I, \\
		\tilde{\mathbf{C}}_{\mathrm{out}} &= \mathbf{C}_{\mathrm{in}} + \Delta \mathbf{C}.
	\end{aligned}
	\label{eq:residual_add}
\end{equation}

To ensure the output components stay within the feasible domain of the decoupled representation, explicit constraints are applied before inverse transformation.

For the intensity component, a lower-bound constraint is applied to avoid numerical instability in dark regions:
\begin{equation}
	I_{\max,\mathrm{out}} = \max\left( \tilde{I}_{\max,\mathrm{out}}, \epsilon \right),
	\label{eq:net_Iout}
\end{equation}
where \(\epsilon\) is a small constant.

For the chromaticity component, we impose an upper bound constraint:
\begin{equation}
	\mathbf{C}_{\mathrm{out}} = \min\left( \tilde{\mathbf{C}}_{\mathrm{out}}, 0 \right),
	\label{eq:net_Cout}
\end{equation}
which prevents abnormal positive responses caused by excessive amplification of individual channels.

The enhanced RGB image is then reconstructed using the inverse transformation:
\begin{equation}
	\mathbf{I}_{\mathrm{out}} = \left( I_{\max,\mathrm{out}} + \epsilon \right) \odot \exp\left( \mathbf{C}_{\mathrm{out}} \right) - \epsilon,
	\label{eq:net_rgb_recon_new}
\end{equation}
followed by clipping the values to the valid range \([0, 1]\).

Since \(\mathbf{C}_{\mathrm{out}} \leq 0\), we have:
\begin{equation}
	\exp\left( \mathbf{C}_{\mathrm{out}} \right) \leq 1.
\end{equation}
Thus, for any channel \(c \in \{R, G, B\}\), the reconstructed intensity is bounded by the output intensity envelope:
\begin{equation}
	I_{\mathrm{out}, c}(x) \leq I_{\max,\mathrm{out}}(x).
	\label{eq:channel_bound}
\end{equation}

This shows that chromatic correction is always bounded by the output intensity envelope. The inverse reconstruction suppresses abnormal channel over-amplification and reduces chromatic noise, color speckles, and local color distortion under strong enhancement. Such constrained reconstruction ensures stable performance and preserves image consistency in low-light conditions.
\subsection{Training Objective}
\label{sec:training_objective}

The proposed network is trained end-to-end with joint supervision in both the RGB domain and the intensity--chromaticity decoupled space. 
Let
\(
\mathbf{I}_{\mathrm{out}}
\in
[0,1]^{3\times H\times W}
\)
denote the enhanced image and
\(
\mathbf{I}^{*}
\in
[0,1]^{3\times H\times W}
\)
denote the corresponding reference image.

The overall objective consists of an RGB reconstruction loss and an intensity--chromaticity consistency loss.

\textbf{RGB reconstruction loss.}
The RGB reconstruction loss enforces pixel-wise fidelity and structural consistency:
\begin{equation}
	\mathcal{L}_{\mathrm{rgb}}
	=
	\|
	\mathbf{I}_{\mathrm{out}}
	-
	\mathbf{I}^{*}
	\|_{1}
	+
	\mathcal{L}_{\mathrm{SSIM}}
	\left(
	\mathbf{I}_{\mathrm{out}},
	\mathbf{I}^{*}
	\right),
	\label{eq:loss_rgb}
\end{equation}
where
\(
\|\cdot\|_{1}
\)
denotes the pixel-wise \(\ell_{1}\) distance and
\(
\mathcal{L}_{\mathrm{SSIM}}(\cdot,\cdot)
\)
denotes the structural similarity loss.

\textbf{Intensity--chromaticity consistency loss.}
Following the decoupled representation defined in Section~\ref{sec:icd_decoupling}, the intensity and chromaticity components are computed as
\begin{equation}
	\begin{gathered}
		I_{\max}(\mathbf{I})
		=
		\max_{c\in\{R,G,B\}}
		I_c,
		\\
		\mathbf{C}(\mathbf{I})
		=
		\log(\mathbf{I}+\epsilon)
		-
		\log
		\big(
		I_{\max}(\mathbf{I})+\epsilon
		\big),
	\end{gathered}
	\label{eq:IC_def}
\end{equation}
where
\(
\epsilon>0
\)
is the numerical stability constant.

Accordingly,
\begin{equation}
	\begin{aligned}
		I_{\max,\mathrm{out}}
		&=
		I_{\max}(\mathbf{I}_{\mathrm{out}}),
		\qquad
		I_{\max}^{*}
		=
		I_{\max}(\mathbf{I}^{*}),
		\\
		\mathbf{C}_{\mathrm{out}}
		&=
		\mathbf{C}(\mathbf{I}_{\mathrm{out}}),
		\qquad
		\mathbf{C}^{*}
		=
		\mathbf{C}(\mathbf{I}^{*}).
	\end{aligned}
	\label{eq:IC_targets}
\end{equation}

The intensity consistency loss is defined as
\begin{equation}
	\mathcal{L}_{I}
	=
	\|
	I_{\max,\mathrm{out}}
	-
	I_{\max}^{*}
	\|_{1}
	+
	\mathcal{L}_{\mathrm{SSIM}}
	\left(
	I_{\max,\mathrm{out}},
	I_{\max}^{*}
	\right),
	\label{eq:loss_I}
\end{equation}
and the chromaticity consistency loss is defined as
\begin{equation}
	\mathcal{L}_{C}
	=
	\|
	\mathbf{C}_{\mathrm{out}}
	-
	\mathbf{C}^{*}
	\|_{1}.
	\label{eq:loss_C}
\end{equation}

The intensity component constrains exposure compensation and absolute intensity recovery, while the chromaticity component preserves inter-channel ratios and color consistency. 
This supervision preserves representation consistency in the decoupled space and improves robustness against chromatic noise amplification caused by direct RGB-domain enhancement.

The overall training objective is
\begin{equation}
	\mathcal{L}
	=
	\mathcal{L}_{\mathrm{rgb}}
	+
	\lambda_{I}\mathcal{L}_{I}
	+
	\lambda_{C}\mathcal{L}_{C},
	\label{eq:loss_total}
\end{equation}
where
\(
\lambda_{I}
\)
and
\(
\lambda_{C}
\)
are balancing weights.

\section{Experiments}
\label{sec:experiments}

We evaluate the proposed method on standard low-light image enhancement benchmarks. 
This section presents the implementation details, quantitative comparisons, and visual results.

\subsection{Implementation Details}
\label{sec:implementation_details}

All experiments are conducted on a workstation with an AMD EPYC 7Y43 48-core CPU, 256\,GB RAM, and four NVIDIA GeForce RTX 4090 GPUs. 
The software environment is Ubuntu 22.04.5 LTS with PyTorch. 
Multi-GPU training is implemented using DataParallel.
Input images are resized to \(128\times128\) during training. 
The network is optimized with AdamW using a weight decay of \(1\times10^{-6}\). 
The initial learning rate is \(1\times10^{-5}\) and is increased to \(5\times10^{-5}\) during the pretraining-to-finetuning stage, followed by cosine annealing. 
The batch size is 128 and the model is trained for 5000 epochs. 
Gradient clipping is applied with a maximum norm of 1.0.
For the joint objective in Section~\ref{sec:training_objective}, we set
\(\lambda_I=1500\) and \(\lambda_C=2500\). 
In the target-intensity pretraining stage, Smooth \(\ell_1\) loss with \(\beta=0.01\) is used to supervise the predicted intensity mean.
\subsection{Datasets and Evaluation Metrics}
\label{sec:datasets_metrics}

We evaluate the proposed method on five paired low-light benchmarks, including the Real-captured and Synthetic subsets of LOLv2~\cite{Yang_2021LOLv2}, MIT-Adobe FiveK~\cite{fivekMIT,bychkovsky2011learning}, and the Huawei and Nikon subsets of LSRW~\cite{hai2023lsrw}.

LOLv2~\cite{Yang_2021LOLv2} extends the original LOL benchmark~\cite{Chen2018LOLv1} and contains both real-captured and synthetic low-light image pairs. 
MIT-Adobe FiveK consists of high-resolution DSLR images with professionally retouched references. 
LSRW-Huawei and LSRW-Nikon provide real low-light images captured by mobile devices and professional cameras, respectively~\cite{hai2023lsrw}.
We report peak signal-to-noise ratio (PSNR) and structural similarity index (SSIM) for quantitative evaluation.
\newcommand{\panelSix}[2]{
	\begin{minipage}{0.162\textwidth}
		\centering
		\includegraphics[width=\linewidth]{#1}\\[1pt]
		\small #2
	\end{minipage}
}
\newcommand{\panelVisCompare}[2]{
	\begin{minipage}[t]{0.162\textwidth}
		\centering
		\includegraphics[width=\linewidth]{#1}\par\vspace{1pt}
		\small #2\par\vspace{3pt}
	\end{minipage}
}
\newcommand{\panelMetric}[3]{%
	\shortstack[t]{%
		\includegraphics[width=0.16\textwidth]{#1}\\[-0.5ex]
		\small\strut #2\\[-1.0ex]
		\footnotesize\strut #3\\[-2.0ex]%
	}%
}
\begin{figure*}[t]
	\centering
	\setlength{\tabcolsep}{1pt}
	\renewcommand{\arraystretch}{1.0}
	\begin{tabular}{cccccc}
		\panelVisCompare{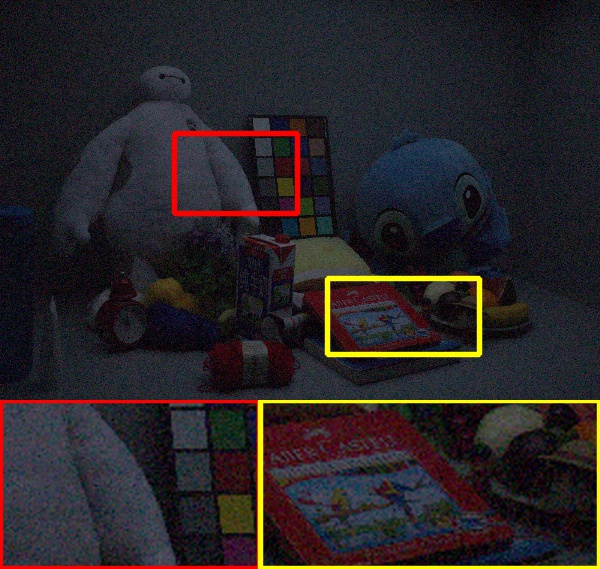}{Input} &
		\panelVisCompare{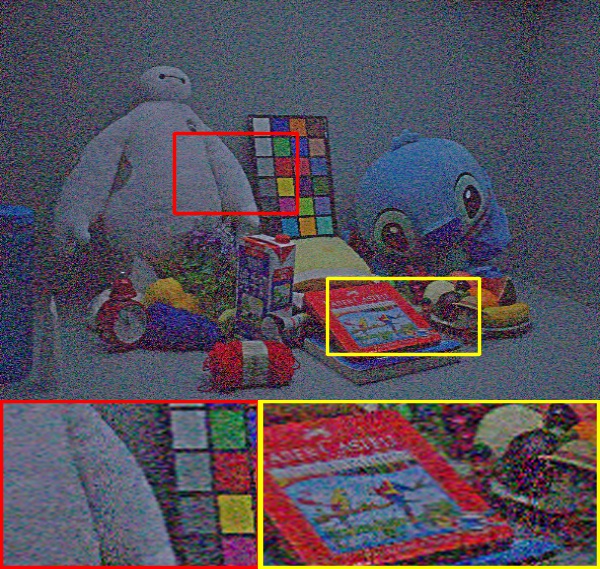}{CLIP-LIT~\cite{liang2023CLIPLIT}} &
		\panelVisCompare{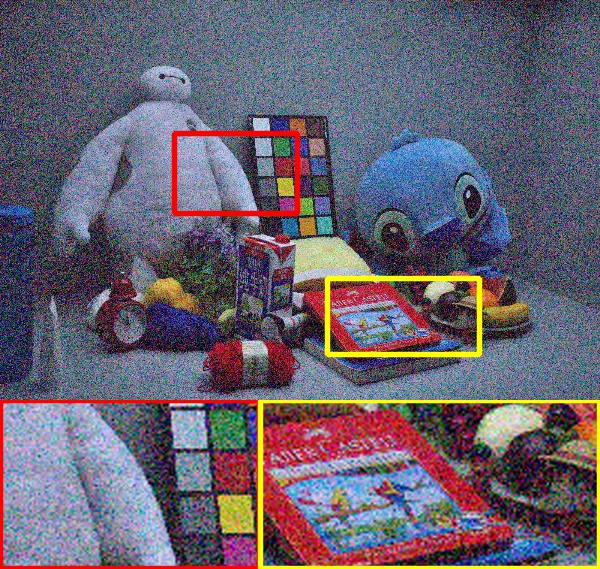}{CoLIE~\cite{chobola2025CoLIE}} &
		\panelVisCompare{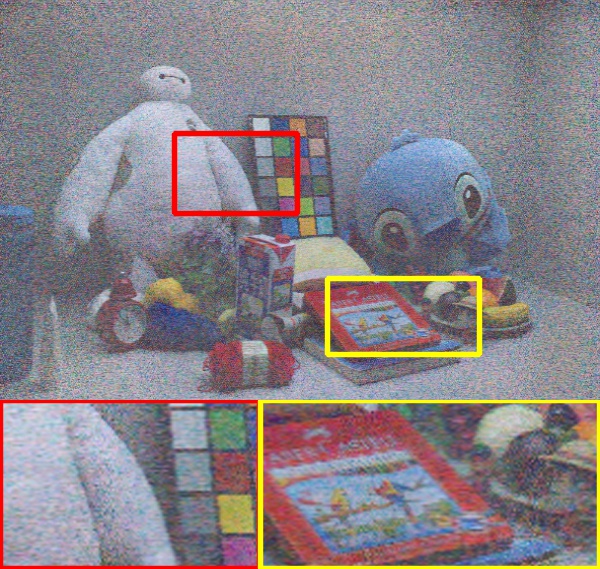}{EnGAN~\cite{jiang2021enlightengan}} &
		\panelVisCompare{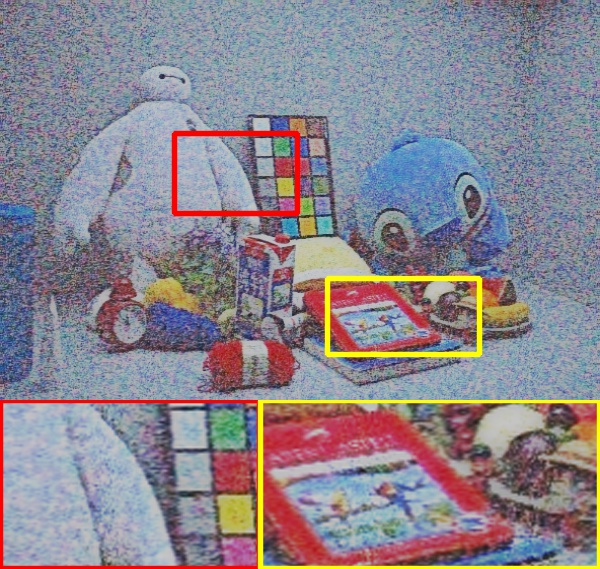}{PairLIE~\cite{fu2023PairLIE}} &
		\panelVisCompare{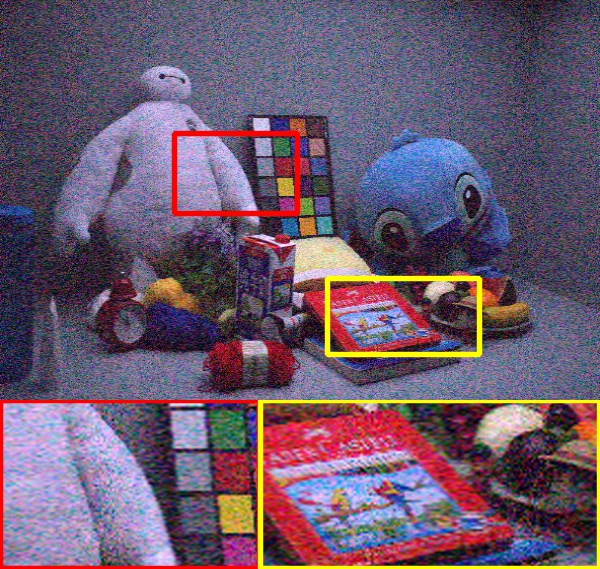}{RUAS~\cite{liu2021ruas}} \\
		
		\panelVisCompare{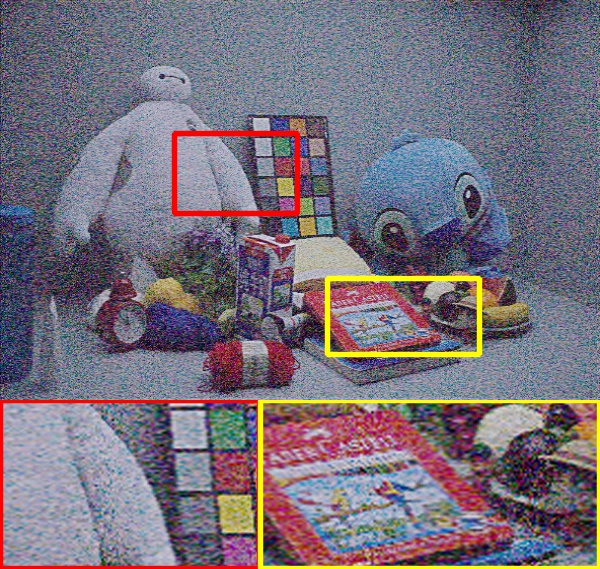}{SCI~\cite{ma2022SCI}} &
		\panelVisCompare{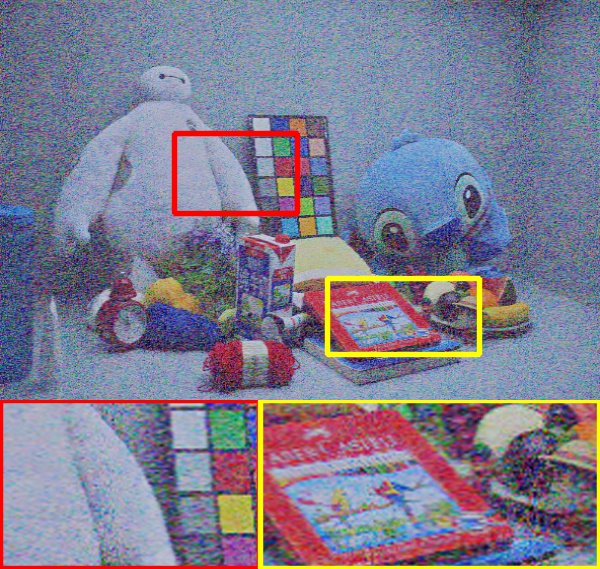}{URetinex~\cite{wu2022uretinex}} &
		\panelVisCompare{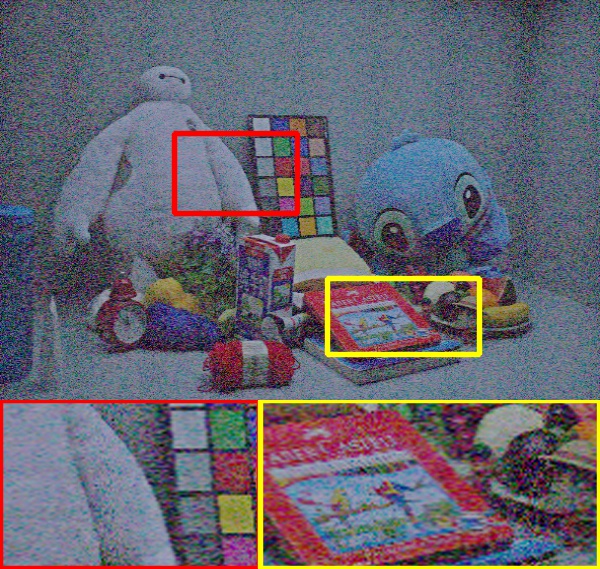}{ZeroDCE~\cite{guo2020zerodce}} &
		\panelVisCompare{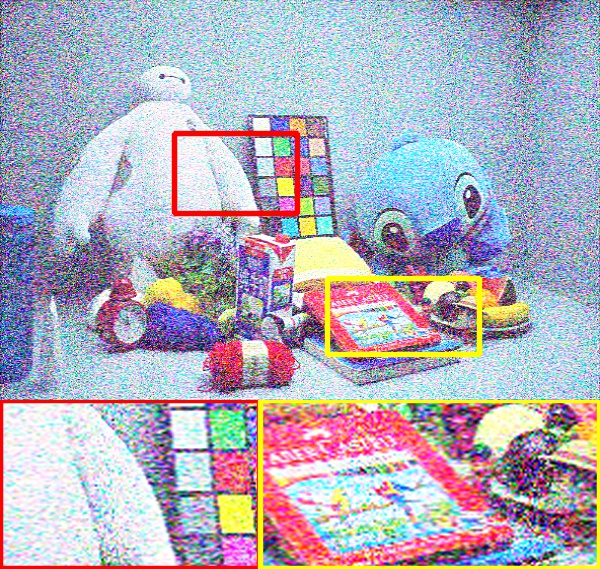}{ZeroIG~\cite{shi2024zeroIG}} &
		\panelVisCompare{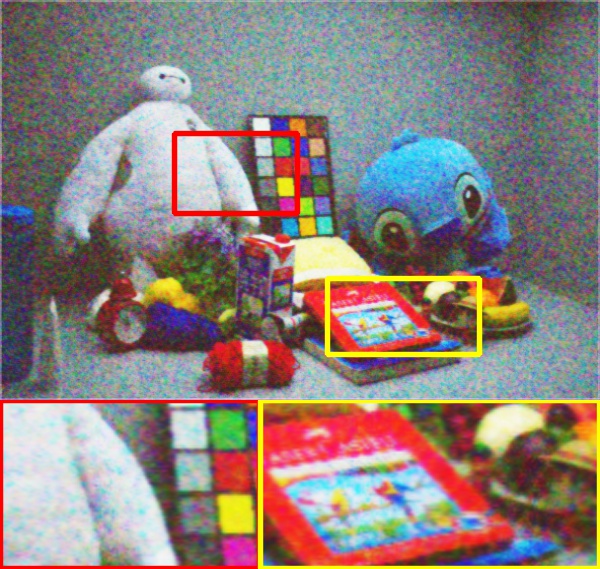}{NoiSER~\cite{zhang2024NoiSER}} &
		\panelVisCompare{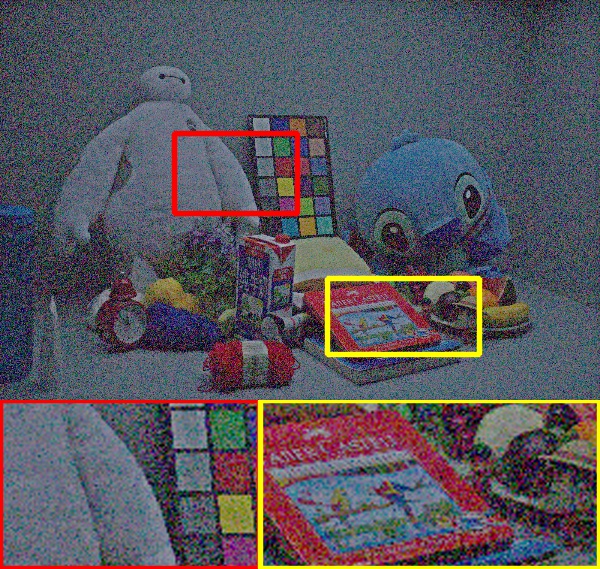}{SCL-LLE~\cite{liang2022SCLLLE}} \\
		
		\panelVisCompare{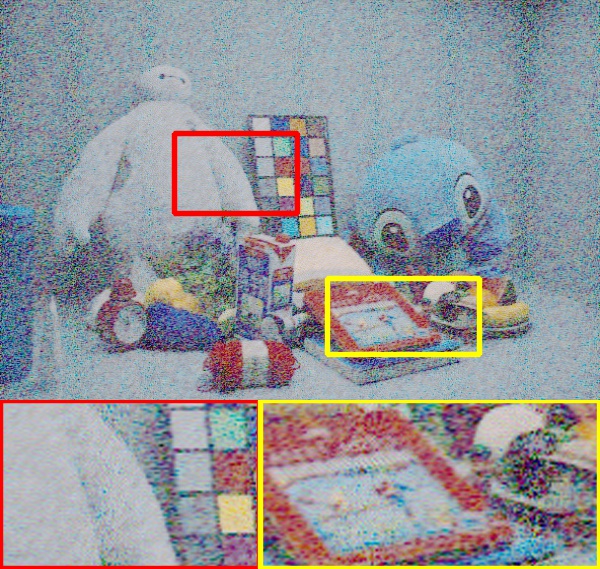}{SCLM~\cite{zhang2023SCLM}} &
		\panelVisCompare{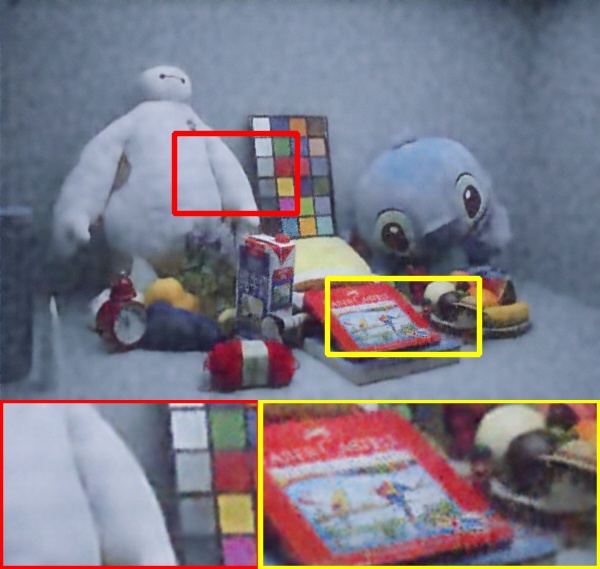}{LightenDiff~\cite{lightdiffusion_2024_ECCV}} &
		\panelVisCompare{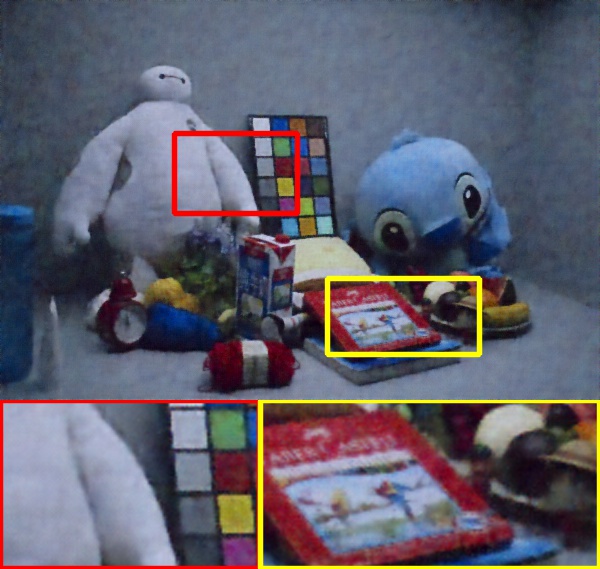}{Retinexformer~\cite{cai2023retinexformer}} &
		\panelVisCompare{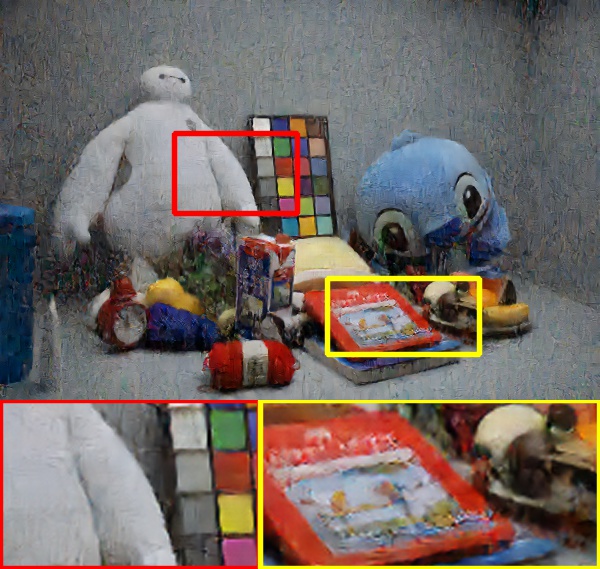}{NERCO~\cite{NeRCo_2023_ICCV}} &
		\panelVisCompare{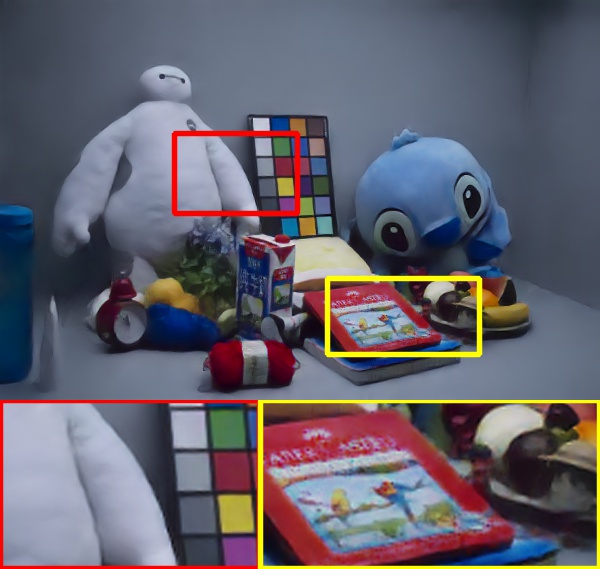}{Ours} &
		\panelVisCompare{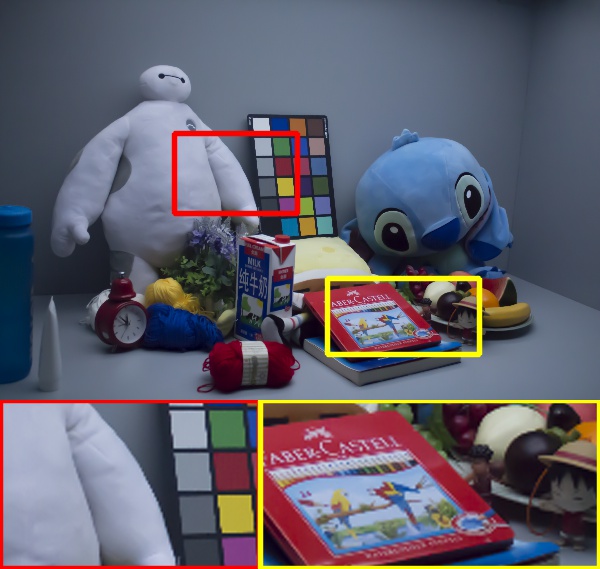}{Ground Truth} \\
	\end{tabular}
	
	\caption{Visual comparison on extremely low-light images under severe noise conditions.}
	\vspace{-1em}
	\label{fig:lolv2_vis_3x6_00321}
\end{figure*}

\begin{figure*}[t]
	\centering
	\setlength{\tabcolsep}{1pt}
	\renewcommand{\arraystretch}{1.0}
	
	\begin{tabular}{cccccc}
		\panelMetric{2/LOLv200009/input.jpg}{Input}{6.22 / 0.1600} &
		\panelMetric{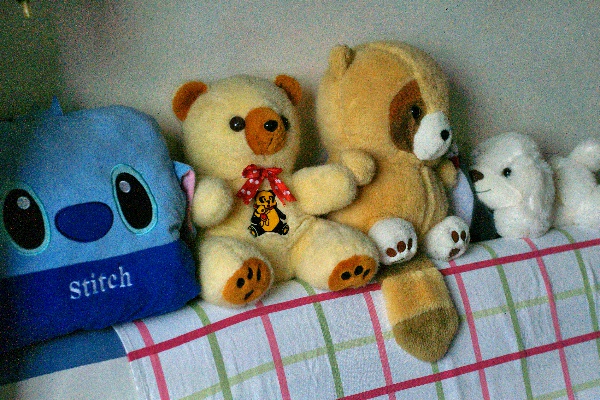}{CoLIE~\cite{chobola2025CoLIE}}{19.54 / 0.404} &
		\panelMetric{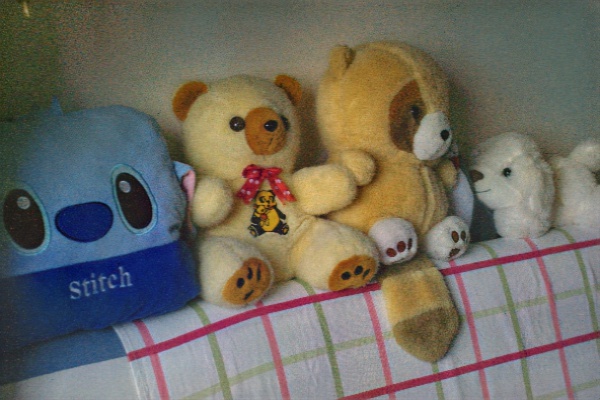}{EnlightenGAN~\cite{jiang2021enlightengan}}{17.54 / 0.728} &
		\panelMetric{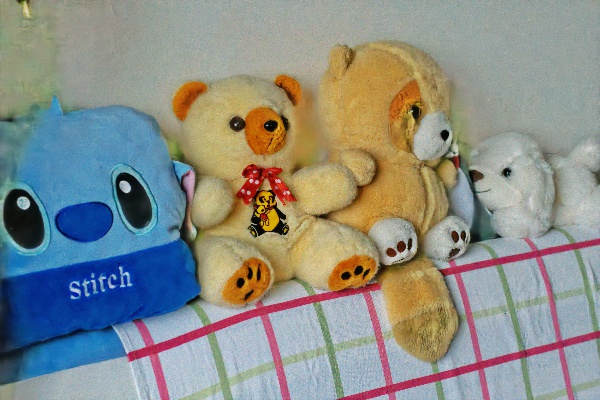}{KinD++~\cite{zhang2021KIND++}}{20.00 / 0.644} &
		\panelMetric{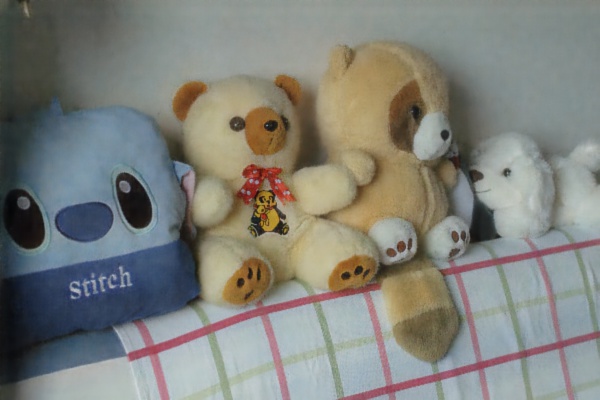}{LightenDiffusion~\cite{lightdiffusion_2024_ECCV}}{24.96 / 0.879} &
		\panelMetric{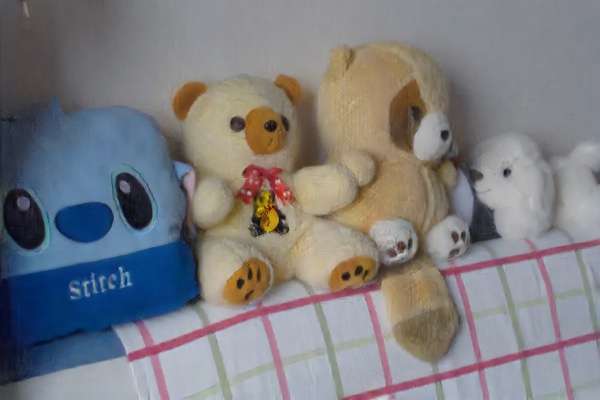}{NeRCo~\cite{NeRCo_2023_ICCV}}{27.88 / 0.875} \\[2pt]
		
		\panelMetric{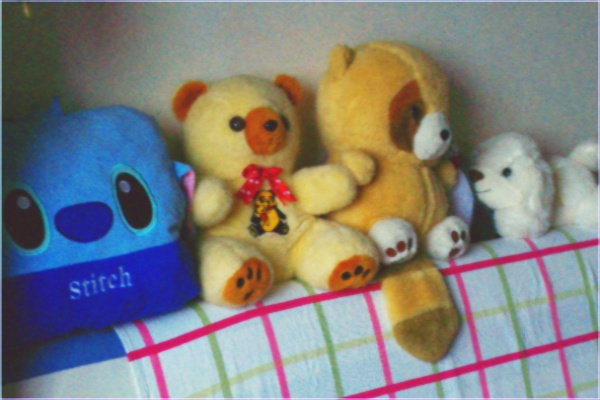}{NoiSER~\cite{zhang2024NoiSER}}{20.62 / 0.744} &
		\panelMetric{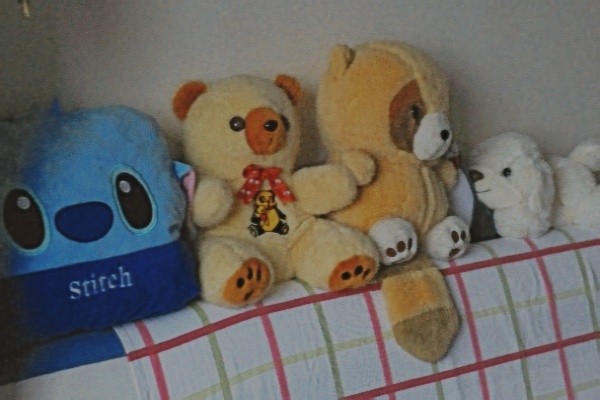}{PairLIE~\cite{fu2023PairLIE}}{17.40 / 0.790} &
		\panelMetric{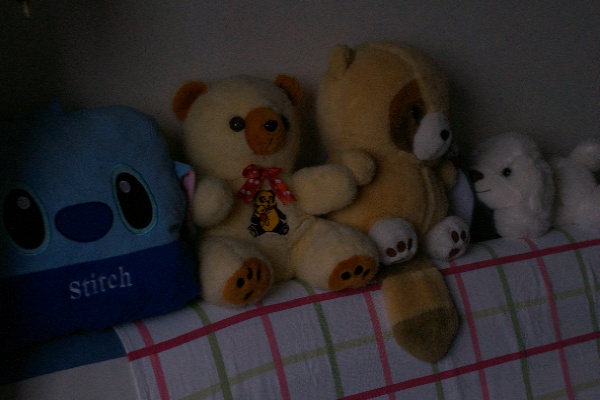}{RUAS~\cite{liu2021ruas}}{8.56 / 0.412} &
		\panelMetric{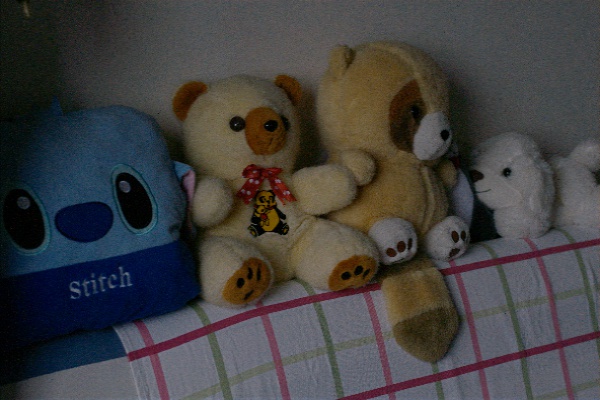}{SCI~\cite{ma2022SCI}}{11.24 / 0.552} &
		\panelMetric{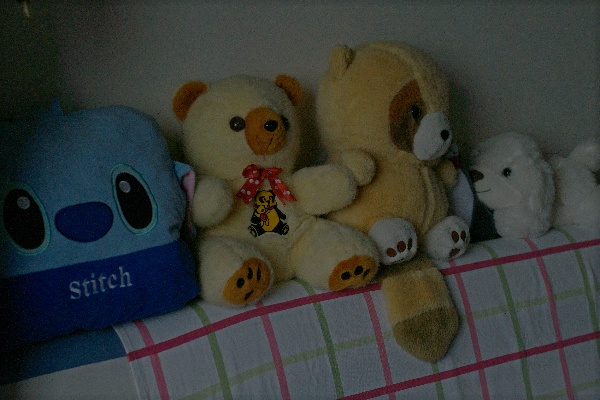}{SCL-LLE~\cite{liang2022SCLLLE}}{10.06 / 0.478} &
		\panelMetric{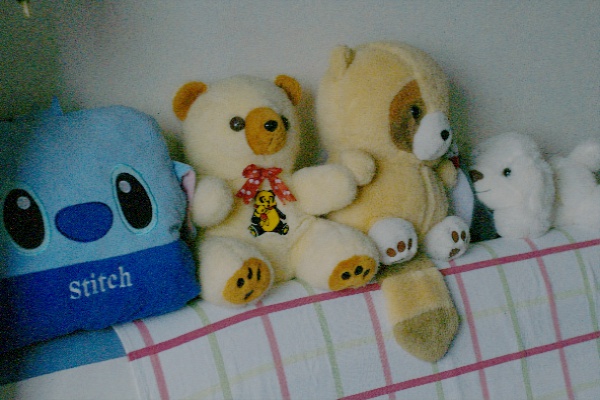}{SCLM~\cite{zhang2023SCLM}}{20.99 / 0.742} \\[2pt]
		
		\panelMetric{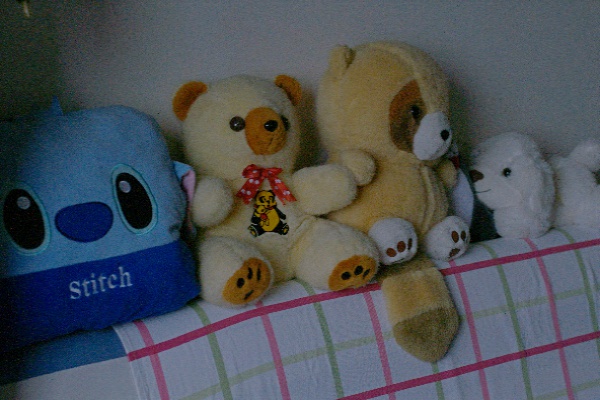}{SGZ~\cite{zheng2022SGZ}}{13.57 / 0.619} &
		\panelMetric{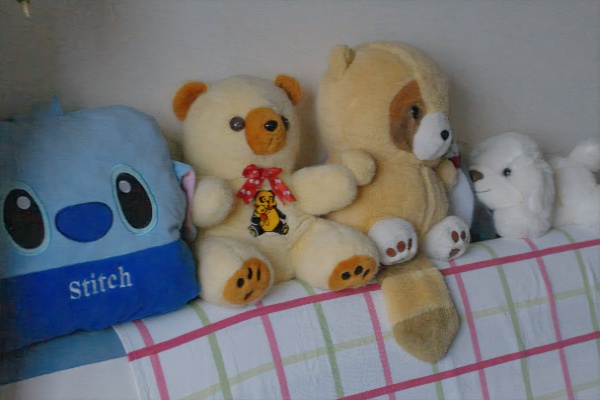}{URetinex~\cite{wu2022uretinex}}{22.20 / 0.871} &
		\panelMetric{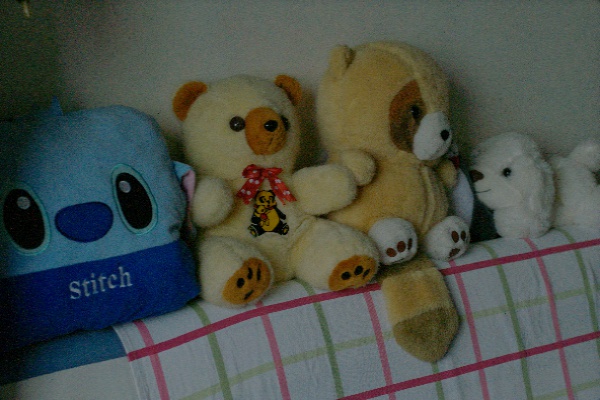}{ZeroDCE~\cite{guo2020zerodce}}{12.77 / 0.608} &
		\panelMetric{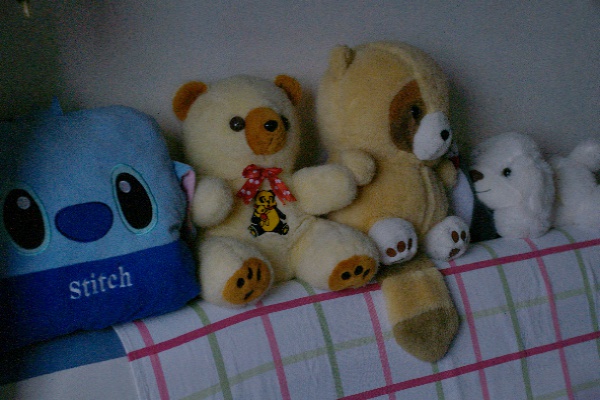}{ZeroDCE++~\cite{li2021zerodce++}}{12.95 / 0.615} &
		\panelMetric{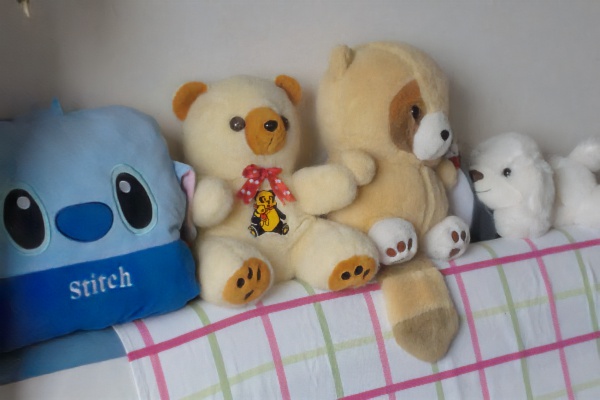}{Ours}{29.7026 / 0.907} &
		\panelMetric{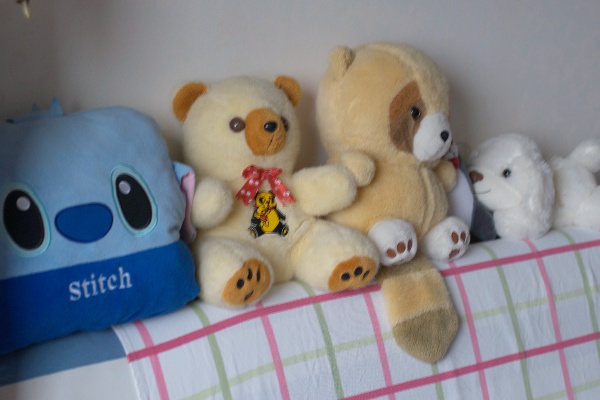}{Ground Truth}{ } 
	\end{tabular}

	\caption{Visual comparisons and PSNR/SSIM results on representative extremely low-light scenes from the LOLv2 dataset.}
	
	\vspace{-1em}
	\label{fig:lol_vis_3x6}
\end{figure*}

\begin{table*}[t]
	\centering
	\caption{Quantitative comparison with state-of-the-art low-light image enhancement methods.}
	\label{tab:quant_comparison_five_datasets}
	\resizebox{\textwidth}{!}{%
		\setlength{\tabcolsep}{4.0pt}
		\renewcommand{\arraystretch}{1.12}
		\begin{tabular}{cccccccccccc}
			\toprule
			\multirow{2}{*}{Methods} & \multirow{2}{*}{Source} & 
			\multicolumn{2}{c}{LOLv2\_Real~\cite{Yang_2021LOLv2}} &
			\multicolumn{2}{c}{LOLv2\_Syn~\cite{Yang_2021LOLv2}} &
			\multicolumn{2}{c}{MIT~\cite{fivekMIT}} &
			\multicolumn{2}{c}{LSRW\_Huawei~\cite{hai2023lsrw}} &
			\multicolumn{2}{c}{LSRW\_Nikon~\cite{hai2023lsrw}} \\
			\cmidrule(lr){3-4}\cmidrule(lr){5-6}\cmidrule(lr){7-8}\cmidrule(lr){9-10}\cmidrule(lr){11-12}
			& & PSNR & SSIM
			& PSNR & SSIM
			& PSNR & SSIM
			& PSNR & SSIM
			& PSNR & SSIM \\
			\midrule
			
			ZeroDCE~\cite{guo2020zerodce}      & CVPR '20  & 18.2654 & 0.6304 & 17.7118 & 0.8123 & 15.8983 & 0.7473 & 16.3458 & 0.4655 & 15.0306 & 0.4145 \\
			
			ZeroDCE++~\cite{li2021zerodce++}   & TPAMI '21 & 18.3704 & 0.5852 & 17.6418 & 0.8133 & 16.2662 & 0.7709 & 16.5368 & 0.4743 & 15.6067 & 0.4293 \\
			RUAS~\cite{liu2021ruas}            & CVPR '21  & 13.9748 & 0.4686 & 13.7650 & 0.6345 & 17.8780 & 0.8254 & 12.5352 & 0.3450 & 13.7949 & 0.3737 \\
			EnlightenGAN~\cite{jiang2021enlightengan} & TIP  '21  & 18.4764 & 0.7383 & 16.6097 & 0.7604 & 14.8903 & 0.7113 & 18.0424 & 0.5307 & 15.3147 & 0.4332 \\
			KIND++~\cite{zhang2021KIND++}      & IJCV '21  & 17.6603 & 0.7609 & 17.4774 & 0.7857 & 14.6434 & 0.6850 & 16.9740 & 0.4120 & 14.7525 & 0.3686 \\
			
			SCI~\cite{ma2022SCI}               & CVPR '22  & 17.2529 & 0.5514 & 16.6949 & 0.7413 & 15.9662 & 0.7765 & 15.0965 & 0.4190 & 15.2775 & 0.3852 \\
			URetinex~\cite{wu2022uretinex}     & CVPR '22  & 20.5652 & 0.8348 & 18.7591 & 0.8333 & 14.3570 & 0.7094 & 19.5233 & 0.5502 & \textcolor{blue}{16.6411} & 0.4506 \\
			SGZ~\cite{zheng2022SGZ}            & WACV '22  & 18.4042 & 0.5782 & 18.3682 & 0.8293 & 14.4830 & 0.7266 & 17.0447 & 0.4717 & 15.7309 & 0.4167 \\
			SCL-LLE~\cite{liang2022SCLLLE}     & AAAI  '22  & 15.4014 & 0.5596 & 15.8461 & 0.7574 & 16.6630 & 0.7887 & 13.4531 & 0.4183 & 13.1146 & 0.3978 \\
			CLIP-LIT~\cite{liang2023CLIPLIT}   & ICCV '23  & 15.1817 & 0.5331 & 16.1900 & 0.7716 & 17.2902 & 0.7890 & 13.5625 & 0.4157 & 13.3665 & 0.3745 \\
			NeRCo~\cite{NeRCo_2023_ICCV}       & ICCV '23  & {25.1720} & 0.7867 & 16.0655 & 0.6733 & 18.3693 & 0.7302 & 18.7568 & \textcolor{blue}{0.5757} & 16.4969 & 0.4795 \\
			PairLIE~\cite{fu2023PairLIE}       & CVPR '23  & 19.8845 & 0.7731 & 19.0744 & 0.7941 & 13.2601 & 0.6764 & 19.0102 & 0.5587 & 15.5299 & 0.4309 \\
			Retinexformer~\cite{cai2023retinexformer} & ICCV '23  & 22.7938 & 0.8387 & \textcolor{red}{25.6697} &\textcolor{blue}{0.9282} & \textcolor{blue}{24.5755} & \textcolor{blue}{0.8858} & 19.1179 & 0.5737 & 16.3568 & \textcolor{blue}{0.4803} \\
			LLformer~\cite{wang2023LLFormer}   & AAAI '23  & \textcolor{blue}{27.7502} & 0.8602 & 17.1631 & 0.7842 & 16.5560 & 0.7706 & 19.2050 & 0.5639 & 16.5100 & 0.4527 \\
			RDHCE~\cite{xia2023RDHCE}          & IJCNN '23 & 16.6360 & 0.4960 & 15.6847 & 0.7284 & 16.9789 & 0.8001 & 15.8857 & 0.4042 & 15.8857 & 0.4042 \\
			
			CoLIE~\cite{chobola2025CoLIE}      & ECCV '24  & 15.0671 & 0.5067 & 14.2961 & 0.6523 & 18.9743 & 0.8430 & 14.7722 & 0.4143 & 12.8747 & 0.3840 \\
			LightenDiffusion~\cite{lightdiffusion_2024_ECCV}
			& ECCV '24  & 22.7419 & 0.8522 & 21.6587 & 0.8651 & 19.7757 & 0.8047 & \textcolor{blue}{19.8080} & 0.5667 & 16.3920 & 0.4628 \\
			SCLM~\cite{zhang2023SCLM}          & CVPR '24  & 16.9267 & 0.5812 & 17.0866 & 0.7326 & 10.0005 & 0.5032 & 18.2258 & 0.4760 & 14.0507 & 0.4108 \\
			ZeroIG~\cite{shi2024zeroIG}        & CVPR '24  & 15.6593 & 0.4442 & 14.8271 & 0.7209 & 8.3333  & 0.5916 & 16.8910 & 0.4163 & 12.6374 & 0.3435 \\

			NoiSER~\cite{zhang2024NoiSER}      & TPAMI '25 & 16.4733 & 0.6776 & 14.9785 & 0.6355 & 16.8542 & 0.7090 & 15.7901 & 0.5299 & 15.5844 & 0.4450 \\
			HVI~\cite{yan2025hvi}              & CVPR '25  & 23.9040 & \textcolor{blue}{0.8663} & 25.1294 & \textcolor{red}{0.9388} & 24.4587 & 0.8769 & 19.1124 &  0.5522 & 16.1812 & 0.4536 \\
			
			\textbf{Ours (ICD)}              & \textbf{--} 
			& \textcolor{red}{29.7100} & \textcolor{red}{0.8895} & \textcolor{blue}{25.2272} & {0.9078} & \textcolor{red}{25.4061} & \textcolor{red}{0.9184} & \textcolor{red}{21.2399} & \textcolor{red}{0.6330} & \textcolor{red}{17.8946} & \textcolor{red}{0.5190} \\
			
			\bottomrule
		\end{tabular}
	}
	\vspace{-1em}
\end{table*}

\begin{figure*}[t]
	\centering
	\setlength{\tabcolsep}{1pt}
	\renewcommand{\arraystretch}{1.0}
	
\begin{tabular}{cccccc}
	\panelMetric{2/LOLv200268/input.jpg}{Input}{7.2046 / 0.0748} &
	\panelMetric{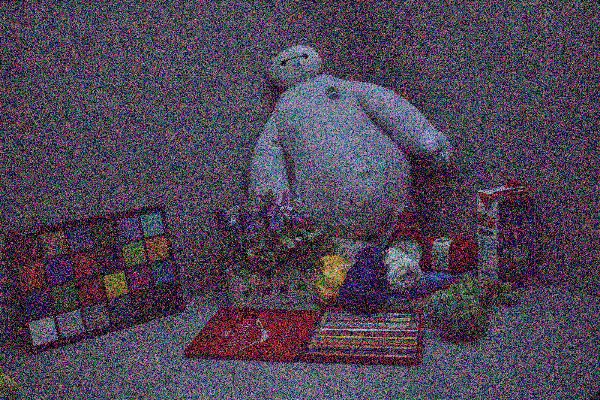}{CoLIE~\cite{chobola2025CoLIE}}{10.5778 / 0.0466} &
	\panelMetric{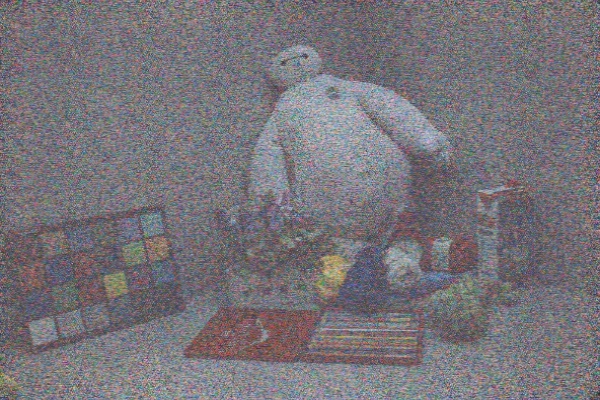}{EnlightenGAN~\cite{jiang2021enlightengan}}{17.0795 / 0.1934} &
	\panelMetric{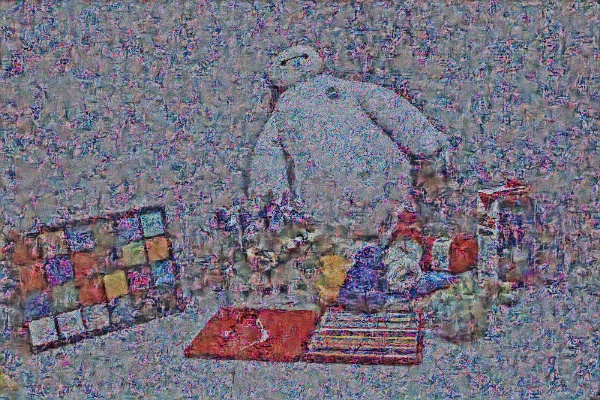}{KinD++~\cite{zhang2021KIND++}}{15.6463 / 0.1414} &
	\panelMetric{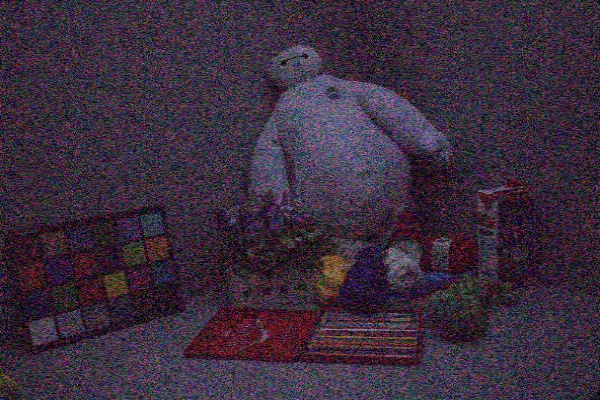}{RUAS~\cite{liu2021ruas}}{10.1043 / 0.0895} &
	\panelMetric{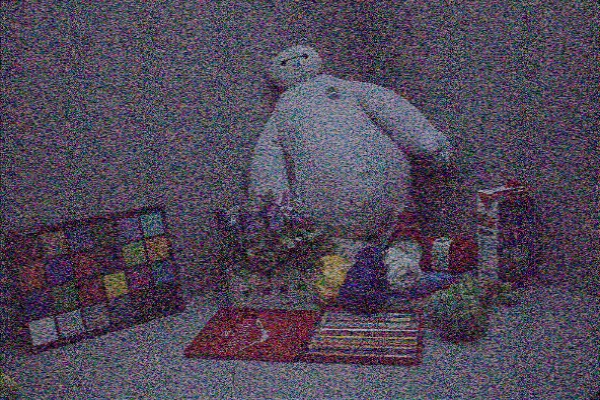}{SCI~\cite{ma2022SCI}}{12.3322 / 0.0852} \\[2pt]
	
	\panelMetric{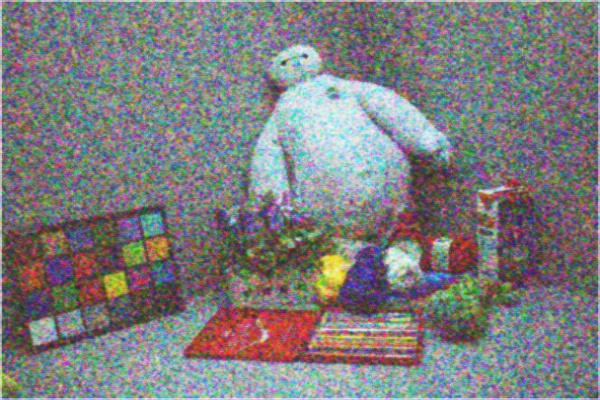}{NoiSER~\cite{zhang2024NoiSER}}{17.3824 / 0.1790} &
	\panelMetric{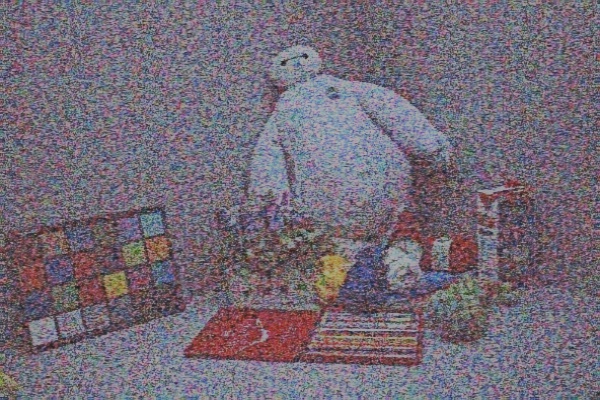}{PairLIE~\cite{fu2023PairLIE}}{17.0117 / 0.1832} &
	\panelMetric{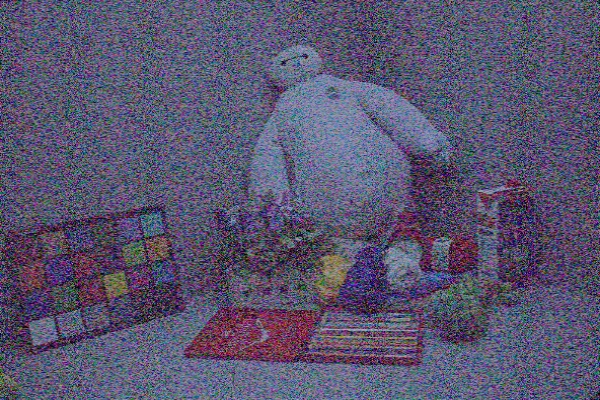}{SGZ~\cite{zheng2022SGZ}}{13.4907 / 0.0808} &
	\panelMetric{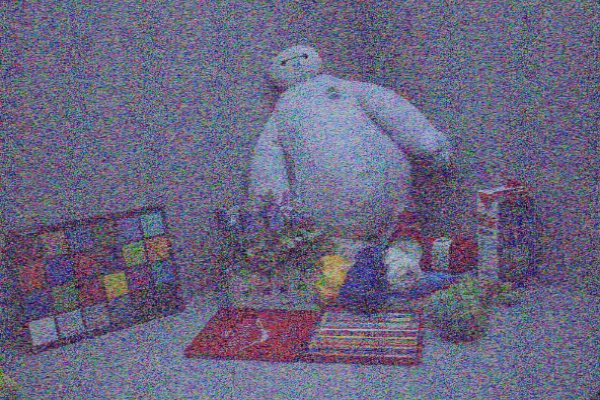}{URetinex~\cite{wu2022uretinex}}{15.4031 / 0.0976} &
	\panelMetric{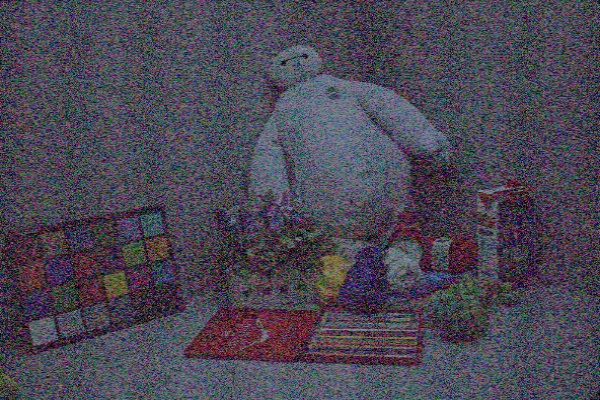}{ZeroDCE~\cite{guo2020zerodce}}{12.0855 / 0.0895} &
	\panelMetric{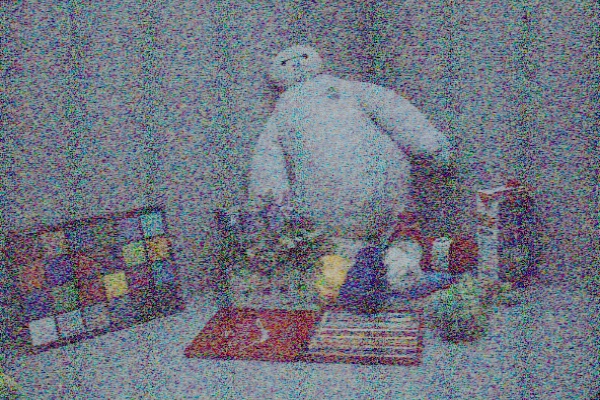}{SCLM~\cite{zhang2023SCLM}}{16.2703 / 0.1248} \\[2pt]
	
	\panelMetric{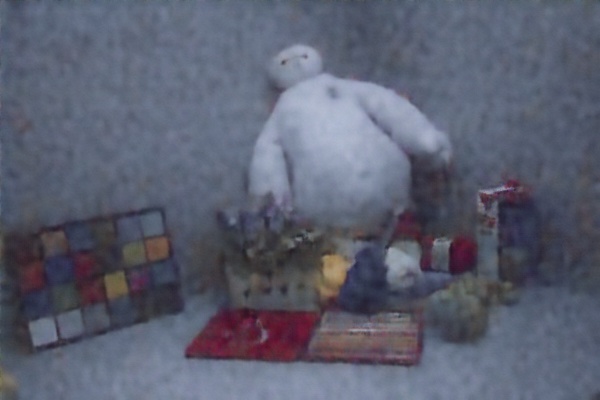}{LightenDiffusion~\cite{lightdiffusion_2024_ECCV}}{18.3162 / 0.713} &
	\panelMetric{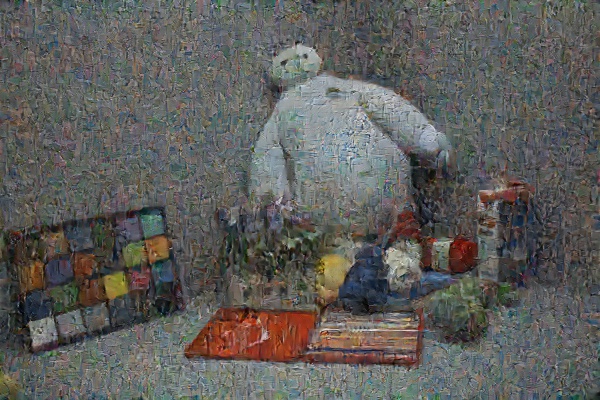}{NeRCo~\cite{NeRCo_2023_ICCV}}{15.7500 / 0.2656} &
	\panelMetric{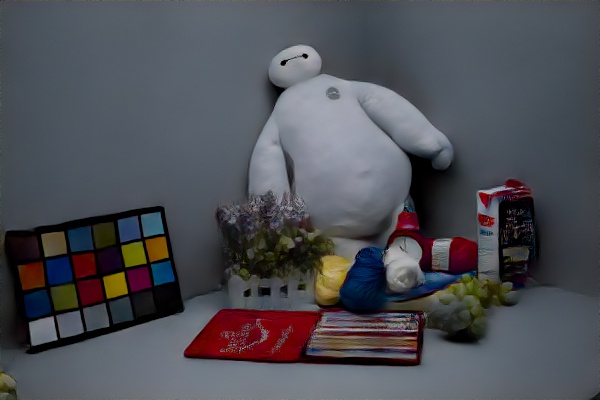}{HVI~\cite{yan2025hvi}}{14.9075 / 0.7349} &
	\panelMetric{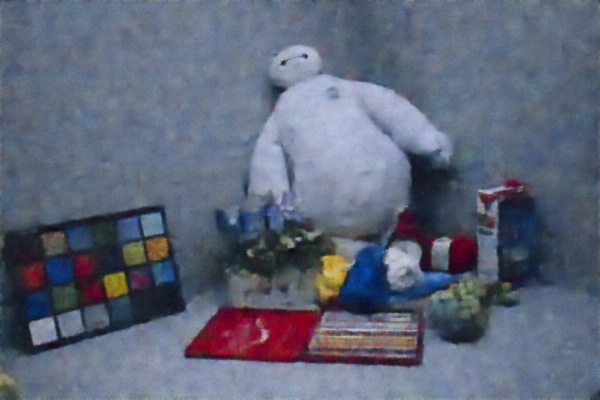}{Retinexformer~\cite{cai2023retinexformer}}{20.6726 / 0.6859} &
	\panelMetric{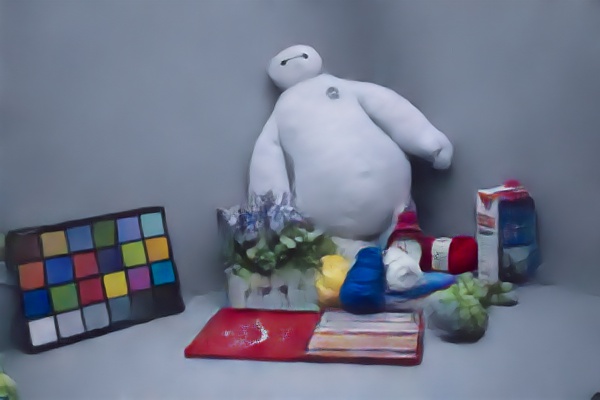}{Ours}{24.7443 / 0.8324} &
	\panelMetric{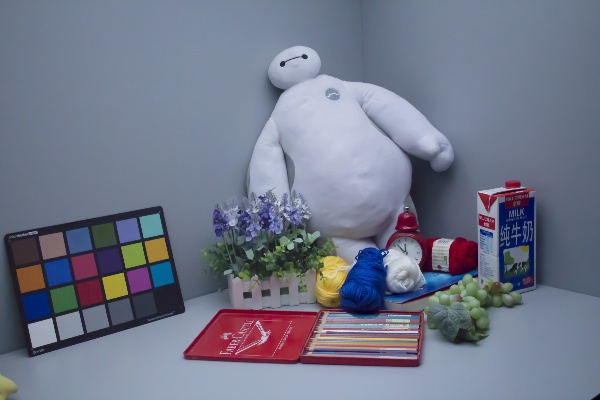}{Ground Truth}{ }
\end{tabular}
	\caption{Visual comparisons and PSNR/SSIM results on representative extremely low-light scenes from LOLv2.}
	\label{fig:lol_vis_3x6}
	\vspace{-1em}
\end{figure*}

\begin{figure*}[p] 
	\centering
	\setlength{\tabcolsep}{1pt}
	\renewcommand{\arraystretch}{1.0}
	
	\begin{tabular}{cccccc}
		\panelMetric{2/LOLv200011/input.jpg}{Input}{5.6819 / 0.0463} &
		\panelMetric{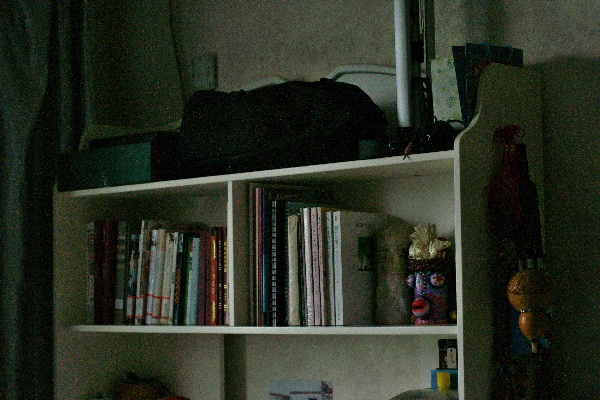}{CoLIE}{8.8660 / 0.2666} &
		\panelMetric{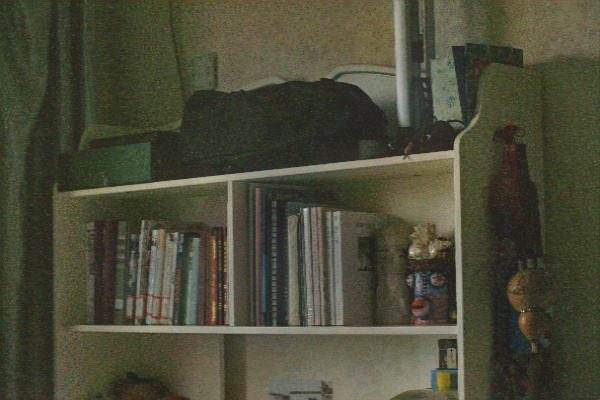}{EnlightenGAN}{11.3639 / 0.5538} &
		\panelMetric{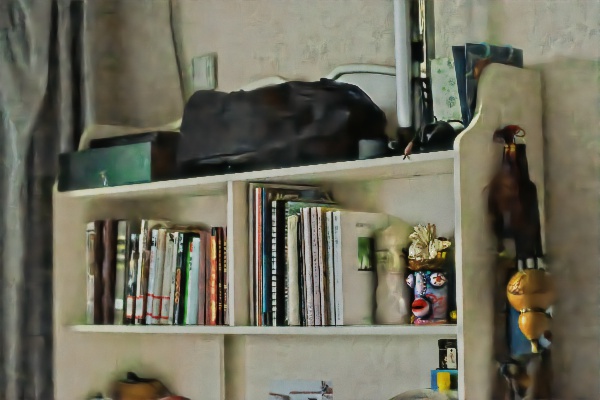}{KIND++}{17.3284 / 0.7992} &
		\panelMetric{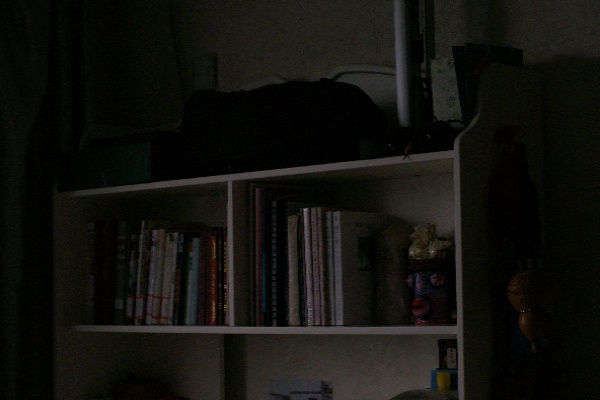}{RUAS}{6.6219 / 0.1427} &
		\panelMetric{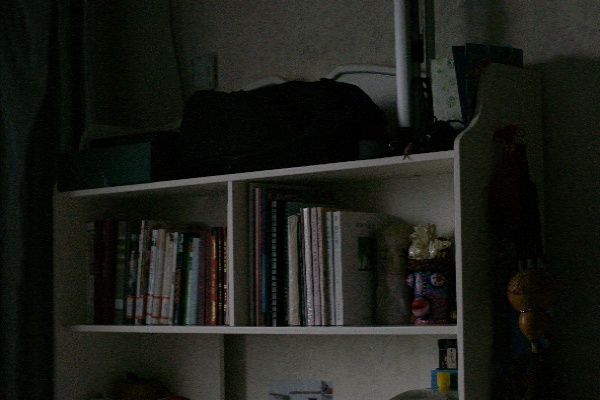}{SCI}{7.6414 / 0.2325} \\[2pt]
		
		\panelMetric{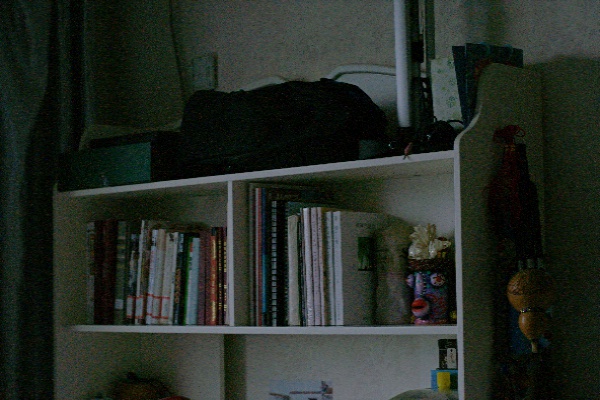}{SGZ}{8.7580 / 0.3094} &
		\panelMetric{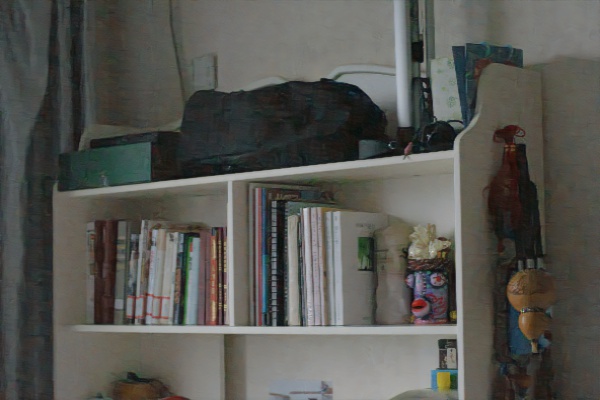}{URetinex-Net}{14.6529 / 0.7844} &
		\panelMetric{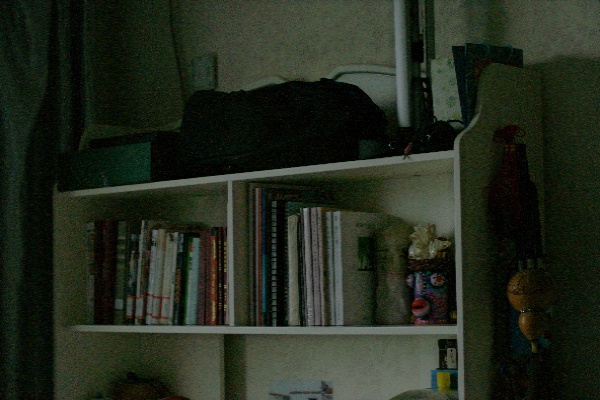}{ZeroDCE}{8.3964 / 0.2871} &
		\panelMetric{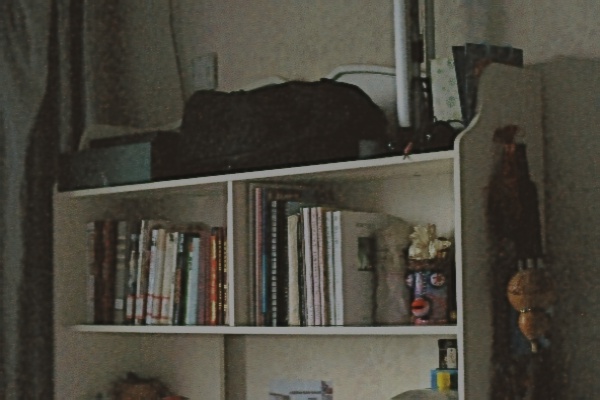}{PairLIE}{11.7224 / 0.6441} &
		\panelMetric{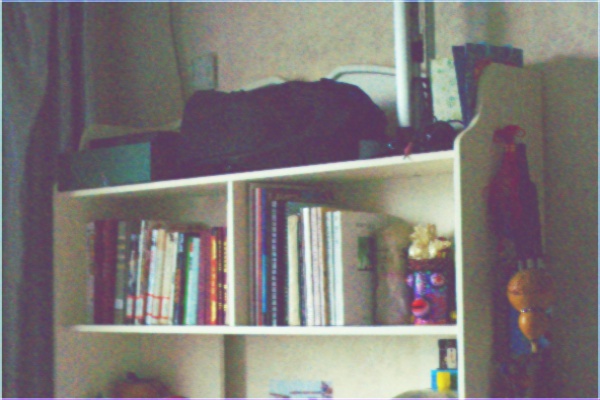}{NoiSER}{19.8051 / 0.6733} &
		\panelMetric{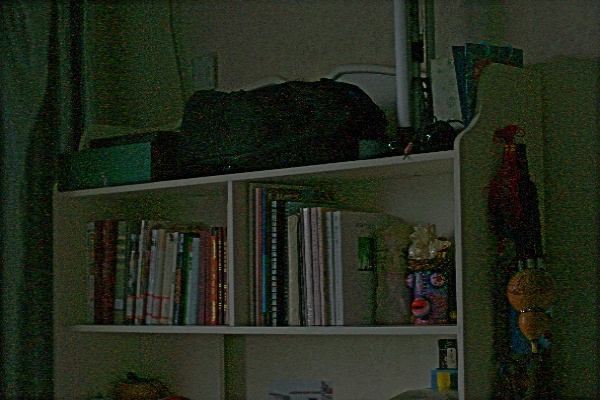}{CLIP-LIT}{7.9845 / 0.2642} \\[2pt]
		
		\panelMetric{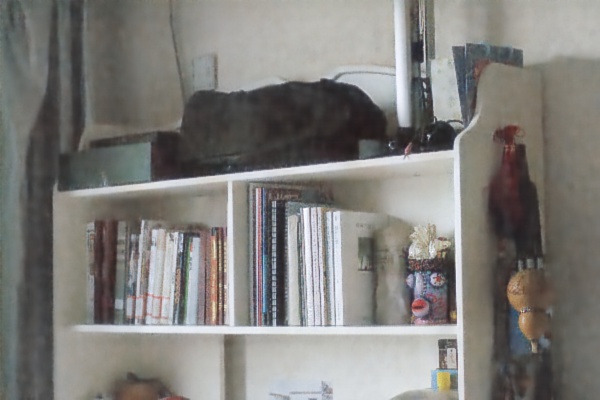}{LightenDiffusion}{26.4180 / 0.8676} &
		\panelMetric{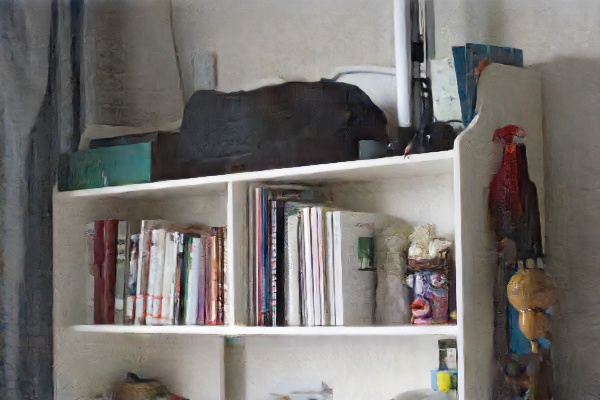}{NeRCo}{22.8367 / 0.8077} &
		\panelMetric{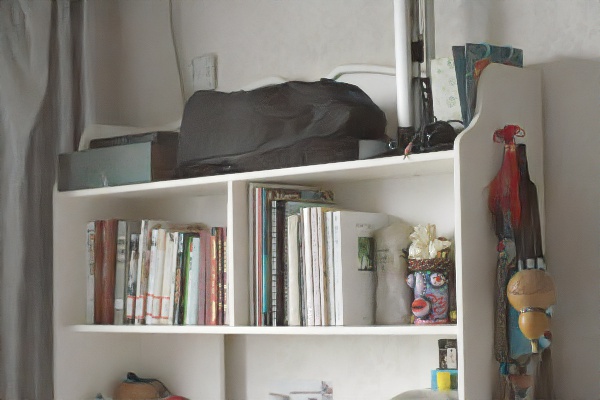}{HVI}{27.1427 / 0.8765} &
		\panelMetric{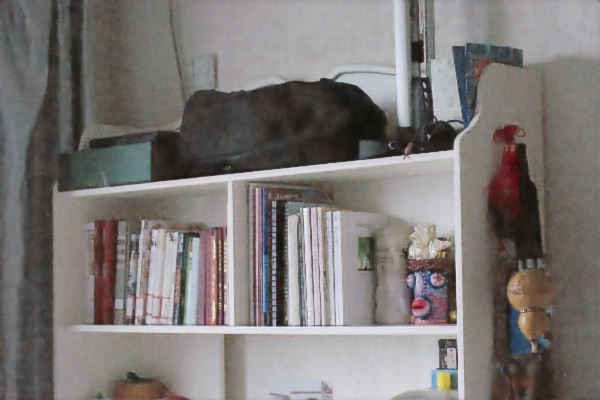}{Retinexformer}{25.4302 / 0.8614} &
		\panelMetric{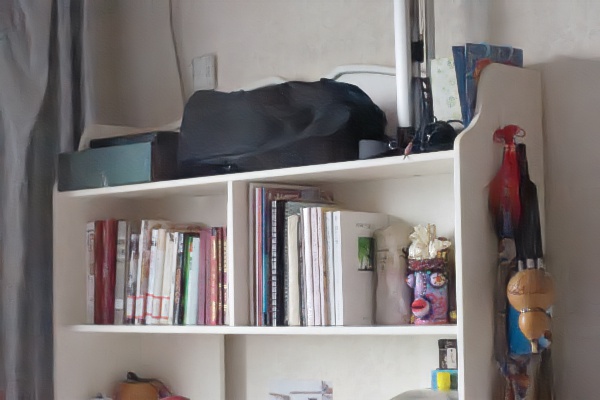}{ICD}{29.0350 / 0.9022} &
		\panelMetric{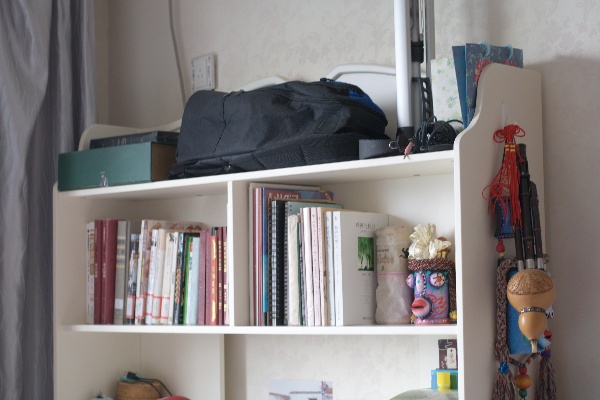}{Ground Truth}{ } \\[-1pt]
	\end{tabular}
	\vspace{-0.2em} 
	\caption{Visual comparisons and per-image PSNR/SSIM results on a high dynamic range scene.}
	\label{fig:lol_vis_3x6_200036}
	
	\vspace{-0.6em} 
	
	\begin{tabular}{cccccc}
		\panelMetric{2/LOLv200042/input.jpg}{Input}{5.3130 / 0.0863} &
		\panelMetric{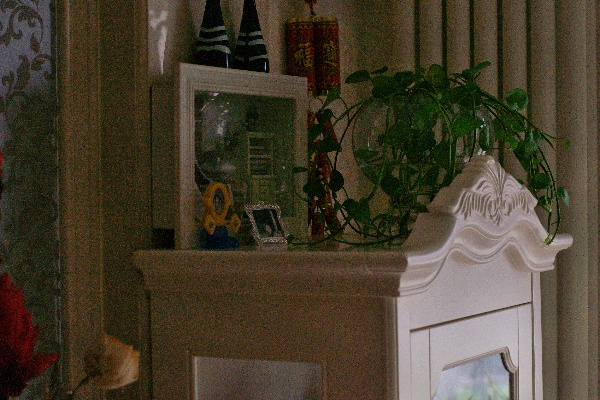}{CoLIE}{9.4905 / 0.3354} &
		\panelMetric{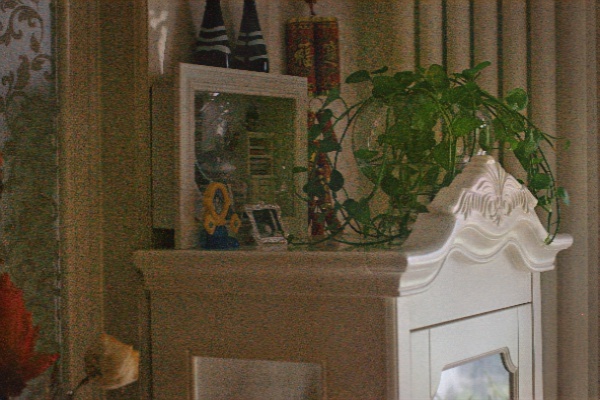}{EnlightenGAN}{11.3766 / 0.5625} &
		\panelMetric{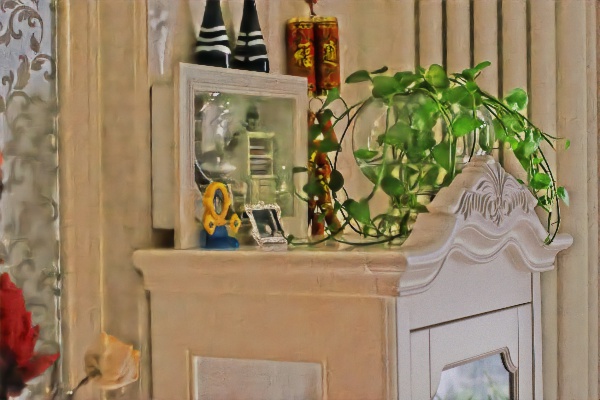}{KIND++}{16.8081 / 0.7921} &
		\panelMetric{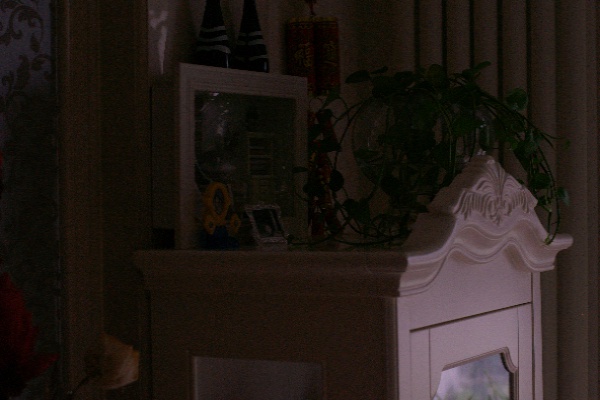}{RUAS}{6.7977 / 0.2338} &
		\panelMetric{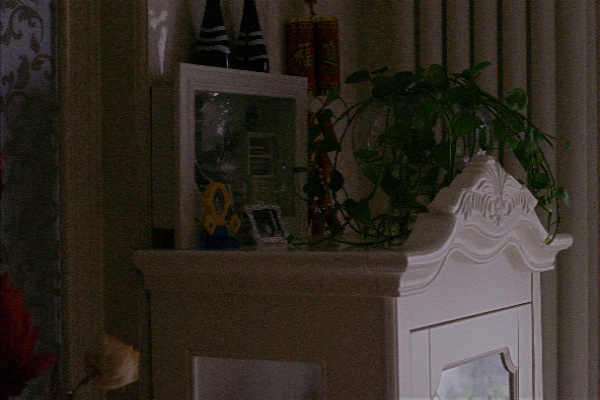}{SCI}{8.0405 / 0.3424} \\[2pt]
		
		\panelMetric{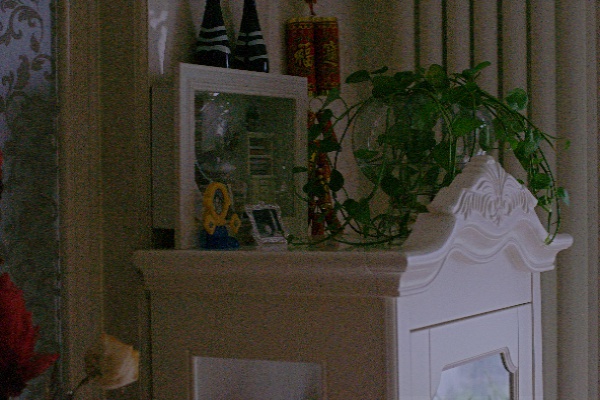}{SGZ}{9.4187 / 0.4272} &
		\panelMetric{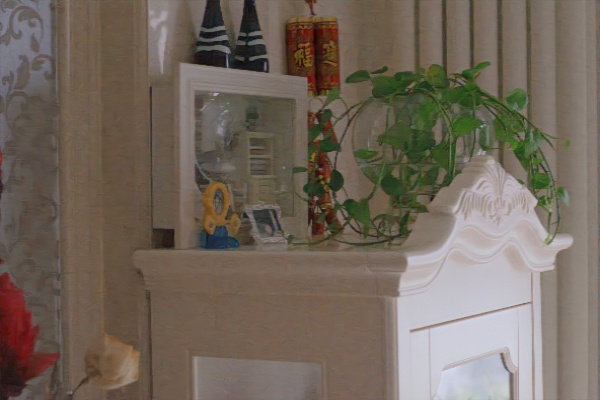}{URetinex-Net}{15.8685 / 0.8359} &
		\panelMetric{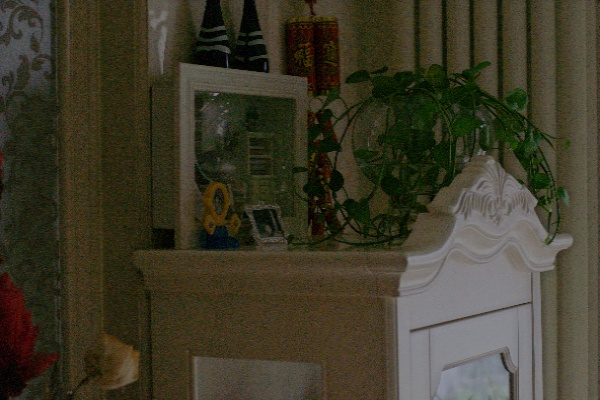}{ZeroDCE}{8.9652 / 0.4065} &
		\panelMetric{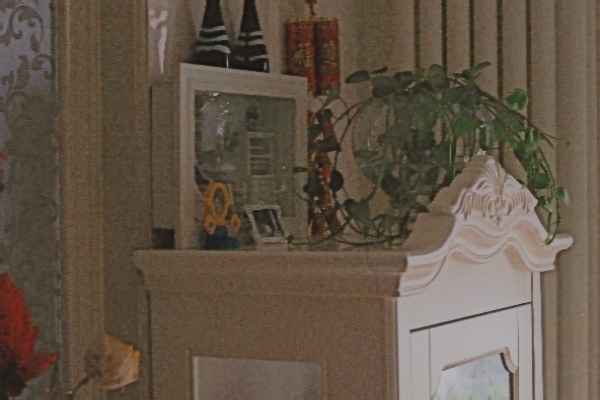}{PairLIE}{12.4111 / 0.7115} &
		\panelMetric{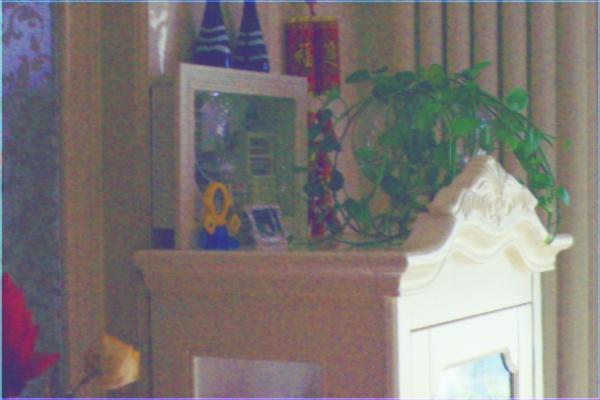}{NoiSER}{18.5992 / 0.6779} &
		\panelMetric{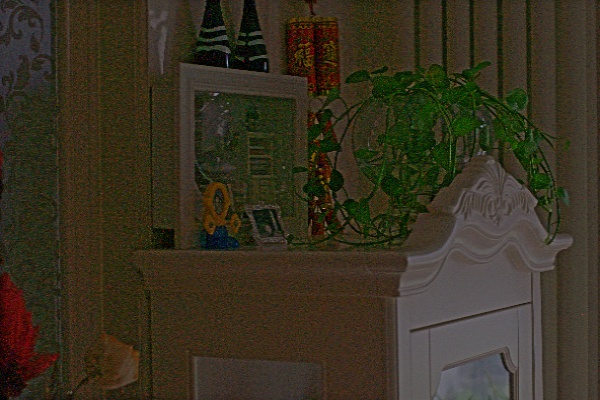}{CLIP-LIT}{7.9865 / 0.3497} \\[2pt]
		
		\panelMetric{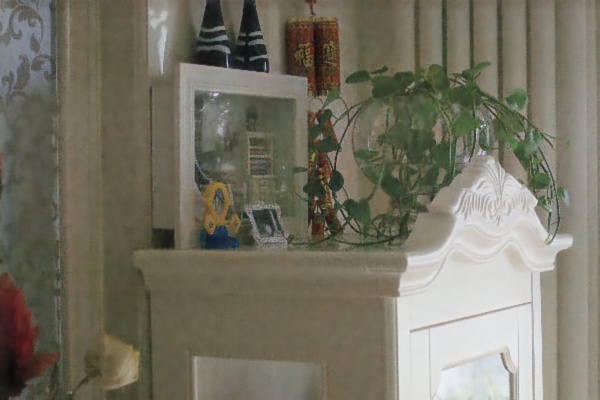}{LightenDiffusion}{17.4306 / 0.8561} &
		\panelMetric{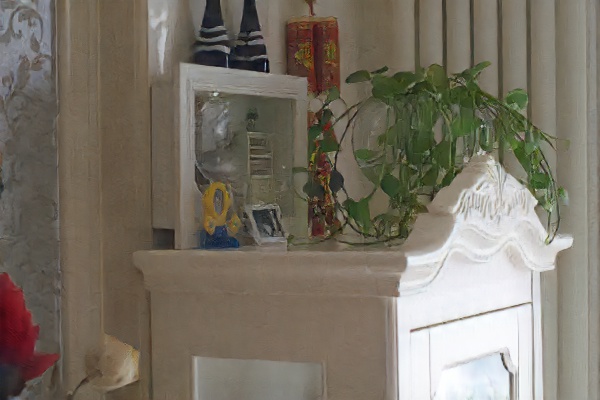}{NeRCo}{18.0689 / 0.8179} &
		\panelMetric{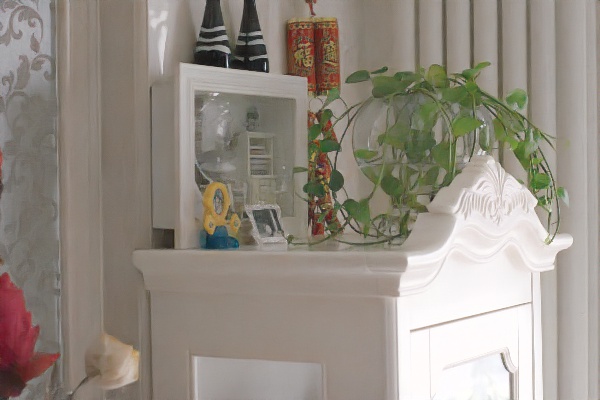}{HVI}{30.5060 / 0.9028} &
		\panelMetric{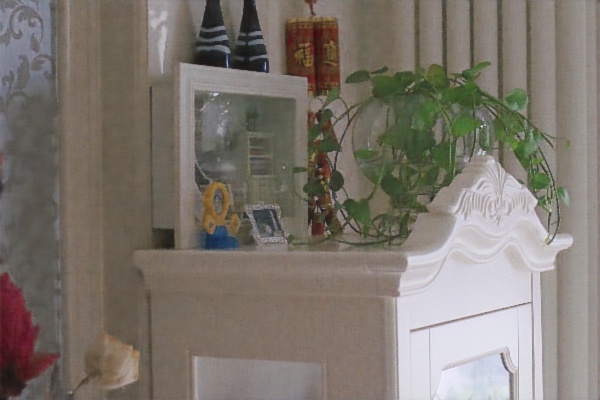}{Retinexformer}{19.6386 / 0.8707} &
		\panelMetric{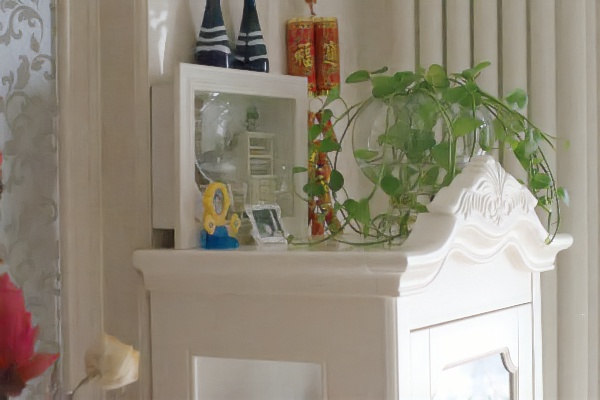}{ICD}{30.8231 / 0.9193} &
		\panelMetric{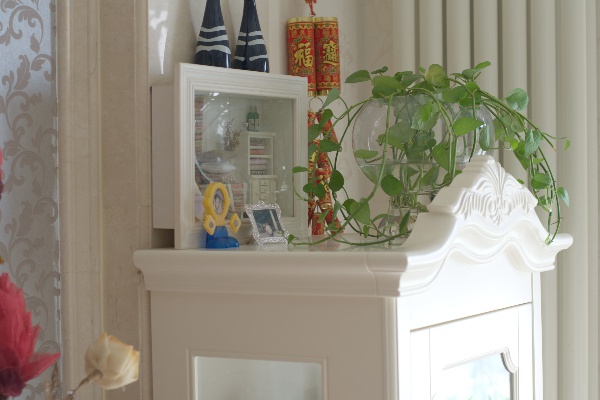}{Ground Truth}{ } \\[-1pt]
	\end{tabular}
	\vspace{-0.2em} 
	\caption{Visual comparisons and per-image PSNR/SSIM results on a low-texture scene from LOLv2.}
	\label{fig:lol_vis_3x6_200146}
	
	\vspace{-0.6em} 
	
	\begin{tabular}{cccccc}
		\panelMetric{2/LOLv200513/input.jpg}{Input}{7.71 / 0.089} &
		\panelMetric{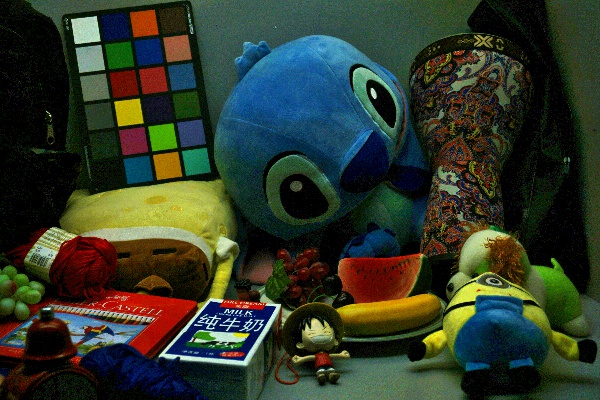}{CoLIE}{13.61 / 0.457} &
		\panelMetric{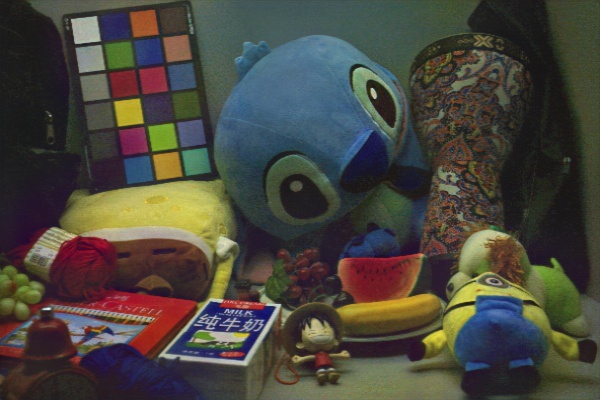}{EnlightenGAN}{15.51 / 0.576} &
		\panelMetric{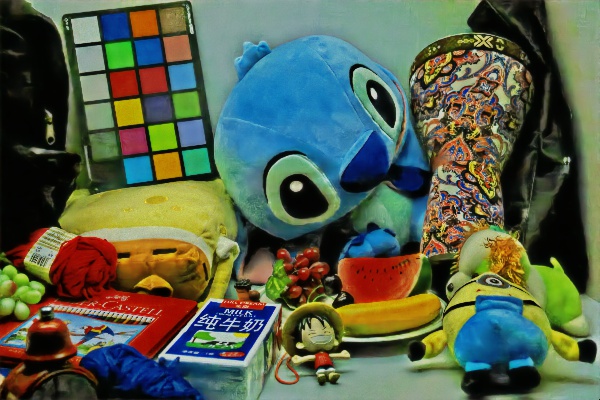}{KIND++}{17.24 / 0.704} &
		\panelMetric{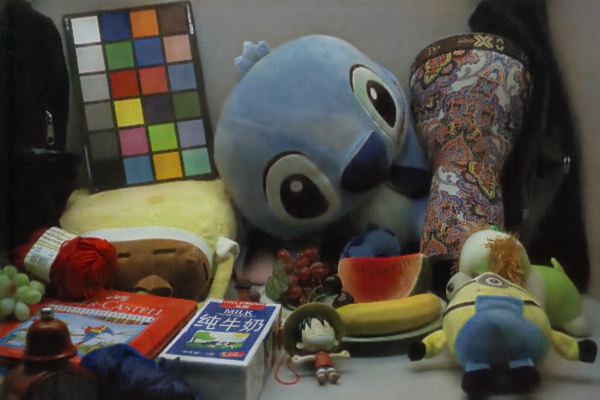}{LightenDiffusion}{17.80 / 0.671} &
		\panelMetric{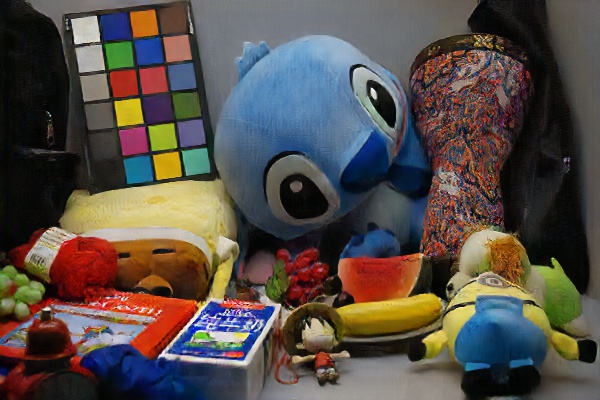}{NeRCo}{19.31 / 0.666} \\[2pt]
		
		\panelMetric{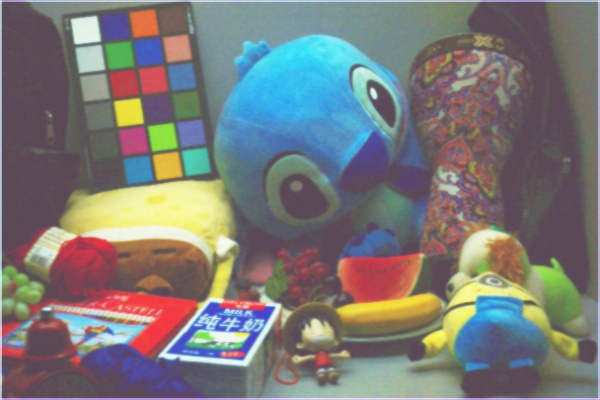}{NoiSER}{13.86 / 0.499} &
		\panelMetric{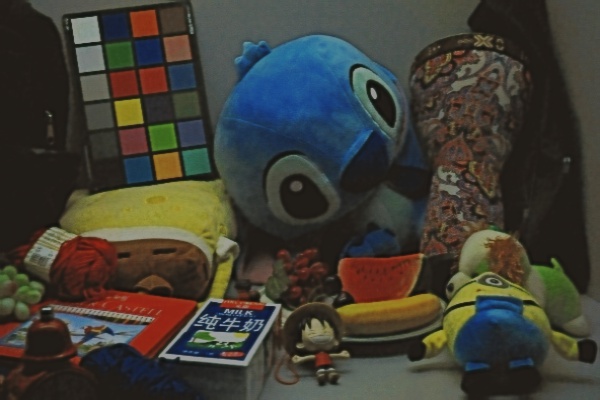}{PairLIE}{14.00 / 0.546} &
		\panelMetric{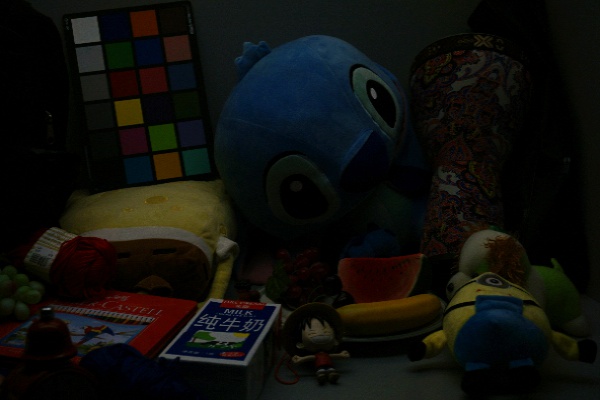}{RUAS}{8.87 / 0.184} &
		\panelMetric{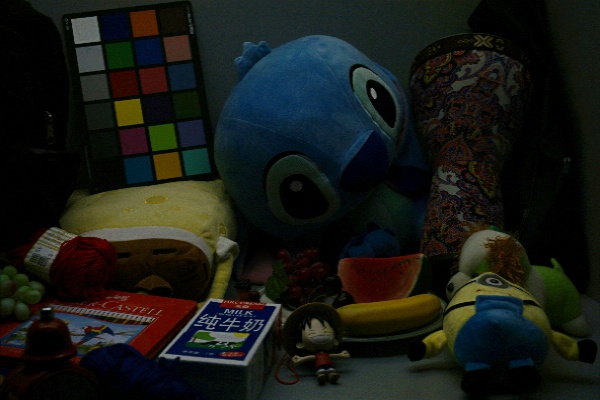}{SCI}{10.04 / 0.289} &
		\panelMetric{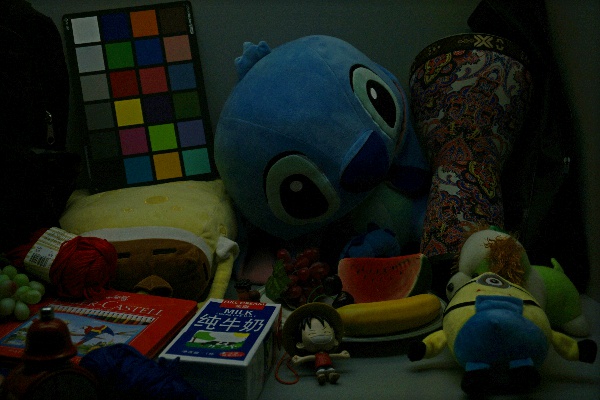}{SCLLLE}{9.96 / 0.295} &
		\panelMetric{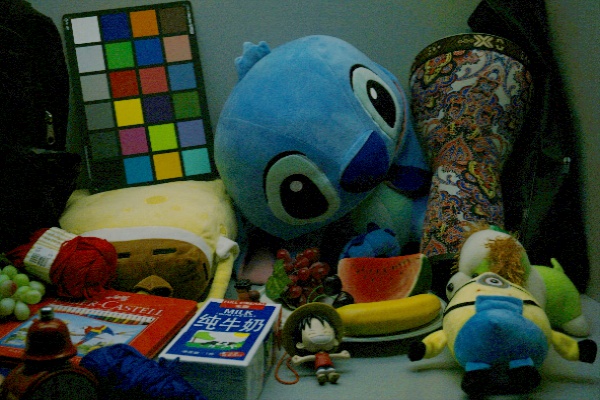}{SCLM}{14.69 / 0.569} \\[2pt]
		
		\panelMetric{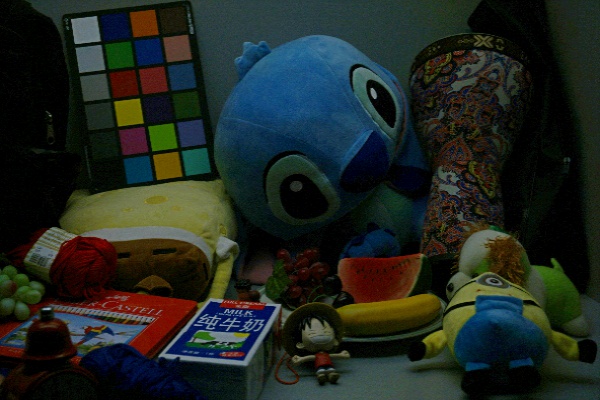}{SGZ}{11.46 / 0.412} &
		\panelMetric{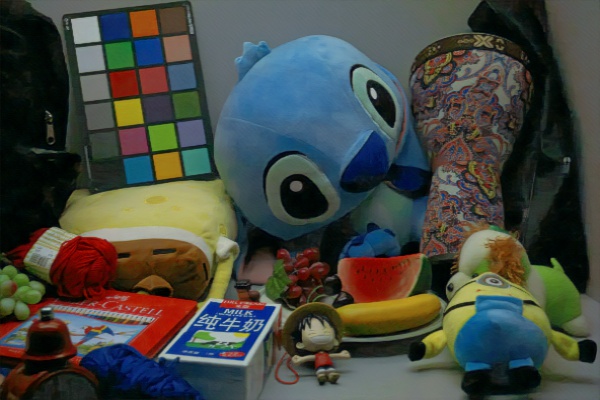}{URetinex-Net}{16.92 / 0.710} &
		\panelMetric{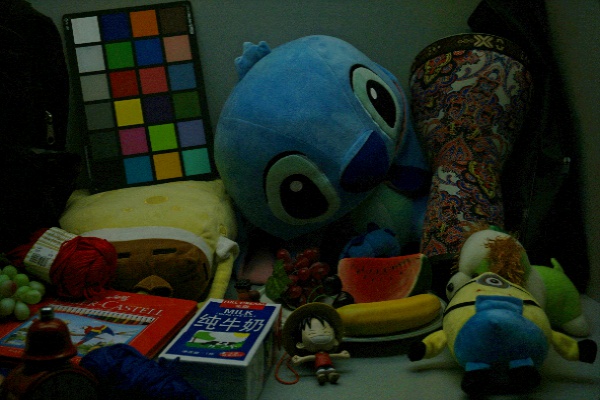}{ZeroDCE}{11.12 / 0.383} &
		\panelMetric{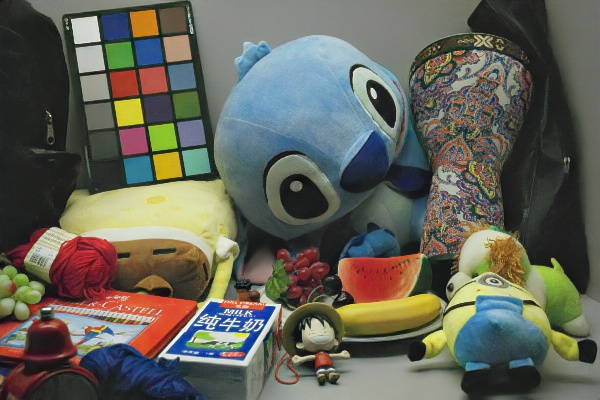}{HVI}{20.41 / 0.70} &
		\panelMetric{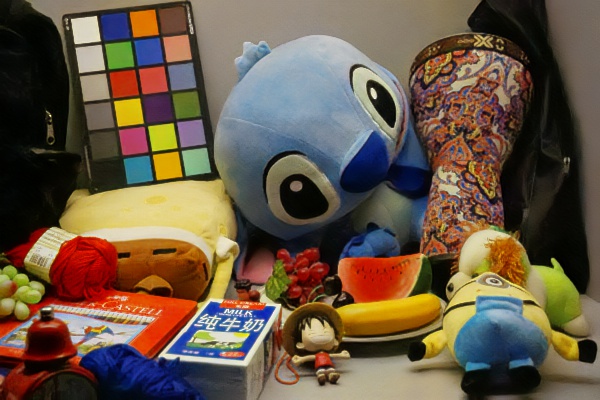}{Ours}{{24.57} / {0.814}} &
		\panelMetric{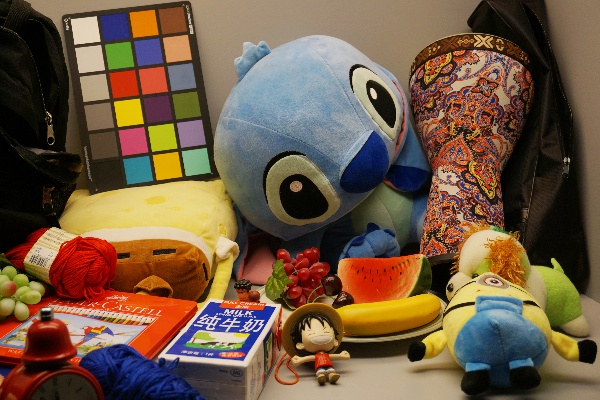}{Ground Truth}{ } \\[-1pt]
	\end{tabular}
	\vspace{-0.2em} 
	\caption{Visual comparisons and per-image PSNR/SSIM results on an extremely low-light color image.}
	\label{fig:lol_vis_3x6_200258}
	
\end{figure*}

\subsection{Quantitative Comparison}
\label{sec:quantitative_comparison}

We compare the proposed intensity--chromaticity decoupled network (ICD) with representative low-light image enhancement methods on LOLv2 (Real and Synthetic), LSRW (Huawei and Nikon), and MIT-Adobe FiveK.
The compared methods cover several representative paradigms. 
We include curve-based and zero-reference methods, such as Zero-DCE~\cite{guo2020zerodce} and Zero-DCE++~\cite{li2021zerodce++}; architecture search and adversarial learning methods, including RUAS~\cite{liu2021ruas} and EnlightenGAN~\cite{jiang2021enlightengan}; and decomposition-based or physics-guided methods, including KinD++~\cite{zhang2021KIND++}, SCI~\cite{ma2022SCI}, URetinex-Net~\cite{wu2022uretinex}, and SGZ~\cite{zheng2022SGZ}. 
We further compare with recent transformer-, diffusion-, and representation-based methods, including CLIP-LIT~\cite{liang2023CLIPLIT}, NeRCo~\cite{NeRCo_2023_ICCV}, PairLIE~\cite{fu2023PairLIE}, Retinexformer~\cite{cai2023retinexformer}, LLFormer~\cite{wang2023LLFormer}, RDHCE~\cite{xia2023RDHCE}, CoLIE~\cite{chobola2025CoLIE}, LightenDiffusion~\cite{lightdiffusion_2024_ECCV}, SCLM~\cite{zhang2023SCLM}, ZeroIG~\cite{shi2024zeroIG}, SCL-LLE~\cite{liang2022SCLLLE}, NoiSER~\cite{zhang2024NoiSER}, and HVI~\cite{yan2025hvi}.

Table~\ref{tab:quant_comparison_five_datasets} reports quantitative comparisons on the five paired datasets. 
ICD achieves the highest PSNR and SSIM on LOLv2-Real, MIT-Adobe FiveK, LSRW-Huawei, and LSRW-Nikon. 
On LOLv2-Syn, ICD achieves the highest PSNR, while its SSIM is slightly lower than HVI~\cite{yan2025hvi}.
On datasets with relatively regular degradation, including LOLv2-Real, LOLv2-Syn, and MIT-Adobe FiveK, recent models such as Retinexformer~\cite{cai2023retinexformer}, LLFormer~\cite{wang2023LLFormer}, and HVI~\cite{yan2025hvi} already achieve strong performance. 
ICD achieves comparable or improved results on these datasets.
On LSRW-Huawei and LSRW-Nikon, which contain real-captured images with more complex degradations, most methods show lower performance. 
ICD outperforms Retinexformer~\cite{cai2023retinexformer}, LLFormer~\cite{wang2023LLFormer}, HVI~\cite{yan2025hvi}, and LightenDiffusion~\cite{lightdiffusion_2024_ECCV} in both PSNR and SSIM, suggesting improved robustness under complex noise and severe underexposure.

%
%
%
%

\subsection{Visual Comparison}
\label{sec:visual_comparison}

Qualitative comparisons on representative indoor scenes from the LOLv2 Real-captured dataset are shown in Fig.~\ref{fig:lolv2_vis_3x6_00321}--Fig.~\ref{fig:lol_vis_3x6_200258}. 
The examples cover diverse scenarios, including close-up objects, clothing racks, hanging decorations, and high-dynamic-range scenes. 
PSNR and SSIM are reported below each image.
Different methods show distinct behaviors in brightness restoration, noise suppression, and color preservation. 
Some methods increase global brightness but amplify noise in dark regions, often introducing color bias or local over-exposure. 
Others preserve structural content but under-enhance extremely dark regions. 
ICD yields stable brightness recovery, fewer chromatic artifacts, and clearer structural details.

Under more challenging conditions with severe underexposure and higher noise levels, performance differences become more pronounced. 
Some methods remain under-enhanced, resulting in low visibility, while others amplify weak signals and introduce noise patterns or local artifacts. 
Transformer-based methods improve global brightness but may produce blurred textures or local inconsistencies in high-frequency regions. 
ICD maintains brightness recovery while preserving structural details and limiting chromatic noise.
Across all examples in Fig.~\ref{fig:lolv2_vis_3x6_00321}--Fig.~\ref{fig:lol_vis_3x6_200258}, ICD consistently improves visibility, suppresses noise, and preserves structural details, which is consistent with the quantitative results in Section~\ref{sec:quantitative_comparison}.

\section{Ablation and Analysis}
\label{sec:analysis}

This section provides additional analysis of the proposed method, including quality--efficiency trade-offs, model complexity, loss ablation, decoupled mapping strategies, noise behavior, and downstream face detection on DarkFace.

\subsection{Ablation Study}
\label{sec:ablation_analysis}

Table~\ref{tab:ablation_results} reports the ablation results on key loss terms and model components. 
Removing any supervision term degrades performance, indicating that the different losses play complementary roles in optimizing reconstruction fidelity, intensity recovery, and chromatic consistency. 
Among the loss-ablation variants, removing the RGB reconstruction loss \(\mathcal{L}_{\mathrm{rgb}}\) leads to the largest performance drop, with PSNR decreasing from 29.7100 dB to 24.2315 dB on LOLv2-Real and from 26.1978 dB to 22.8437 dB on LOLv1. 
This suggests that RGB-domain supervision remains necessary for preserving overall reconstruction quality after constrained inverse reconstruction. 
Removing the intensity loss \(\mathcal{L}_{I}\) weakens the alignment of the intensity envelope and affects exposure recovery, while removing the chromaticity loss \(\mathcal{L}_{C}\) degrades inter-channel consistency and color stability.

We further compare enhancement strategies under different representations. 
As shown in Table~\ref{tab:ablation_results}, direct enhancement in the RGB domain performs worse than the full ICD model on both datasets. 
The visual results in Fig.~\ref{fig:lol00766_vis_2x4} show that RGB-domain enhancement tends to introduce color bias and chromatic noise, especially in dark regions and fine-detail areas. 
In contrast, the decoupled representation assigns intensity adjustment to a shared intensity component and models channel-relative variations in the chromaticity component, thereby reducing channel-wise inconsistency and suppressing chromatic artifacts during enhancement.

We also evaluate the choice of the intensity baseline in the decoupled space. 
In the \(I_{\min}\)-based variant, the pixel-wise minimum channel response is used as the intensity baseline:
\begin{equation}
	I_{\min}(\mathbf{I})(x)
	=
	\min_{c\in\{R,G,B\}} I_c(x).
	\label{eq:imin_def}
\end{equation}
For each channel \(c\in\{R,G,B\}\), the corresponding chromaticity component is defined as
\begin{equation}
	C^{\min}_c(\mathbf{I})(x)
	=
	\log\!\big(I_c(x)+\epsilon\big)
	-
	\log\!\big(I_{\min}(\mathbf{I})(x)+\epsilon\big).
	\label{eq:cmin_def}
\end{equation}
The \(I_{\min}\)-based forward mapping is then written as
\begin{equation}
	\begin{aligned}
		\mathbf{I}(x)
		&\longmapsto
		\big(I_{\min}(\mathbf{I})(x),\mathbf{C}^{\min}(\mathbf{I})(x)\big),\\
		\mathbf{C}^{\min}(\mathbf{I})(x)
		&=
		\big(C^{\min}_R(\mathbf{I})(x),
		C^{\min}_G(\mathbf{I})(x),
		C^{\min}_B(\mathbf{I})(x)\big)^\top .
	\end{aligned}
	\label{eq:imin_forward}
\end{equation}
For each channel \(c\in\{R,G,B\}\), the inverse transformation is
\begin{equation}
	I_c(x)
	=
	\big(I_{\min}(\mathbf{I})(x)+\epsilon\big)
	\exp\!\big(C^{\min}_c(\mathbf{I})(x)\big)
	-\epsilon.
	\label{eq:imin_inverse}
\end{equation}
This setting corresponds to the \(I_{\min}\)-based ICD variant in Table~\ref{tab:ablation_results}.

\newcommand{\panelFour}[2]{%
	\shortstack[t]{%
		\includegraphics[width=0.185\textwidth]{#1}\\[0.3ex]
		\small #2%
	}%
}

\begin{figure*}[t]
	\centering
	\setlength{\tabcolsep}{1pt}
	\renewcommand{\arraystretch}{1.0}
	
	\begin{tabular}{ccccc}
		\panelFour{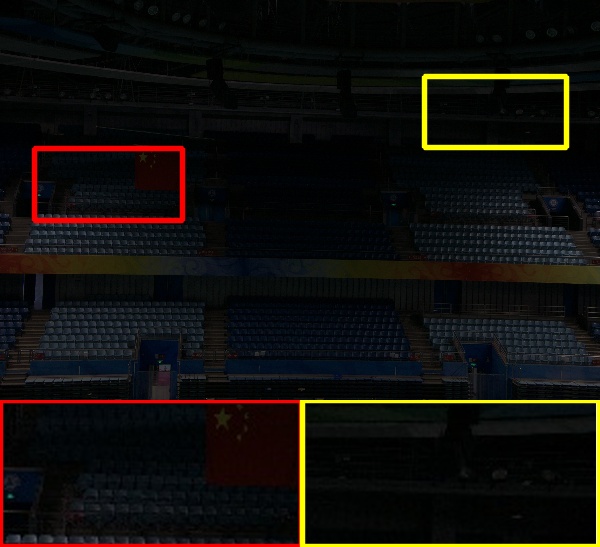}{Input} &
		\panelFour{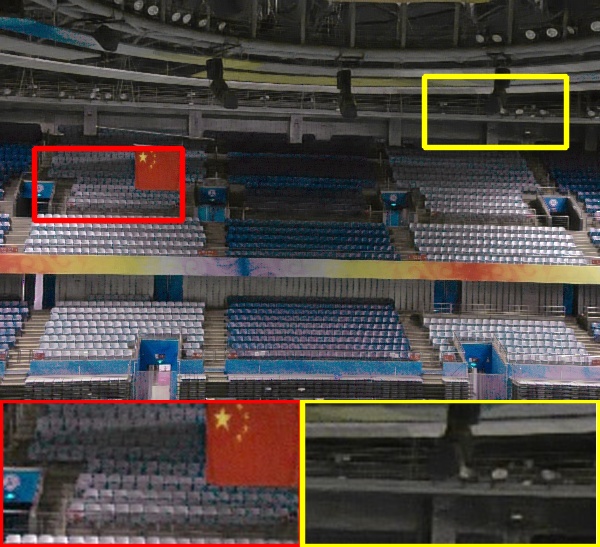}{RGB-only} &
		\panelFour{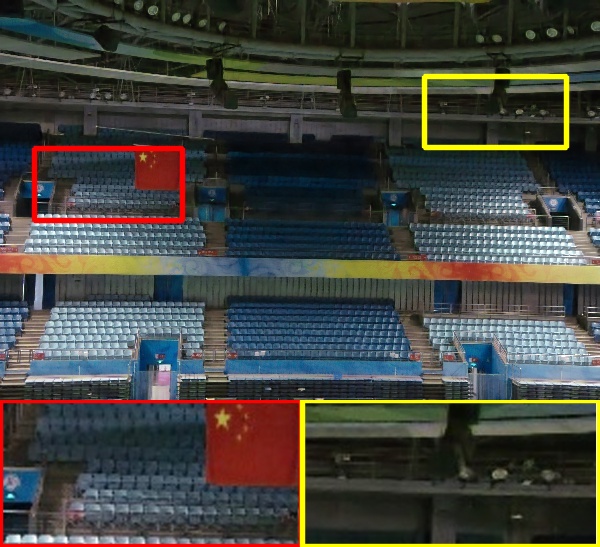}{w/o $\mathcal{L}_{\mathrm{rgb}}$} &
		\panelFour{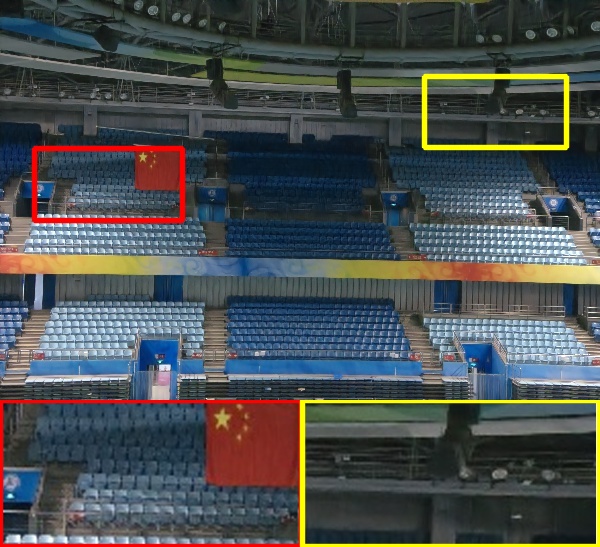}{$I_{\min}$ as intensity} &
		\panelFour{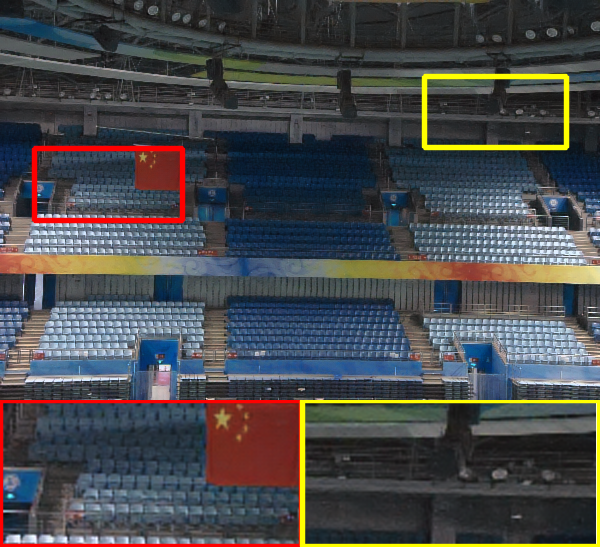}{$I_{\mathrm{ave}}$ as intensity}\\[2pt]
		
		
		\panelFour{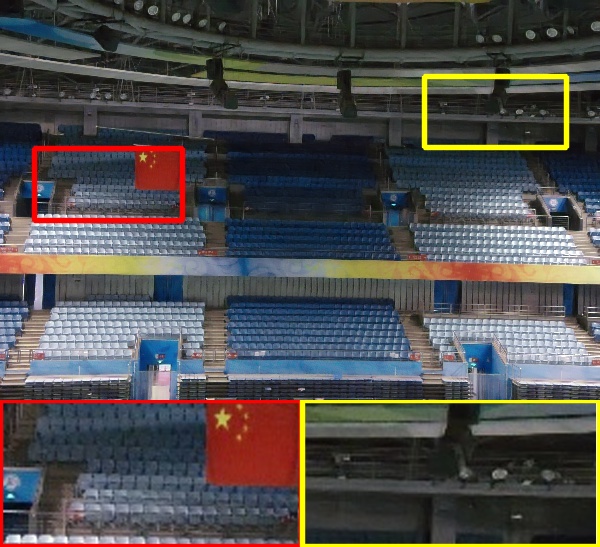}{w/o $\mathcal{L}_{I}$} &
		\panelFour{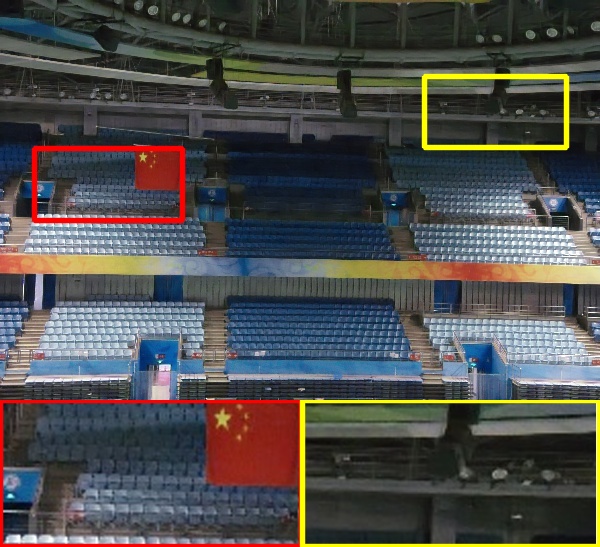}{w/o $\mathcal{L}_{C}$} &
		\panelFour{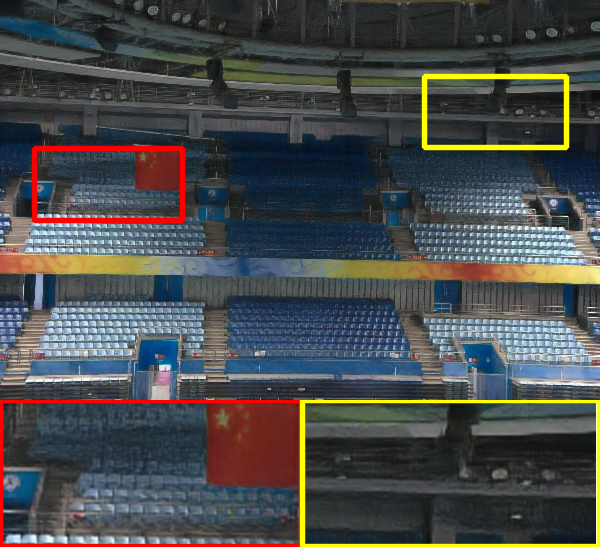}{w/o chroma gate} &
		\panelFour{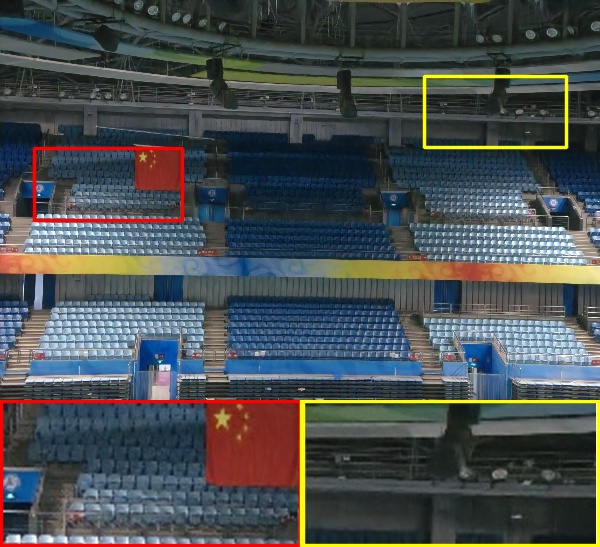}{Full model}  &
		\panelFour{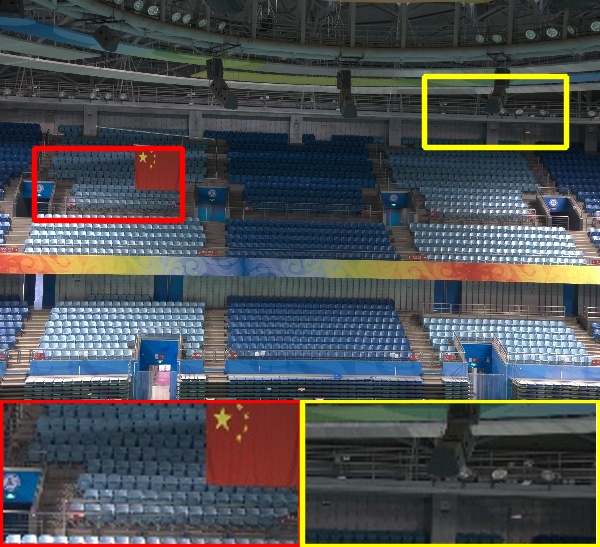}{Ground truth} \\
	\end{tabular}
	
	\caption{Visual comparison of different ablation settings on LOLv2.}
	\vspace{-1em}
	\label{fig:lol00766_vis_2x4}
\end{figure*}

\begin{table*}[t]
	\centering
	\small
	\caption[Ablation results]{Ablation results on the LOLv2-Real~\cite{Yang_2021LOLv2} and LOLv1~\cite{Chen2018LOLv1} test sets.}
	\label{tab:ablation_results}
	\setlength{\tabcolsep}{5.2pt}
	\renewcommand{\arraystretch}{1.15}
	\resizebox{\textwidth}{!}{%
		\begin{tabular}{clccccccccc}
			\toprule
			& \multirow{2}{*}{Method}
			& \multirow{2}{*}{Gate}
			& \multicolumn{3}{c}{LOLv2-Real~\cite{Yang_2021LOLv2}}
			& \multicolumn{3}{c}{LOLv1~\cite{Chen2018LOLv1}}
			& \multirow{2}{*}{Params (M)$\downarrow$}
			& \multirow{2}{*}{FLOPs (G)$\downarrow$} \\
			\cmidrule(lr){4-6}\cmidrule(lr){7-9}
			&
			&
			& PSNR$\uparrow$ & SSIM$\uparrow$ & LPIPS$\downarrow$
			& PSNR$\uparrow$ & SSIM$\uparrow$ & LPIPS$\downarrow$
			& & \\
			\midrule
			1 & w/o $\mathcal{L}_{\mathrm{rgb}}$ & \checkmark
			& 24.2315 & 0.8654 & 0.1382
			& 22.8437 & 0.8214 & 0.1652
			& 1.10 &5.57 \\
			
			2 & w/o $\mathcal{L}_{I}$ & \checkmark
			& 26.6113 & 0.8764 & 0.1284
			& 23.4123 & 0.8355 & 0.1476
			& 1.10 & 5.57 \\
			
			3 & w/o $\mathcal{L}_{C}$ & \checkmark
			& 26.4135 & 0.8712 & 0.1326
			& 23.0741 & 0.8292 & 0.1531
			& 1.10 & 5.57 \\
			\midrule
			
			4 & RGB-only enhancement & \checkmark
			& 25.8257 & 0.8223 & 0.1942
			& 21.9254 & 0.7953 & 0.2115
			& 1.10 & \textbf{4.16}\\
			
			5 & ICD with $I_{\min}$ & \checkmark
			& 28.3112 & 0.8521 & 0.1322
			& 25.7780 & 0.8354 & 0.1356
			& 1.10 & 5.57 \\
			
			6 & ICD with $I_{\mathrm{ave}}$ & \checkmark
			& 28.8192 & 0.8864 & 0.0962
			& 25.1451 & 0.8533 & 0.1181
			& 1.10 & 5.57 \\
			
			7 & w/o chroma gate & $\times$
			& 23.5105 & 0.8495 & 0.2608
			& 20.0605 & 0.7984& 0.3154
			& \textbf{1.08} & 5.56 \\
			\midrule
			
			& \textbf{Full model} & \checkmark
			& \textbf{29.7100} & \textbf{ 0.8895} & \textbf{0.0867 }
			& \textbf{26.1978 } & \textbf{0.8567} & \textbf{0.1061}
			& {1.10} & {5.57} \\
			\bottomrule
		\end{tabular}%
	}
\end{table*}
The \(I_{\min}\)-based variant still achieves competitive enhancement performance, but its results are consistently lower than those of the full model. 
This indicates that decomposing enhancement into an intensity scale and relative chromatic ratios is effective, while the choice of the intensity baseline affects reconstruction quality. 
More generally, the purpose of intensity--chromaticity decoupling is to define an intensity baseline and its corresponding chromaticity variables. 
From this perspective, \(I_{\min}\), \(I_{\mathrm{ave}}\), and \(I_{\max}\) can all provide feasible decoupled parameterizations. 
However, compared with \(I_{\min}\) and \(I_{\mathrm{ave}}\), using \(I_{\max}\) as the intensity envelope better preserves the non-positive chromaticity constraint and provides a more stable upper envelope for RGB reconstruction.

The \(I_{\mathrm{ave}}\)-based variant also obtains strong performance, especially on LOLv2-Real, where it achieves 28.8192 dB PSNR and 0.8864 SSIM. 
However, it still underperforms the full \(I_{\max}\)-based model on both datasets. 
This suggests that the average intensity can provide a useful global brightness descriptor, but it does not impose the same channel-wise upper-envelope constraint as \(I_{\max}\). 
Therefore, it is less effective in stabilizing the chromaticity representation and constrained RGB reconstruction.

Finally, removing the chroma gate causes a substantial performance degradation. 
The PSNR drops to 23.5105 dB on LOLv2-Real and 20.0605 dB on LOLv1, while LPIPS increases markedly to 0.2608 and 0.3154, respectively. 
This indicates that the chroma gate plays an important role in regulating chromaticity updates and suppressing unstable color variations in dark regions. 
Together with the visual results in Fig.~\ref{fig:lol00766_vis_2x4}, these results support the effectiveness of the proposed ICD formulation and its gated chromaticity regulation.

\begin{table*}[t]
	\centering
	\caption{Quality--efficiency comparison on LOLv2-Real. Inference time is measured on an NVIDIA RTX 4060 Ti GPU with \(600 \times 400\) input images.}
	\label{tab:lolv2real_hvi_compare}
	\setlength{\tabcolsep}{6pt}
	\renewcommand{\arraystretch}{1.2}
	\small
	
	\begin{tabular}{ccccccccc}
		\toprule
		Method & PSNR$\uparrow$ & SSIM$\uparrow$ & LPIPS$\downarrow$ & Time (s)$\downarrow$ & FPS$\uparrow$ & Params (M)$\downarrow$ & FLOPs (G)$\downarrow$ & Type \\
		\midrule
		KinD++~\cite{zhang2021KIND++}             
		& 17.660 & 0.760 & 0.768 & 0.9611 & 1.04  & 8.27 & 2719.57 & CNN \\ 
		
		
		LEDNet~\cite{zhou2022lednet}              
		& 23.394 & 0.837 & 0.115 & \textcolor{blue}{0.0189} & \textcolor{blue}{52.91} & 7.45 & 122.92 & CNN \\
		
		NeRCo~\cite{NeRCo_2023_ICCV}              
		& 25.170 & 0.790 & 0.108 & 0.1960 & 5.10  & 11.39 & 183.96 & CNN \\
		
		GSAD~\cite{hou2023GSAD}                  
		& 23.715 & \textcolor{blue}{0.876} & \textcolor{blue}{0.103} & 0.1097 & 9.12  & 17.36 & 350.46 & Diffusion \\
		
		SNR-Aware~\cite{xu2022snr}               
		& 22.251 & 0.840 & 0.117 & 0.0244 & 40.98 & 39.12 & 261.79 & CNN+Transformer \\
		
		
		LLFormer~\cite{wang2023LLFormer}         
		& \textcolor{blue}{27.750} & 0.860 & 0.117 & 0.0483 & 20.70 & 24.52 & 81.77 & Transformer \\
		
		HVI~\cite{yan2023hvi}                    
		& 24.111 & 0.871 & 0.108 & 0.0617 & 16.21 & 1.98 & \textcolor{blue}{29.78} & Transformer \\
		
		Retinexformer~\cite{cai2023retinexformer} 
		& 22.790 & 0.840 & 0.108 & 0.0242 & 41.32 & \textcolor{blue}{1.61} & 57.72 & Transformer \\
		
		\midrule
		Ours                                     
		& \textcolor{red}{29.710} & \textcolor{red}{0.8895} & \textcolor{red}{0.102} & \textcolor{red}{0.0133} & \textcolor{red}{75.19} & \textcolor{red}{1.10} & \textcolor{red}{5.57} & Transformer \\
		
		\bottomrule
	\end{tabular}
		\vspace{-1em}
\end{table*}
%
\subsection{Efficiency Analysis}
\label{sec:efficiency_analysis}

We analyze the quality--efficiency trade-off of the proposed method on LOLv2-Real. 
Table~\ref{tab:lolv2real_hvi_compare} reports quantitative performance, inference time, FPS, parameter count, and FLOPs for representative methods.
ICD achieves the best PSNR, SSIM, and LPIPS among the compared methods. 
It obtains 29.710 dB PSNR, 0.8895 SSIM, and 0.102 LPIPS. 
In terms of efficiency, ICD requires 0.0133 s per image on an NVIDIA RTX 4060 Ti, corresponding to 75.19 FPS. 
It is faster than all compared baselines in Table~\ref{tab:lolv2real_hvi_compare}. 
ICD also has the lowest model complexity among the listed methods, with 1.10M parameters and 5.57G FLOPs.

Compared with HVI~\cite{yan2023hvi}, ICD improves PSNR from 24.111 dB to 29.710 dB and SSIM from 0.871 to 0.8895, while reducing LPIPS from 0.108 to 0.102. 
The inference time is reduced from 0.0617 s to 0.0133 s, and FLOPs decrease from 29.78G to 5.57G. 
Compared with Retinexformer~\cite{cai2023retinexformer}, ICD achieves higher PSNR, SSIM, and LPIPS with fewer parameters and lower FLOPs.
The results in Tables~\ref{tab:lolv2real_hvi_compare} show that ICD improves enhancement quality while maintaining low model complexity and practical inference efficiency.

\begin{table*}[t]
	\centering
	\caption{Mapping definitions of different variants in the decoupled mapping ablation. The table lists only the core mapping forms; unified clipping operations for numerical stability and physical boundary constraints are retained for all variants but omitted for compactness.}
	
	\label{tab:mapping_variants}
	\setlength{\tabcolsep}{10pt}
	\renewcommand{\arraystretch}{1.5}
	\small
	
	\begin{tabular}{c c c c}
		\toprule
		\textbf{Variant} & \textbf{Type} & \textbf{Intensity Mapping} \(I_{\max,\mathrm{out}}\) & \textbf{Chromaticity Mapping} \(\mathbf{C}_{\mathrm{out}}\) \\
		\midrule
		
		Residual mapping
		& Joint
		& \(I_{\max,\mathrm{in}}+\Delta I\)
		& \(\mathbf{C}_{\mathrm{in}}+\Delta\mathbf{C}\)
		\\
		
		End-to-end mapping
		& Joint
		& \(\Delta I\)
		& \(\Delta\mathbf{C}\)
		\\
		
		Intensity division
		& Intensity
		& \(I_{\max,\mathrm{in}}/L\)
		& \(\mathbf{C}_{\mathrm{in}}+\Delta\mathbf{C}\)
		\\
		
		Intensity fractional mapping
		& Intensity
		& \(\displaystyle \frac{u\,I_{\max,\mathrm{in}}}{u\,I_{\max,\mathrm{in}}+(1-I_{\max,\mathrm{in}})+\epsilon}\)
		& \(\mathbf{C}_{\mathrm{in}}+\Delta\mathbf{C}\)
		\\
		
		Intensity quadratic mapping
		& Intensity
		& \(I_{\max,\mathrm{in}} + a\,I_{\max,\mathrm{in}}(1-I_{\max,\mathrm{in}})\)
		& \(\mathbf{C}_{\mathrm{in}}+\Delta\mathbf{C}\)
		\\
		
		Chromaticity gamma mapping
		& Chromaticity
		& \(I_{\max,\mathrm{in}}+\Delta I\)
		& \(\boldsymbol{\gamma}\odot \mathbf{C}_{\mathrm{in}}\)
		\\
		
		Gated chromaticity residual
		& Chromaticity
		& \(I_{\max,\mathrm{in}}+\Delta I\)
		& \(\mathbf{C}_{\mathrm{in}}+w\odot\Delta\mathbf{C}\)
		\\
		
		Chromaticity affine mapping
		& Chromaticity
		& \(I_{\max,\mathrm{in}}+\Delta I\)
		& \(\boldsymbol{\alpha}\odot\mathbf{C}_{\mathrm{in}}+\boldsymbol{\beta}\)
		\\
		
		\bottomrule
	\end{tabular}
	
\end{table*}

\subsection{Ablation on Decoupled Mapping Strategies}
\label{sec:mapping_strategy_ablation}

We compare different mapping strategies in the same ICD space to analyze the effect of intensity and chromaticity update forms. 
All variants share the decoupled representation and constrained inverse reconstruction defined in Sections~\ref{sec:icd_decoupling} and~\ref{sec:constrained_reconstruction}; only the mappings for \(I_{\max,\mathrm{out}}\) and \(\mathbf{C}_{\mathrm{out}}\) are changed. 
Table~\ref{tab:mapping_variants} summarizes the core mapping forms of the evaluated variants. 
For clarity, the unified clipping operations used for numerical stability and physical boundary constraints are omitted from the table.

Table~\ref{tab:quant_mapping_variants} reports the quantitative comparison of different decoupled mapping strategies on the LOLv1~\cite{Chen2018LOLv1} test set. 
Residual mapping achieves the best full-reference performance, with 24.0799 dB PSNR, 0.8463 SSIM, 0.0754 Rel-MAE, 0.1096 LPIPS, and the lowest MSE of 0.0073. 
These results suggest that residual updates for both the intensity envelope and the log-chromaticity component provide a relatively stable mapping strategy in the ICD space.

From the intensity branch perspective, residual updating based on the input intensity envelope \(I_{\max,\mathrm{in}}\) performs better than direct regression, division mapping, and nonlinear intensity transformations. 
End-to-end mapping reduces PSNR and SSIM to 22.9178 dB and 0.8368, respectively, indicating that directly regressing the target intensity weakens the constraint provided by the input intensity prior. 
Intensity division, quadratic, and fractional mappings obtain PSNR values of 22.6549 dB, 22.8594 dB, and 23.0753 dB, respectively, all lower than residual mapping. 
Although the intensity fractional mapping obtains the lowest NIQE and PI, suggesting improved no-reference naturalness, its full-reference metrics remain lower than those of residual mapping. 
This reflects a trade-off between reference fidelity and no-reference perceptual naturalness.

From the chromaticity branch perspective, additive residual mapping in the ICD space shows better stability. 
Chromaticity affine mapping is the closest variant to residual mapping, achieving 23.7866 dB PSNR and 0.8434 SSIM. 
However, it does not improve LPIPS, NIQE, or PI over residual mapping, suggesting that a more flexible chromaticity transformation does not necessarily lead to better perceptual quality. 
Chromaticity gamma mapping degrades more substantially, with PSNR and SSIM dropping to 21.3830 dB and 0.6673, respectively. 
This suggests that a simple nonlinear remapping of chromaticity is insufficient to compensate for chromatic shifts caused by low-light degradation and noise. 
Gated chromaticity residual performs the worst, indicating that excessive suppression of chromaticity updates in dark regions may weaken necessary color recovery.
\begin{figure*}[t]
	\centering
	\includegraphics[width=1.0\textwidth]{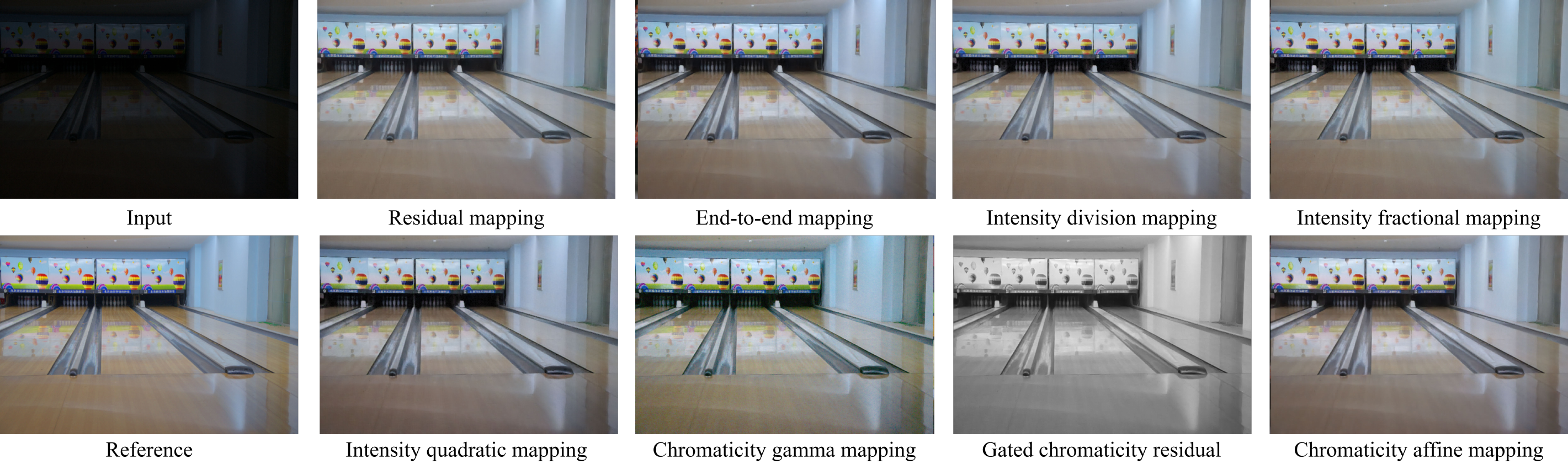}
	\caption{Visual comparison of different decoupled mapping strategies.}
	\label{fig:mapping_visual_results}
	\vspace{-1em}
\end{figure*}

\begin{table*}[t]
	\centering
	\caption{Quantitative comparison of different decoupled mapping strategies on the LOLv1~\cite{Chen2018LOLv1} test set.}
	\label{tab:quant_mapping_variants}
	\setlength{\tabcolsep}{6pt}
	\renewcommand{\arraystretch}{1.15}
	\begin{tabular}{cccccccc}
		\toprule
		Method & PSNR$\uparrow$ & SSIM$\uparrow$ & MSE$\downarrow$ & Rel-MAE$\downarrow$ & LPIPS$\downarrow$ & NIQE$\downarrow$ & PI$\downarrow$ \\
		\midrule
		Residual mapping & \textbf{26.1978} & \textbf{0.8567} & \textbf{0.0073} & \textbf{0.0754} & \textbf{0.1061} & 4.1603 & 3.4465 \\
		Chromaticity affine mapping & 25.8787 & 0.8538 & 0.0080 & 0.0758 & 0.1203 & 4.2258 & 3.5067 \\
		Intensity fractional mapping & 25.1048 & 0.8342 & 0.0104 & 0.0873 & 0.1201 & \textbf{3.9969} & \textbf{3.1204} \\
		End-to-end mapping & 24.9335 & 0.8471 & 0.0111 & 0.0868 & 0.1159 & 4.0803 & 3.3764 \\
		Intensity quadratic mapping & 24.8700 & 0.8323 & 0.0088 & 0.0823 & 0.1342 & 4.1120 & 3.1395 \\
		Intensity division mapping & 24.6475 & 0.8312 & 0.0107 & 0.0887 & 0.1231 & 4.2558 & 3.2163 \\
		Chromaticity gamma mapping & 23.2637 & 0.6755 & 0.0122 & 0.0981 & 0.1944 & 6.7770 & 4.1227 \\
		Gated chromaticity resual & 21.3153 & 0.8105 & 0.0189 & 0.1101 & 0.2930 & 4.1989 & 3.4158 \\
		\bottomrule
	\end{tabular}
	\vspace{-1em}
\end{table*}

Fig.~\ref{fig:mapping_visual_results} presents visual comparisons of different mapping strategies. 
Residual mapping produces results closer to the reference image in terms of brightness recovery, color stability, and detail preservation. 
End-to-end mapping and several intensity-based variants show less stable tonal distributions, while chromaticity-based variants tend to affect color rendition. 
In particular, chromaticity gamma mapping and gated chromaticity residual introduce more visible color shifts or saturation artifacts. 
Chromaticity affine mapping performs closer to residual mapping, but still shows slightly weaker visual quality, which is consistent with the quantitative results.
Based on these results, residual mapping is adopted as the default decoupled mapping strategy in the proposed method. 

\newcommand{\panelFive}[2]{%
	\shortstack[t]{%
		\includegraphics[width=0.19\textwidth]{#1}\\[0.3ex]
		\small #2%
	}%
}

%
%
%

\begin{figure*}[t]
	\centering
	\setlength{\tabcolsep}{1pt}
	\renewcommand{\arraystretch}{1.0}
	
	\begin{tabular}{ccccc}
		
		\panelFive{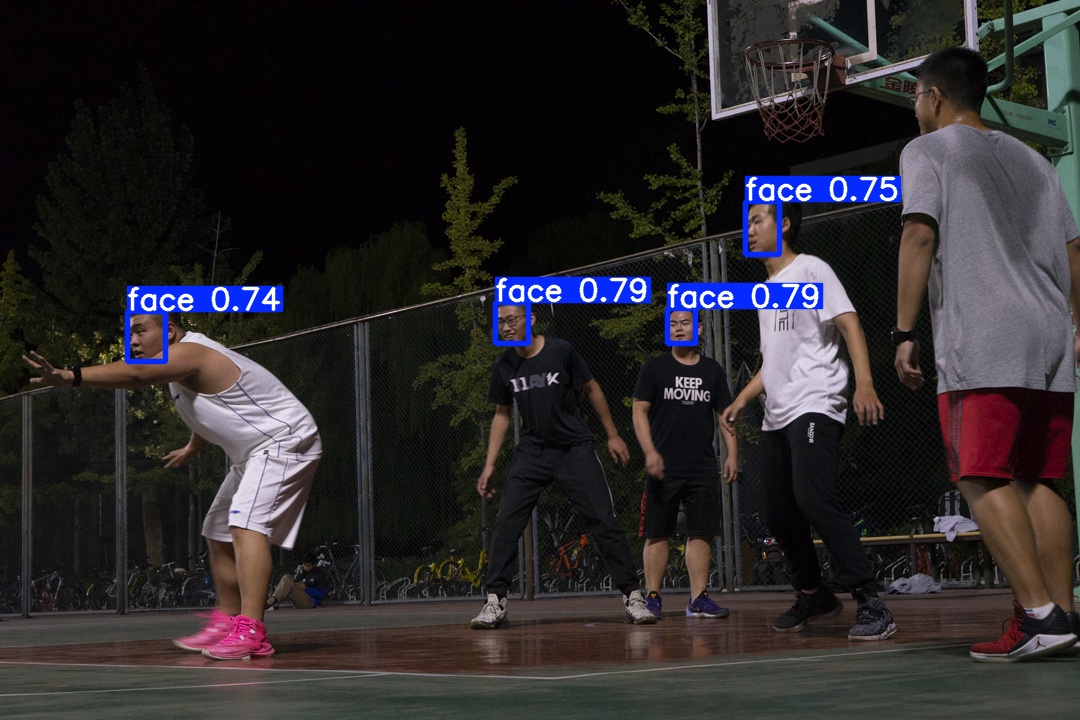}{Input} &
		\panelFive{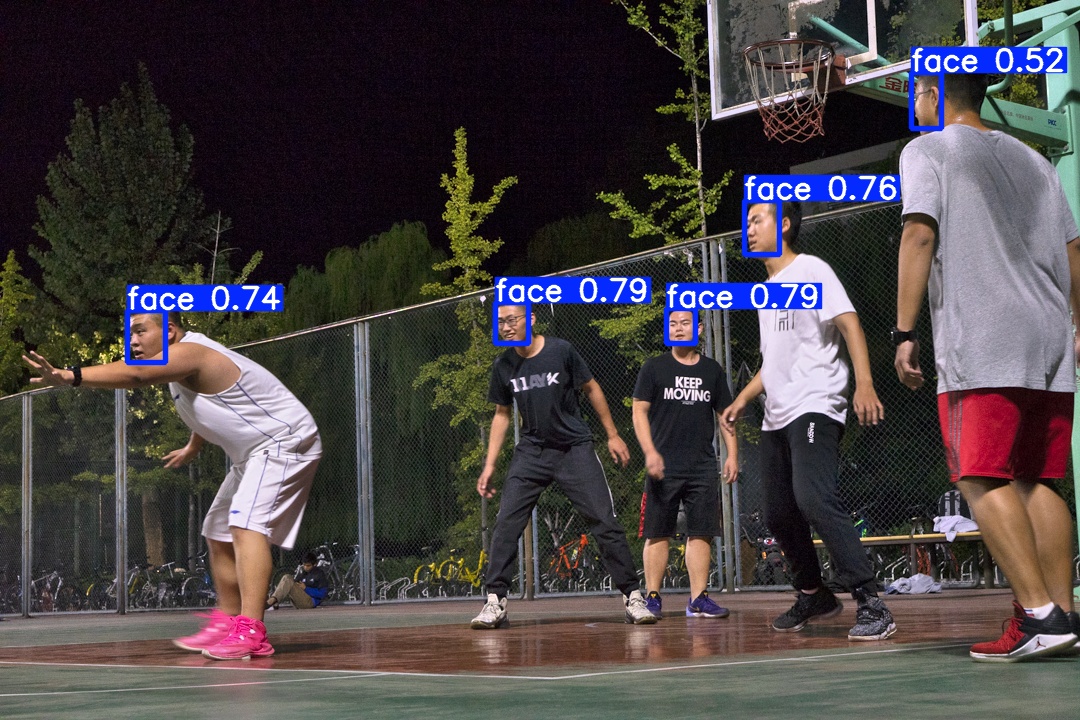}{CoLIE~\cite{chobola2025CoLIE}} &
		\panelFive{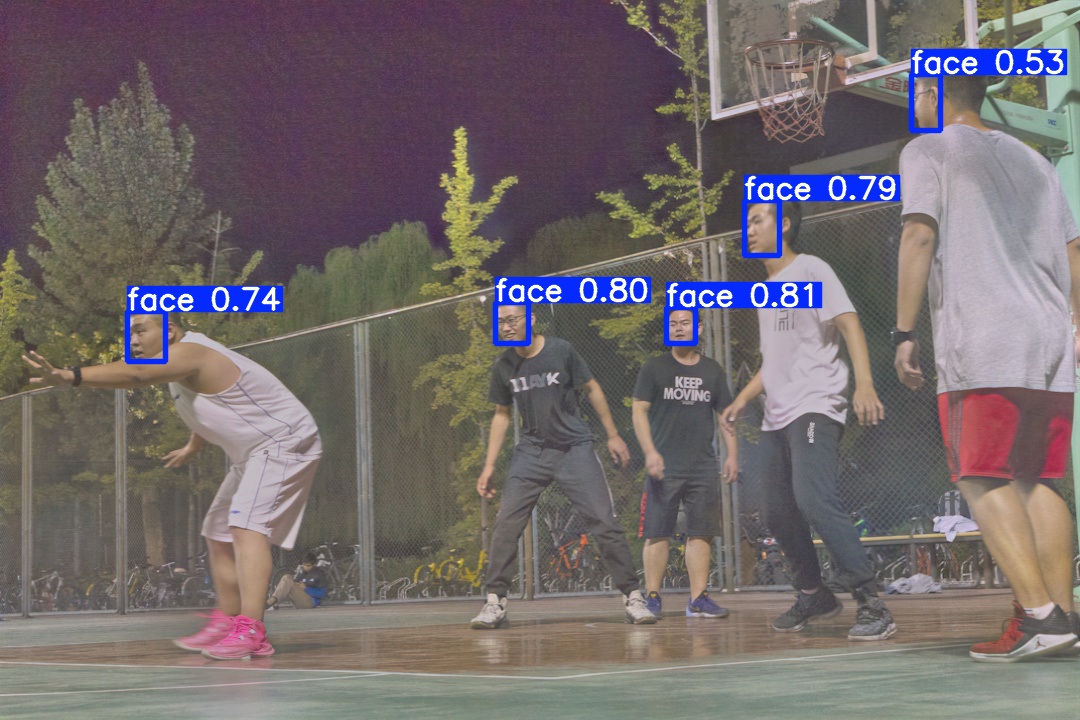}{EnlightenGAN~\cite{jiang2021enlightengan}} &
		\panelFive{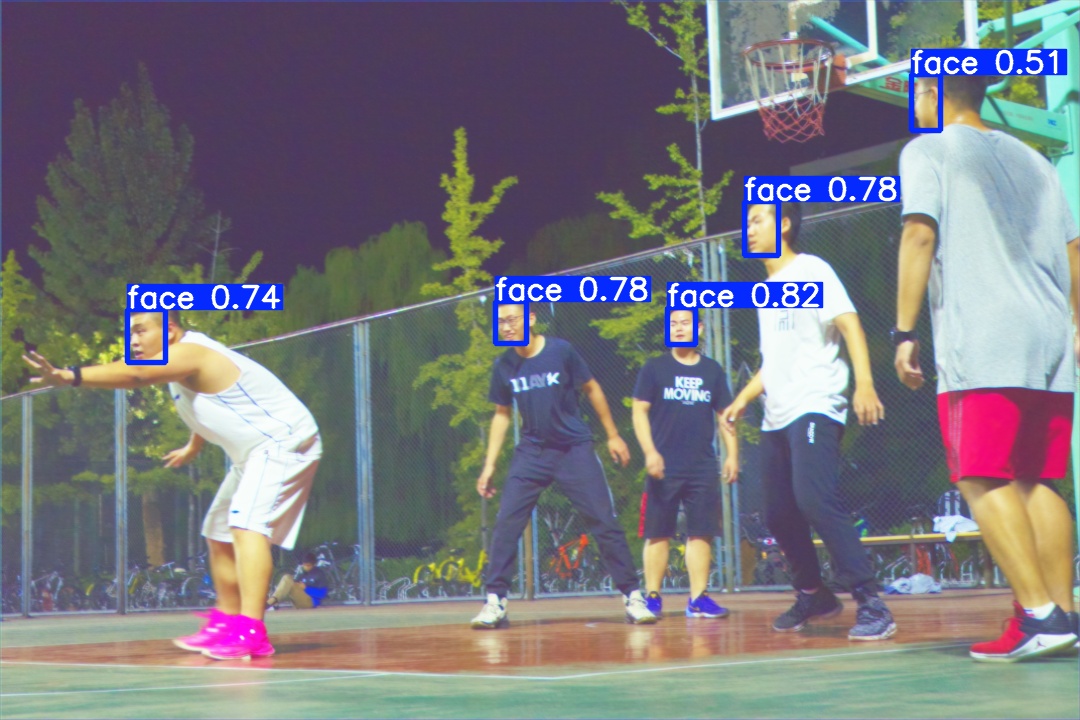}{NoiSER~\cite{zhang2024NoiSER}} &
		\panelFive{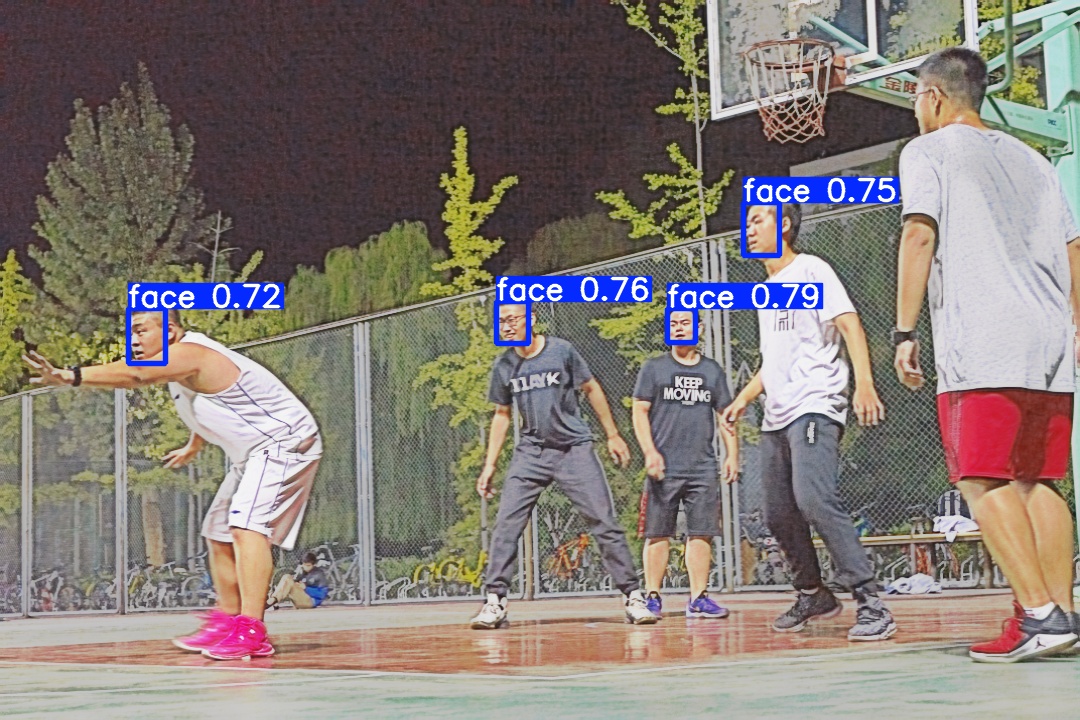}{PairLIE~\cite{fu2023PairLIE}} \\[2pt]
		
		\panelFive{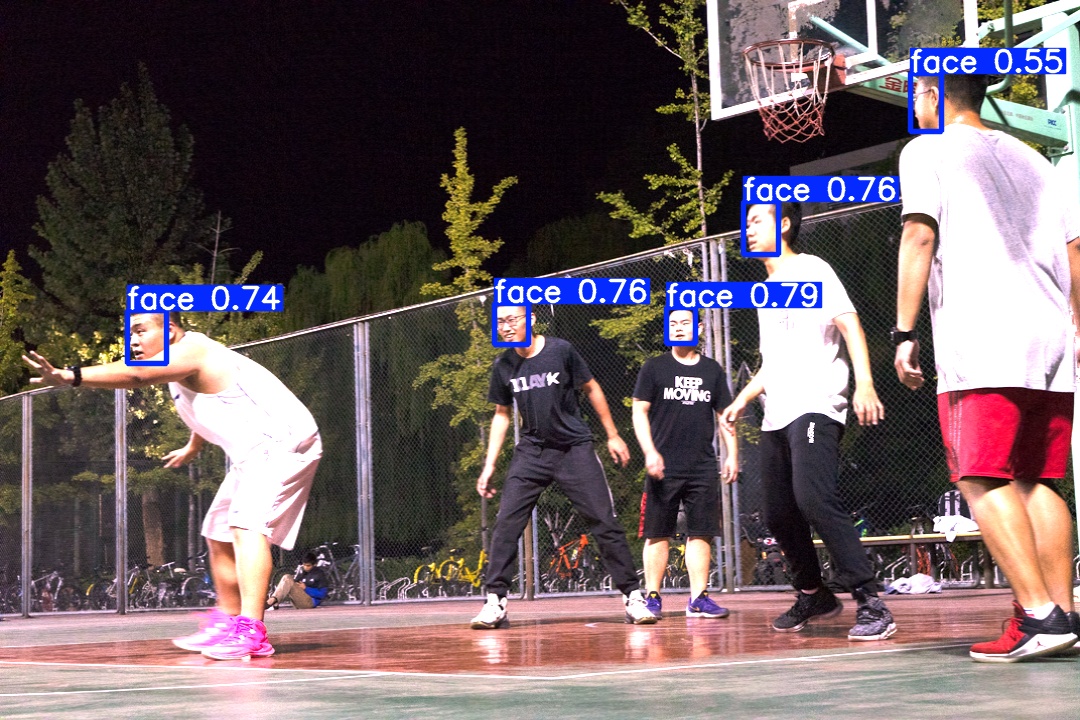}{RUAS~\cite{liu2021ruas}} &
		\panelFive{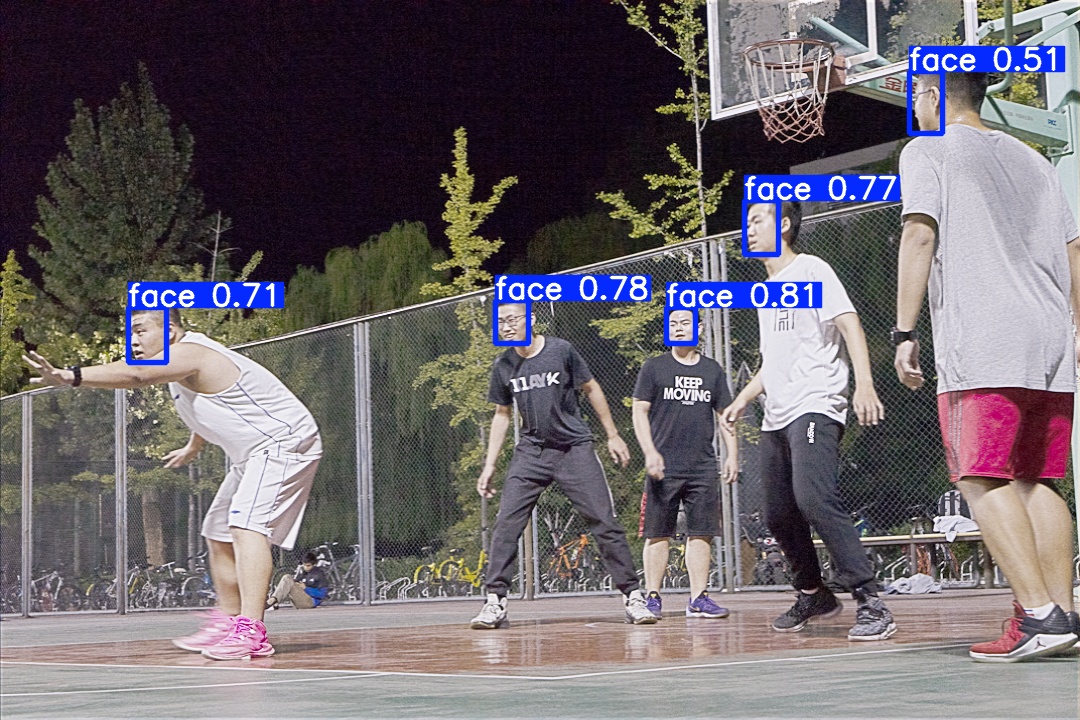}{SCI~\cite{ma2022SCI}} &
		\panelFive{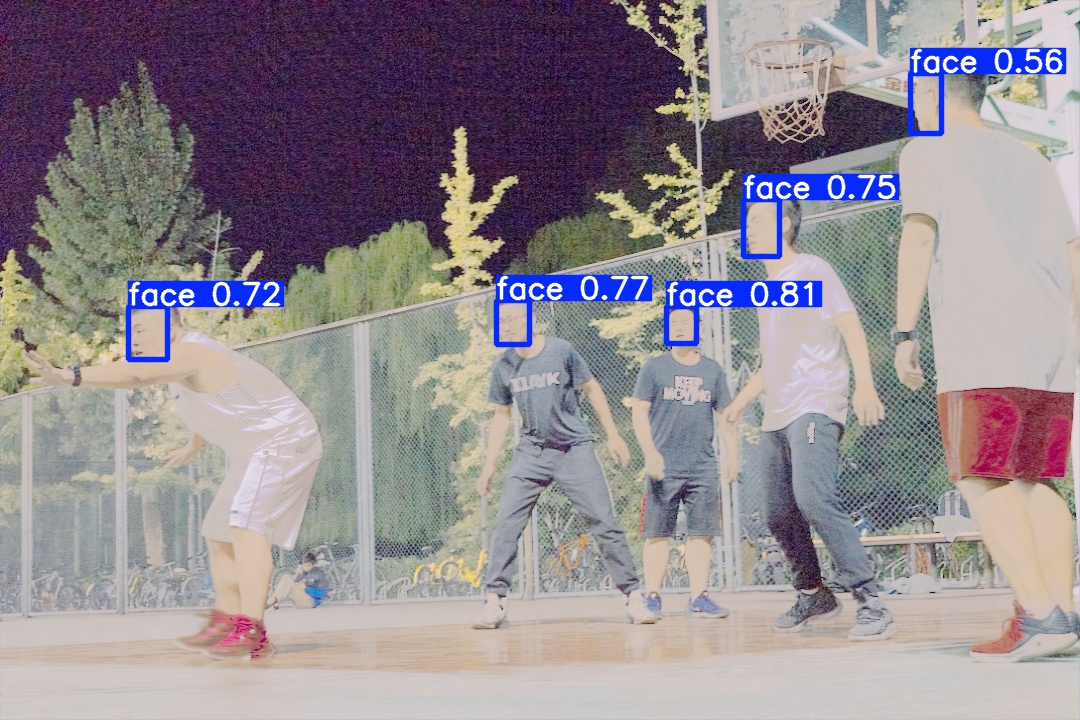}{SCLM~\cite{zhang2023SCLM}} &
		\panelFive{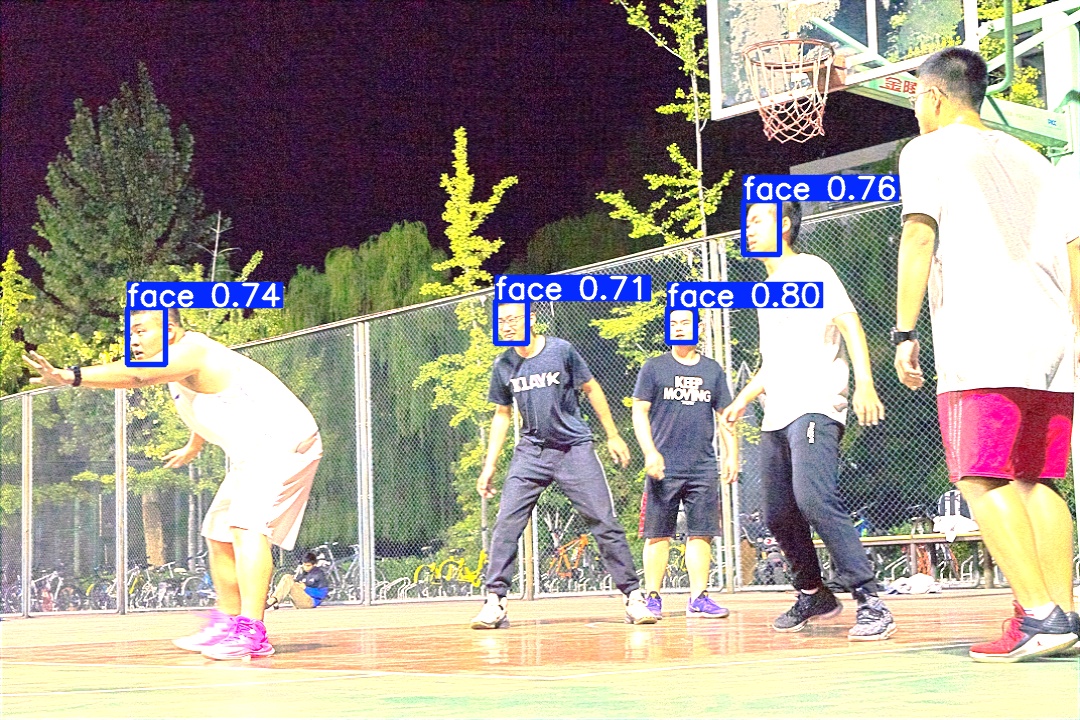}{ZeroIG~\cite{shi2024zeroIG}} &
		\panelFive{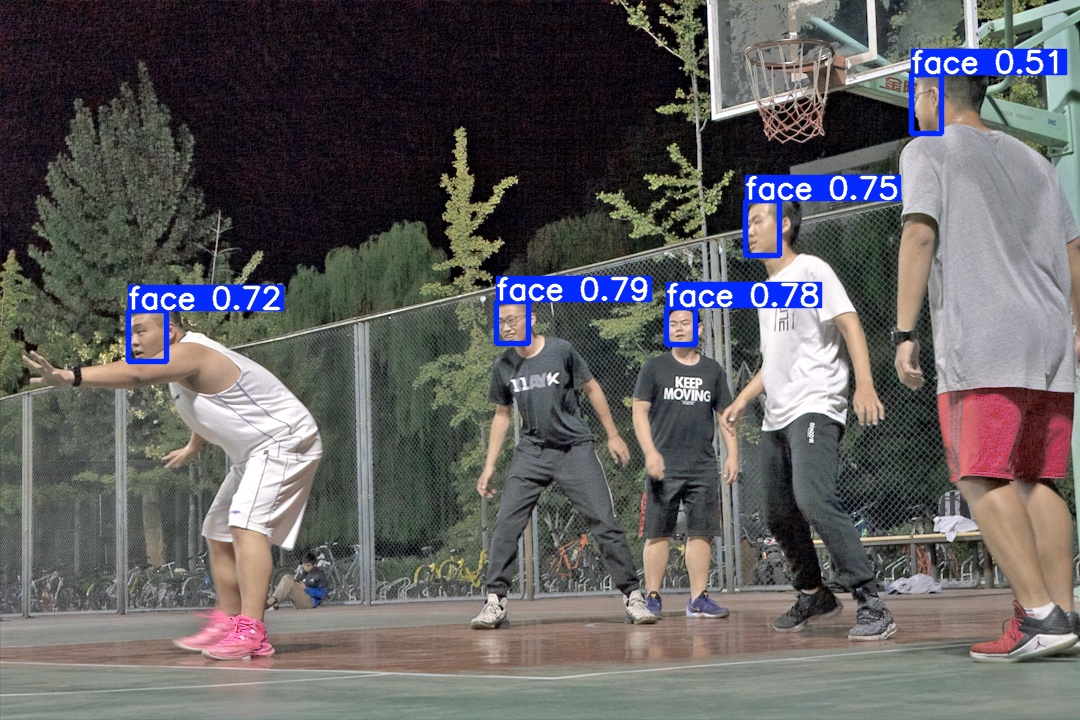}{\textbf{Ours}} \\
		
	\end{tabular}
	
 \caption{Visual comparison of face detection results on a backlit image from DarkFace.}
 \label{fig:face_detection_darkface697}
 	\vspace{-1em}
\end{figure*}

\newcommand{\panelThree}[2]{%
	\shortstack[t]{%
		\includegraphics[width=0.32\textwidth]{#1}\\[0.3ex]
		\small #2%
	}%
}

\begin{figure*}[t]
	\centering
	\setlength{\tabcolsep}{2pt}
	\renewcommand{\arraystretch}{1.0}
	
	\begin{tabular}{ccc}
		\panelThree{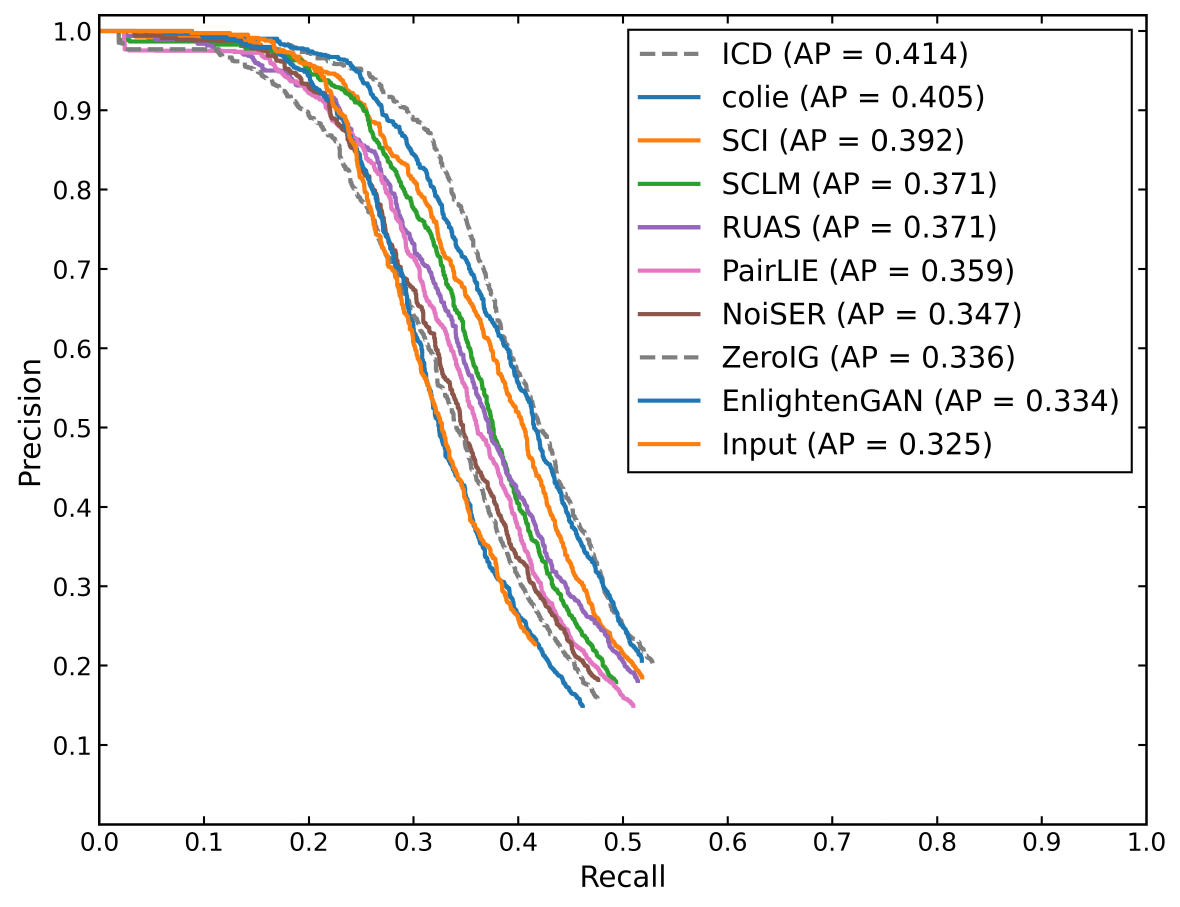}{PR curve, IoU = 0.3} &
		\panelThree{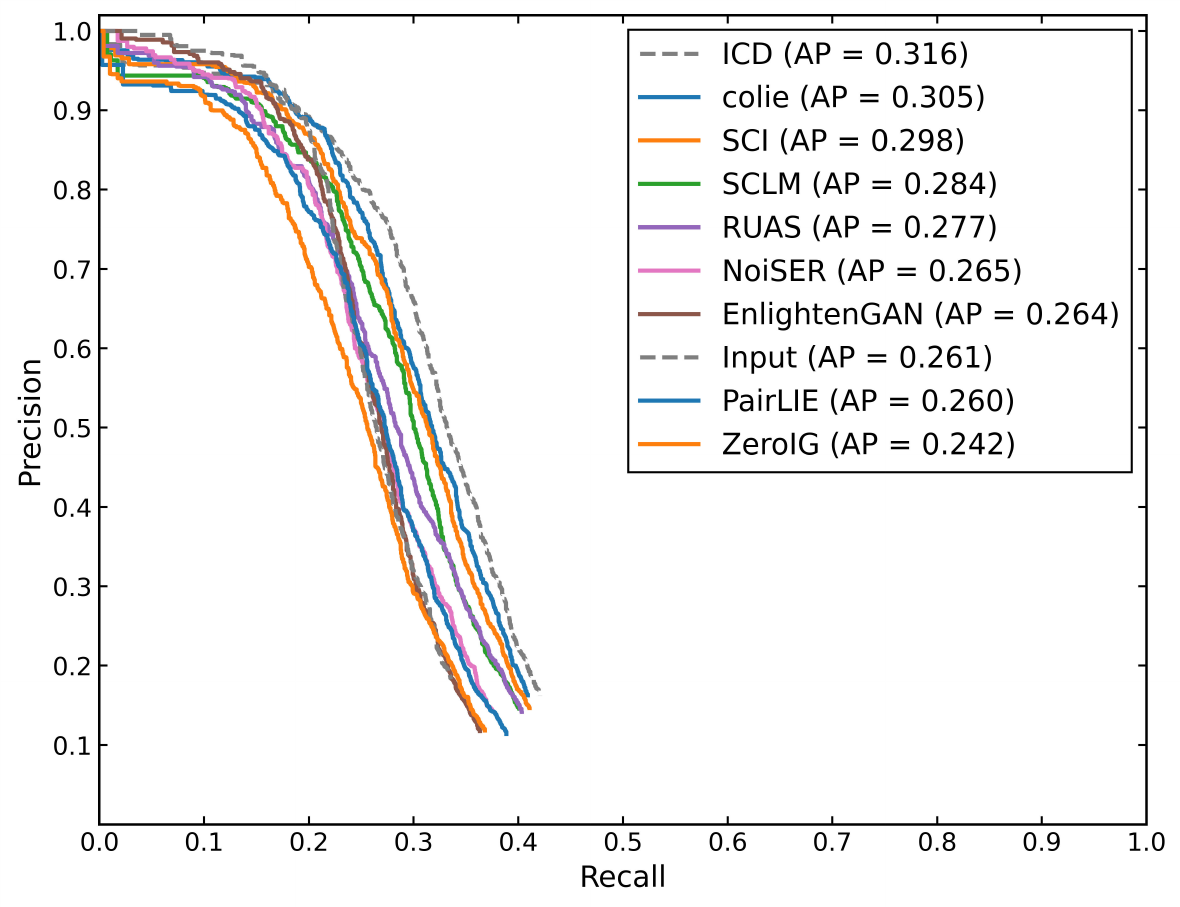}{PR curve, IoU = 0.5} &
		\panelThree{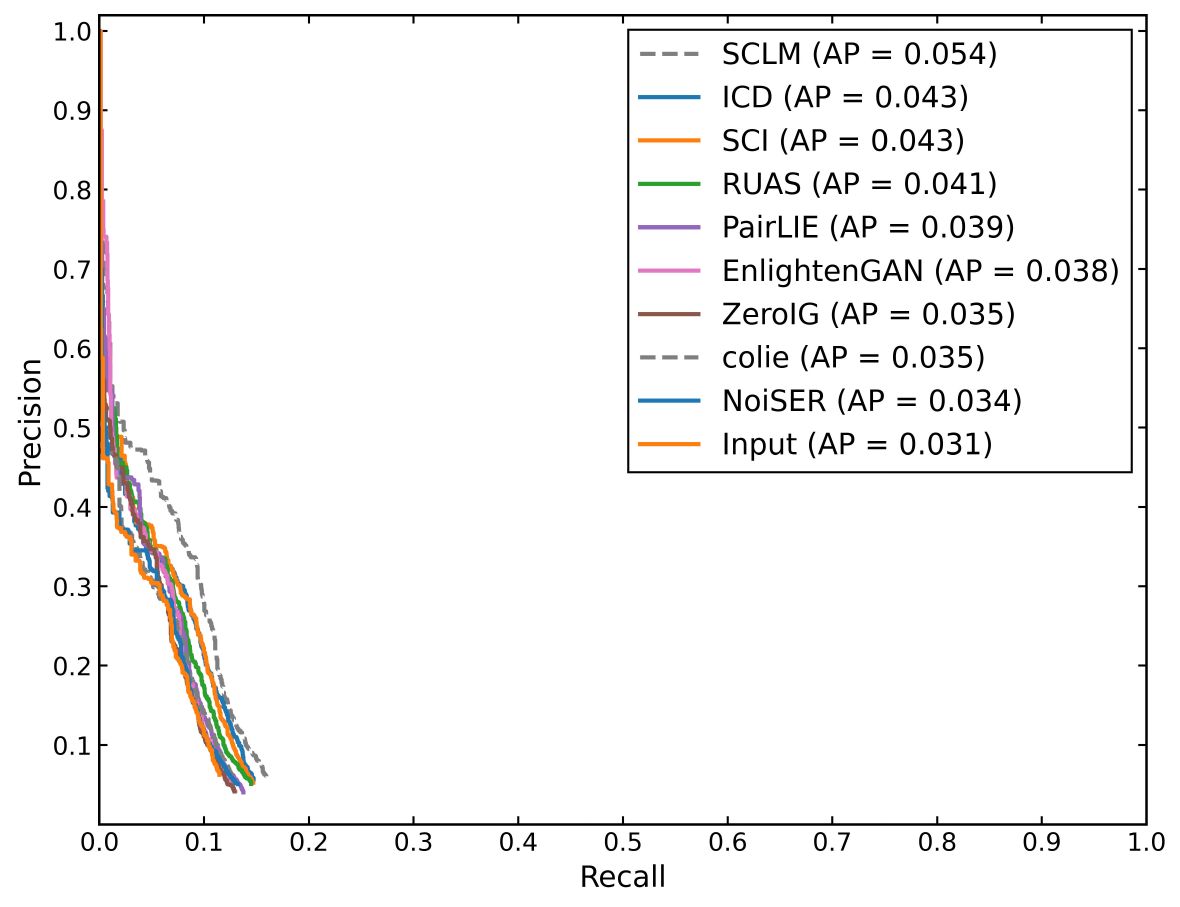}{PR curve, IoU = 0.7}
	\end{tabular}
 
	\caption[PR curves under different IoU thresholds]{
		PR curves under different IoU thresholds on the DarkFace dataset. 
		The curves report detection performance at IoU thresholds of 0.3, 0.5, and 0.7.
	}
	\label{fig:darkface_pr_curves}
	\vspace{-1em}
\end{figure*}

\subsection{DarkFace Face Detection}
\label{sec:darkface_detection}

To evaluate the effect of ICD on downstream detection, we conduct face detection experiments on the DarkFace dataset~\cite{DARKFACE}. 
The detector is implemented using YOLOv8 from Ultralytics~\cite{ultralytics_yolov8_docs,ultralytics_yolov8_repo}. 
Precision--recall curves are reported under IoU thresholds of 0.3, 0.5, and 0.7, as shown in Fig.~\ref{fig:darkface_pr_curves}.

ICD achieves favorable performance under loose and moderate matching criteria. 
At IoU=0.3, ICD obtains an AP of 0.414, outperforming CoLIE~\cite{chobola2025CoLIE} and SCI~\cite{ma2022SCI}, which obtain AP values of 0.405 and 0.392, respectively. 
At IoU=0.5, ICD achieves an AP of 0.316, which is also higher than CoLIE~\cite{chobola2025CoLIE} and SCI~\cite{ma2022SCI}, with AP values of 0.305 and 0.298, respectively. 
These results suggest that ICD can improve the visibility of low-light face regions while retaining local contrast cues that are useful for detection.

When the IoU threshold increases to 0.7, the AP values of all methods decrease substantially, indicating that accurate bounding-box localization remains challenging under low-light conditions. 
In this stricter setting, SCLM~\cite{zhang2023SCLM} obtains the highest AP of 0.054, while ICD achieves an AP of 0.043. 
Although ICD does not achieve the best result at this threshold, its performance remains comparable to SCI~\cite{ma2022SCI} and higher than RUAS~\cite{liu2021ruas} and PairLIE~\cite{fu2023PairLIE}. 
Overall, ICD shows clear advantages at IoU=0.3 and IoU=0.5, while maintaining reasonable performance at IoU=0.7. 
This result indicates that ICD is beneficial for low-light face detection, whereas high-precision localization under severe low-light degradation remains difficult.

Visual examples are shown in Fig.~\ref{fig:face_detection_darkface697}. 
The qualitative results are generally consistent with the quantitative trends. 
In the original low-light images, face targets located in dark regions are difficult to detect, leading to missed detections. 
Some enhancement methods improve global brightness but introduce over-enhancement, local whitening, or structural degradation, which may interfere with face localization. 
For example, EnlightenGAN~\cite{jiang2021enlightengan} and ZeroIG~\cite{shi2024zeroIG} exhibit noticeable brightness imbalance and detail degradation in some cases. 
In contrast, ICD improves the visibility of dark regions while preserving local structural information and contrast, allowing more face instances in low-light regions to be correctly detected.

\section{Conclusion}
\label{sec:conclusion}

This paper presents an intensity--chromaticity decoupled framework for low-light image enhancement. 
Instead of directly enhancing coupled RGB intensities, we formulate low-light enhancement in a log-domain decoupled space, where intensity recovery and chromatic correction are modeled separately. 
This representation helps mitigate noise amplification, exposure inconsistency, color shift, and detail degradation in complex low-light scenes. 
Based on this formulation, we develop a dual-branch interaction network to estimate intensity and chromaticity components, followed by constrained RGB reconstruction for improved color stability and chromatic noise suppression.
Extensive experiments on multiple public benchmarks validate the effectiveness of the proposed method. 
ICD achieves favorable performance in terms of PSNR, SSIM, and LPIPS, and produces stable visual results under different low-light conditions. 
Ablation studies further verify the contributions of the decoupled representation, residual mapping strategy, and intensity--chromaticity consistency losses. 
Experiments on DarkFace also show that the enhanced images can improve downstream face detection in low-light environments.
Despite its effectiveness, the computational cost of ICD increases with image resolution, which may limit its deployment on resource-constrained platforms. 
Future work will focus on more efficient architectures and deployment-oriented optimization for high-resolution and edge-device applications.

\printcredits

\bibliographystyle{cas-model2-names}

\bibliography{cas-refs}





\end{document}